\documentclass[5p,twocolumn]{elsarticle}

\usepackage[T1]{fontenc}

\usepackage[english]{babel}

\usepackage{amssymb}

\usepackage{amsmath}

\usepackage{graphicx}

\usepackage{booktabs}

\usepackage{multirow}

\usepackage{xcolor}

\usepackage{algorithm}

\usepackage{algorithmic}

\usepackage{float}

\usepackage{placeins}

\usepackage{lineno}

\usepackage{url}

\usepackage{hyperref} 



\hypersetup{ 
colorlinks=true,
linkcolor=blue,
urlcolor=blue,
citecolor=blue,
pageanchor=true,
unicode=true
}

\usepackage{booktabs}

\usepackage{tabularx}

\usepackage{siunitx}

\usepackage[table]{xcolor}

\makeatletter

\def\ps@pprintTitle{%

\let\@oddhead\@empty

\let\@evenhead\@empty

\def\@oddfoot{\hfil\thepage\hfil}%

\let\@evenfoot\@oddfoot

}

\def\@maketitle{%

\newpage

\null \vspace*{-10pt}%

\vspace{16pt}%

\begin{center}%

{\LARGE\bfseries\@title\par}%

\vspace{12pt}%

{\large\@author\par}%

\end{center}%

\vspace{8pt}%

}

\makeatother

\usepackage{fancyhdr}

\fancypagestyle{firstpage}{%

\fancyhf{}

\fancyfoot[L]{}  

}

\fancypagestyle{resultsreproduction}{%

\fancyhf{}

\fancyfoot[C]{\hfil\thepage\hfil}

}

\fancypagestyle{conclusionfooter}{%

\fancyhf{}

\fancyfoot[C]{\hfil\thepage\hfil}

}

\begin{document}

\thispagestyle{firstpage}

\begin{frontmatter}


\title{Mecellem Models: Turkish Models Trained from Scratch and Continually Pre-trained for the Legal Domain}


\author[a]{\"{O}zg\"{u}r U\u{g}ur}

\author[a]{Mahmut G\"{o}ksu}

\author[a]{Mahmut \c{C}imen}

\author[a]{Musa Y{\i}lmaz}

\author[a]{Esra \c{S}avirdi}

\author[a]{Alp Talha Demir}

\author[a]{Rumeysa G\"{u}ll\"{u}ce}

\author[a]{\.{I}clal \c{C}etin}

\author[a]{\"{O}mer Can Sa\u{g}ba\c{s}}

\cortext[cor1]{Corresponding author. E-mail: aiteam@newmind.ai}

\affiliation[a]{organization={NewmindAI}, country={T\"{u}rkiye}}


\begin{abstract}

This paper presents Mecellem models, a framework for developing specialized language models for the Turkish legal domain through domain adaptation strategies. We make two contributions: (1)\textbf{Encoder Model Pre-trained from Scratch}: ModernBERT-based bidirectional encoders pre-trained on a Turkish-dominant corpus of 112.7 billion tokens. Effective domain adaptation requires joint evaluation of pre-training objectives and downstream task performance. Rather than relying solely on training loss minimization, we implement a checkpoint selection strategy that evaluates downstream retrieval performance throughout training, revealing that optimal checkpoints achieve best retrieval scores before pre-training loss reaches its minimum. This demonstrates the importance of integrating downstream evaluation into pre-training checkpoint selection. Our encoder models achieve top-3 rankings on the Turkish retrieval leaderboard, with smaller models (155M parameters) achieving comparable performance to larger reference models (307M-567M parameters). Our approach achieves 92.36\% production efficiency compared to state-of-the-art models (embeddinggemma-300m: 100.00\%, BAAI/bge-m3: 99.54\%, newmindai/bge-m3-stsb: 94.38\%), ranking fourth overall despite requiring less computational resources. SOTA models rely on multi-stage, computationally intensive training pipelines involving RetroMAE, contrastive learning, and unified fine-tuning phases, making our single-stage pre-training followed by efficient post-training approach a cost-effective alternative; (2)\textbf{Decoder Model with Continual Pre-training (CPT)}: Qwen3-1.7B and Qwen3-4B models adapted to Turkish legal domain through controlled curriculum learning. Domain adaptation is achieved through four-phase CPT with optimal sample ratios identified via initialization ablation studies, enabling gradual transition from general language knowledge to specialized legal terminology and long-context reasoning. This approach achieves 36.2\% perplexity reduction on Turkish legal text, demonstrating domain adaptation gains.

\end{abstract}



\end{frontmatter}


\section{Introduction}

\label{sec:introduction}

Large Language Models (LLMs) acquire powerful generalization capabilities through large-scale, multilingual pre-training. However, when these models are predominantly trained on English-centric data, performance degradation is observed in non-English languages and highly formal domains. The legal domain presents particular challenges for domain adaptation due to intensive terminology usage, long and complex sentence structures, and strict contextual and normative constraints. These challenges become even more pronounced in morphologically rich languages such as Turkish, where syntactic variation and inflectional complexity demand more precise linguistic modeling \cite{turkish2024}.

The rapid advancement of LLMs has transformed natural language processing applications across various domains \cite{emergent2022,gpt4}. However, the Turkish legal domain remains underserved, lacking specialized language resources essential for building effective AI-powered legal tools. This gap is critical given the growing demand for intelligent legal information systems, automated document analysis, and Retrieval-Augmented Generation (RAG) applications tailored to Turkey's unique legal framework.

Legal text processing presents distinct challenges compared to general domain NLP \cite{legalai2020,legalbert2020}. Legal documents employ specialized terminology, complex syntactic structures, and domain-specific reasoning patterns that require dedicated training approaches. These characteristics necessitate consideration of corpus composition, training objectives, and evaluation metrics to ensure models capture nuanced linguistic patterns inherent in legal discourse. Effective adaptation requires balancing domain-specific knowledge acquisition with preservation of general language capabilities, critical for morphologically rich languages where linguistic complexity amplifies adaptation challenges.

In this work, we introduce Mecellem, a framework for developing Turkish legal language models through two complementary approaches.

First, we develop ModernBERT-based bidirectional encoder models (155M base, 403M large parameters) pre-trained entirely from scratch on a carefully curated Turkish-dominant corpus totaling 112.7 billion tokens. The pre-training process employs Masked Language Modeling (MLM) as the primary objective, achieving competitive accuracy scores. Rather than relying solely on pre-training loss minimization, we implement a checkpoint selection strategy that evaluates downstream retrieval performance throughout training. This approach reveals non-linear relationships between pre-training loss and embedding quality: optimal checkpoints achieve the best retrieval scores before the training loss reaches its minimum. These pre-trained encoders are subsequently post-trained for embedding tasks using multiple contrastive learning techniques, including standard InfoNCE, Qwen3-style InfoNCE, and GISTEmbed with cached guide models (BGE-M3 and EmbeddingGemma-300M), achieving competitive performance on Turkish legal retrieval benchmarks.

Second, we apply Continual Pre-training (CPT) to Qwen3-1.7B and Qwen3-4B decoder models using datasets predominantly composed of Turkish legal and official texts. Prior to full-scale CPT training, we conduct ablation studies to determine optimal initialization strategies and dataset sample ratios. These studies evaluate various initialization approaches and data mixing strategies, identifying configurations that balance domain adaptation with preservation of general language capabilities. The CPT process employs a four-phase curriculum learning strategy specifically designed to account for Turkish linguistic complexity, progressively transitioning from general-purpose texts to domain-specific legal content. Experimental results show that Turkish-focused CPT provides meaningful improvements in legal terminology usage, long-context reasoning, and normative text generation. We also conduct comparative analysis of learning dynamics and performance gains across different model scales, providing insights into CPT effectiveness under varying model capacity conditions.

We curate a large-scale dataset combining Turkish legal and general text sources, processed using SemHash-based semantic deduplication~\cite{fineweb2024,semhash2024} and FineWeb quality filtering (see Section~\ref{sec:methodology} for dataset composition details). The project leverages MareNostrum 5 supercomputing infrastructure (see Section~\ref{sec:methodology} for hardware details).

\section{Related Work}

\label{sec:related}

\subsection{Legal Natural Language Processing}

Legal NLP has emerged as a critical research area, driven by the need for intelligent legal information systems. Prior work has explored various tasks including legal document classification, case retrieval, and document summarization \cite{butler2025massivelegalembeddingbenchmark,legalai2020}. Recent work has introduced comprehensive benchmarks for evaluating legal text embedding models, such as the Massive Legal Embedding Benchmark (MLEB) \cite{butler2025massivelegalembeddingbenchmark}, developed by Isaacus, which provides evaluation across multiple document types, jurisdictions, and legal tasks. LEGAL-BERT \cite{legalbert2020} demonstrated the effectiveness of domain-specific pre-training for English legal text. For Turkish legal text, existing approaches have primarily relied on traditional machine learning and transformer-based classification methods \cite{turkishbert2024,vbart2024}. Recent work \cite{AITEAM2025124} explored domain-specific fine-tuning of small language models (Llama 3.1 8B) for Turkish legal sub-domains using Parameter-Efficient Fine-Tuning (PEFT) techniques. However, fine-tuning alone proved insufficient for achieving the depth of domain adaptation required for comprehensive legal language understanding, particularly for morphologically rich languages like Turkish where specialized terminology and complex syntactic structures demand more extensive pre-training. This limitation motivated the adoption of continual pre-training approaches, which enable deeper domain adaptation through large-scale exposure to legal corpora while preserving general language capabilities. When benchmarked on MLEB, our Mecellem models demonstrate competitive performance, highlighting the effectiveness of domain-specific pre-training for legal embedding tasks.

\subsection{CPT and Domain Adaptation}

Domain adaptation aims to align general-purpose large language models with the linguistic and conceptual characteristics of a specific domain. In highly formal domains such as law, and for morphologically rich languages like Turkish, standard fine-tuning or training from scratch approaches are often insufficient or computationally inefficient.

In this work, a CPT approach is adopted rather than full pre-training, continuing the training of pre-trained Qwen3 models on large-scale, Turkish-dominant legal datasets. This approach allows the model to preserve general language knowledge while acquiring deeper understanding of Turkish legal terminology, long-context reasoning, and domain-specific linguistic structures, providing more stable and effective domain adaptation. Domain-adaptive pre-training has proven effective for specialized applications \cite{ulmfit2018}.

\subsection{Modern Bidirectional Encoders}

ModernBERT \cite{modernbert2025} introduces an architectural framework designed for pretraining masked language models optimized for downstream tasks such as Named Entity Recognition (NER), semantic embedding generation, and classification. The architecture extends the context window to 8,192 tokens for both the base and large configurations, a substantial increase over the 512-token constraint of the original BERT. This scalability is facilitated by the integration of Rotary Positional Embeddings (RoPE) for long-range dependency modeling, alongside efficient attention implementations, specifically alternating sliding window (local) and global attention layers.

\subsection{Decoder-to-Encoder Conversion for Embeddings}

The Qwen3 Embedding series \cite{qwen3embedding} demonstrates that autoregressive decoder models can be effectively converted into embedding models through multi-stage training pipelines combining large-scale unsupervised pre-training (150M synthetic samples) with supervised fine-tuning on high-quality datasets (7M labeled + 12M synthetic samples). Their approach achieves state-of-the-art results through: (1) synthetic data generation using LLM capabilities for multilingual, multi-domain query-document pairs; (2) progressive training from weakly supervised pre-training to supervised fine-tuning; (3) model merging strategies across checkpoints. Replicating these results requires substantial infrastructure for data synthesis and extensive training resources.

\subsection{Catastrophic Forgetting in Continual Learning}

A critical challenge in CPT is catastrophic forgetting---the tendency of neural networks to overwrite previously learned knowledge when adapting to new tasks \cite{ewc2017}. Several mitigation approaches have been proposed: Elastic Weight Consolidation (EWC) \cite{ewc2017} identifies and protects important parameters; replay-based methods revisit examples from prior tasks; and orthogonal subspace methods \cite{nayak2025sculpting} constrain new learning to parameter subspaces orthogonal to prior knowledge.

\subsection{Curriculum Learning}

Curriculum learning organizes training data in a progressive manner, presenting simpler examples before more complex ones. In language models, this strategy improves training stability and convergence for long and structurally complex texts \cite{catastrophicforgetting2023,climb2025,weborganizer2025}. In domain adaptation scenarios such as legal language modeling, curriculum learning supports a gradual transition from basic terminology to intensive, normative, and long-context documents. Additionally, prior work has shown that the order of knowledge acquisition plays a critical role in reducing catastrophic forgetting (where newly learned information overwrites previous knowledge), a common problem in continual learning contexts.

\section{Methodology}

\label{sec:methodology}

\subsection{System Overview}

All experiments were conducted on the MareNostrum 5 ACC partition at Barcelona Supercomputing Center. Each ACC node is equipped with 4$\times$ NVIDIA Hopper H100 64GB GPUs, 80 CPU cores, 512GB DDR5 memory, and 800 Gb/s InfiniBand interconnect for distributed training. This high-performance computing infrastructure enables efficient large-scale model training with exceptional parallel processing capabilities and low-latency inter-node communication, essential for processing our 112.7 billion token corpus within practical time constraints.

\subsection{Data Preparation}

\subsubsection{Data Sources and Domain-Specific Dataset}

The dataset used in this work consists predominantly of Turkish legal and official texts. Data sources include legislative texts, high court and regional court decisions, academic legal publications, and official gazette content. To enable the model to preserve general language capabilities, limited multilingual and technical content is also included in the dataset.

Training data are aggregated from multiple sources. The legal subset includes decisions from the Court of Cassation (Yarg{\i}tay) with 10.3M sequences and approximately 3.43B tokens, the Council of State (Dan{\i}\c{s}tay) with 151K sequences and approximately 0.11B tokens, and academic theses from Y\"{O}KTEZ with 21.1M sequences and approximately 9.61B tokens after DocsOCR processing. The general Turkish subset is constructed from FineWeb2~\cite{fineweb22025} and CulturaX, contributing 212M sequences and approximately 96.17B tokens.

Following the curriculum learning approach, training is conducted with a four phases data arrangement: (i) general and relatively simple texts, (ii) domain-specific legal content, (iii) long, normative, and conceptually intensive texts, and (iv) extended domain-specific refinement with mixed complexity.

\subsubsection{Data Extraction}

Constructing a large-scale legal and academic corpus from archival sources requires robust text extraction from scanned documents with heterogeneous quality and formatting. Our initial approach employed Tesseract, a mature and computationally inexpensive OCR system. However, during preliminary processing, we observed systematic failures across specific document categories. Analysis of OCR errors revealed consistent patterns tied to document complexity:

\begin{itemize}

\item\textit{Formulas}: Mathematical equations and formulas were frequently misrecognized or omitted entirely.

\item\textit{Tables}: Multi-column layouts caused text reordering and cell boundary confusion.

\item\textit{Tables of Contents}: Hierarchical structures with page numbers produced garbled output.

\item\textit{Scattered Text}: Characters separated by variable spacing were incorrectly segmented.

\item\textit{Noise}: Low-quality scans introduced random character insertions.

\end{itemize}

These error patterns are attributable to the fundamental architecture of traditional OCR systems, which rely on character-level recognition without document-level understanding. Vision-Language Models (VLMs) offer a principled alternative by treating OCR as a vision-to-text generation task, leveraging transformer-based architectures to reason over spatial relationships and document structure. This capability motivated our development of a VLM-based extraction pipeline using DotsOCR \cite{dotsocr2025} as the underlying model.

\paragraph{Pipeline Architecture}

The extraction pipeline consists of two primary stages: document rendering and VLM-based text extraction. In the rendering stage, PDF pages are converted to PNG images using PyMuPDF at 300 DPI, with the longest dimension resized to 1024 pixels. This resolution was empirically determined to preserve text legibility while remaining within the input constraints of the vision encoder.

For text extraction, we deploy DotsOCR \cite{dotsocr2025} served via vLLM (v0.11.2) across multiple compute nodes in a data-parallel configuration, using CUDA 12.2 and Flash Attention v3. Each node processes an independent partition of the document collection, with the model configured for a maximum context length of 12,288 tokens and generation limit of 3,072 tokens, sufficient to capture full-page content from dense academic documents. Inference is performed with greedy decoding to ensure deterministic outputs suitable for corpus construction.

\paragraph{Scalability Considerations}

Deploying this pipeline at scale on the MareNostrum 5 supercomputer introduced challenges related to the General Parallel File System (GPFS) architecture. GPFS employs a centralized metadata server (MDS) that becomes a bottleneck when handling millions of small files, as each file operation requires MDS interaction. This overhead is exacerbated in shared HPC environments where concurrent user activity compounds metadata contention.

We addressed these constraints through two complementary strategies. First, we restructured the storage layout by partitioning intermediate files into separate subdirectory hierarchies, reducing inode-level locking contention. Second, we eliminated redundant directory creation calls by pre-allocating directory structures, as repeated existence checks were generating substantial MDS read load.

\paragraph{Inference Optimization}

To improve extraction throughput, we upgraded the attention implementation from Flash Attention v2 to Flash Attention v3. As shown in Table~\ref{tab:flash-attention}, this upgrade yielded a 56\% improvement in observed throughput, increasing processing speed from 7--9 images per second to 11--14 images per second. This optimization was critical for processing tens of millions of pages within practical time constraints.

\begin{table}[htbp]
\caption{Throughput Comparison Between Flash Attention Versions}
\label{tab:flash-attention}
\centering
\small
\resizebox{\columnwidth}{!}{%
\begin{tabular}{@{}lrr@{}}
\toprule
& Flash Attention v2 & Flash Attention v3\\
\midrule
Throughput (images/s) & 7--9 & 11--14\\
Relative Speedup & 1.0$\times$ & 1.56$\times$\\
\bottomrule
\end{tabular}%

}

\end{table}

Over the course of the extraction process, the pipeline processed 22.8 billion prompt tokens and generated 15.9 billion output tokens across 100 compute nodes.

\paragraph{Output Completeness Verification}

\begin{figure}[htbp]

\centering

\includegraphics[width=0.95\columnwidth,keepaspectratio]{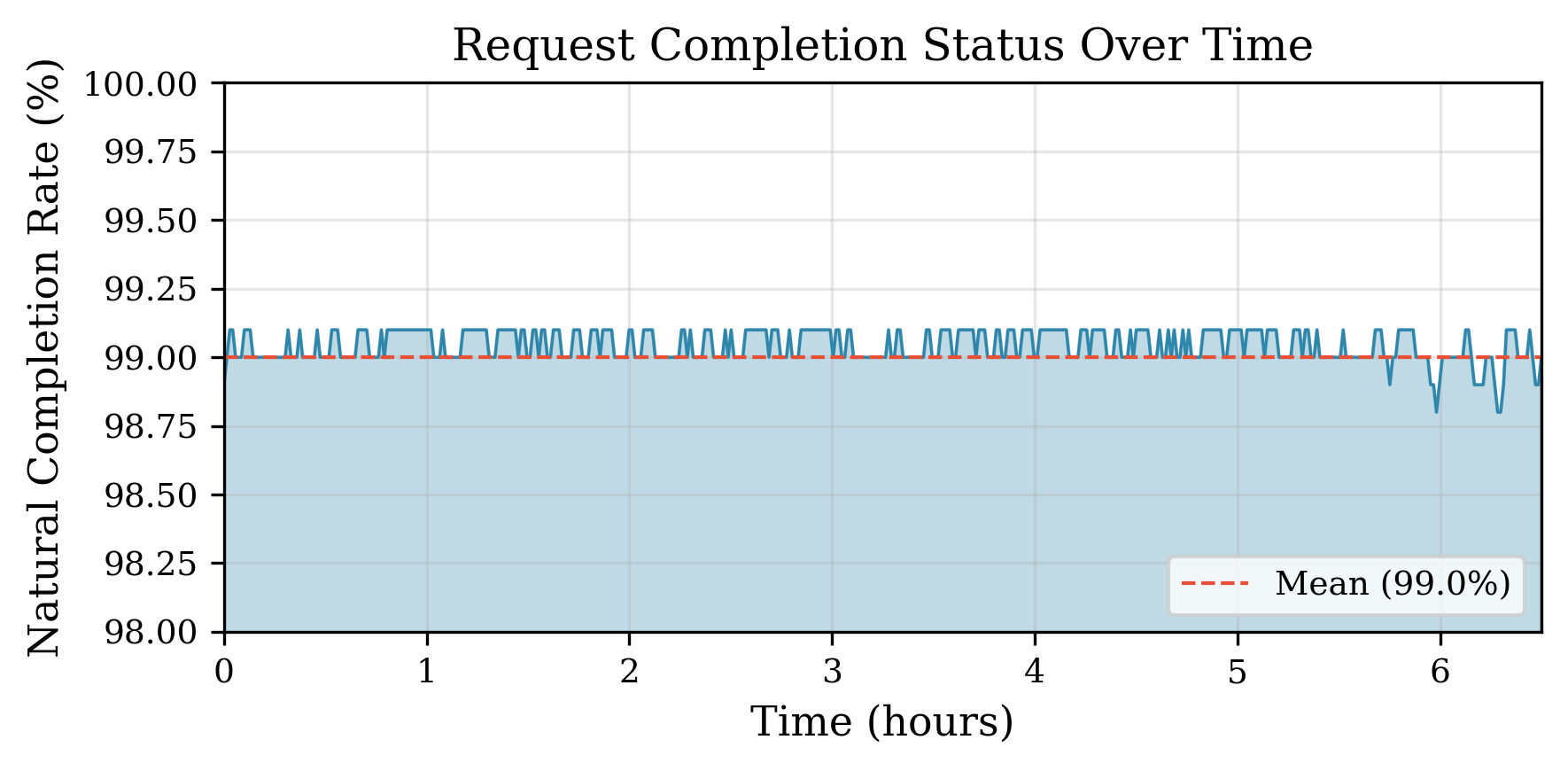}

\caption{Natural completion rate over a 6.5-hour extraction run.}

\label{fig:completion-timeseries}

\end{figure}

To ensure that the extraction pipeline captures complete document content without truncation, we monitored request completion status throughout the inference process. Each request terminates with one of three status codes: natural completion upon generating an end-of-sequence token, length truncation when reaching the maximum token limit, or abort due to timeout or system errors. Figure~\ref{fig:completion-timeseries} presents the distribution of completion status over a representative 6.5-hour extraction run.

As shown in Figure~\ref{fig:completion-timeseries}, the natural completion rate remained stable at 99.0\% (±0.1\%) throughout the extraction process. Only 1.0\% of requests reached the 3,072 token generation limit, indicating consistent pipeline behavior. This high completion rate shows that the configured output length is sufficient for the vast majority of document pages, minimizing information loss due to truncation. The stability of the completion rate across the full runtime suggests that document complexity did not vary systematically in ways that would affect extraction quality.

\subsubsection{Data Preprocessing Pipeline}

\paragraph{Cleaning and Normalization:} All datasets underwent a multi-stage cleaning pipeline to remove duplicates, malformed records, and low-quality content. The pipeline included (i) discarding samples with missing or invalid fields, (ii) intra-batch deduplication to eliminate repeated entries, (iii) filtering out overly short texts that lack sufficient linguistic signal, (iv) excluding excessively long documents that are likely noisy or out-of-distribution and may hinder efficient processing, and (v) removing HTML-derived samples dominated by tabular structure or image references (e.g., pages consisting primarily of tables or image links) rather than natural language. This process ensures that the final corpus is predominantly textual and suitable for language model training.

\paragraph{Filtering:}

A multi-stage filtering pipeline was employed to improve corpus quality and safety while preserving topical and stylistic diversity. The pipeline comprises (i) URL-based safety filtering for web-derived content, (ii) document-level language identification to enforce linguistic consistency, and (iii) content-quality filtering using tuned heuristic indicators.

\textbf{URL Filtering:} We removed records whose URLs contained a curated set of high-frequency inappropriate keywords (e.g., adult, explicit, or gambling-related terms). To reduce false positives, we applied an allowlist of known legitimate sources that were exempt from filtering.

\textbf{Language Filtering:} We ensured linguistic consistency via document-level language identification using The GlotLID classifier~\cite{glotlid2023}. For each document, we predicted the primary language top-$k$ with 0.40 confidence scores and retained only samples whose top prediction matched the target languages, and all other documents or low-confidence cases were filtered out.

\textbf{Quality Filtering:} We applied the FineWeb Quality Filter~\cite{fineweb2024} with tuned thresholds to remove low-quality documents based on indicators such as punctuation patterns, line-length statistics, character duplication, and newline/list-like structure. We set the \textbf{short\_line}, \textbf{char\_duplicates}, \textbf{new\_line} thresholds respectively to \textbf{0.67}, \textbf{0.03}, \textbf{0.4} keeping other parameters at default values. The thresholds were chosen via a small grid search around defaults on a balanced test set built with 5-fold sampling over categories (up to 100 examples per category; random\_state $=42$), selecting the configuration that balanced drop rate and false positives.\footnote{\href{https://github.com/huggingface/datatrove}{github.com/huggingface/datatrove}}

\begin{table}[H]
\caption{The Dropping Ratios of tested Configurations}
\centering
\small
\resizebox{\linewidth}{!}{%
\begin{tabular}{@{}ccccc@{}}
\toprule
short\_line (\%) & char\_duplicates (\%) & char\_dup (\%) & new\_line (\%) & Overall (Drop ratio\%)\\
\midrule
0.1000 & 0.0100 & 0.0018 & 0.3000 & 0.0080\\
0.2500 & 0.1300 & 0.0018 & 0.4500 & 0.0042\\
0.6700 & 0.2500 & 0.0019 & 0.6000 & 0.0042\\
0.7800 & 0.3700 & 0.0017 & 0.7500 & 0.0075\\
0.9000 & 0.5000 & 0.0061 & 0.9000 & 0.0044\\
\bottomrule
\end{tabular}%

}

\end{table}

\textbf{Turkish Quality Filtering:} To quantify Turkish-specific linguistic richness, we compute two morphology-driven indicators from \emph{Zemberek}-based morphological analyses \cite{Akin2007Zemberek}: \emph{suffix entropy} and \emph{lemma diversity}. After tokenization and morphological analysis, we focus on noun and verb tokens and extract their nominal case tags, then estimate the empirical distribution over observed case categories. Suffix entropy is defined as the Shannon entropy~\cite{Shannon1948} of this case-tag distribution:

\begin{equation}
H_{\text{suffix}} = -\sum_{i=1}^{n} p_i\log p_i,
\end{equation}

where $p_i$ denotes the relative frequency of case tag $i$ among noun tokens and $n$ is the number of distinct case tags observed.

Higher values indicate more balanced and varied case usage, which is typical of well-formed, sentence-like prose, whereas lower values reflect dominance of a small number of cases, often associated with templated or repetitive content.

For comparability across texts with different numbers of observed tags, we also report the normalized entropy which lies in $[0,1]$:

\begin{equation}
H_{\text{suffix}}^{\text{norm}} =\frac{H_{\text{suffix}}}{\log(n)}
\end{equation}

Lemma diversity captures lexical variety after morphological normalization. For each token with a valid analysis, we extract its lemma and compute the ratio of unique lemmas to the total number of analyzed tokens:

\begin{equation}
\text{LD} =\frac{|\{l_1, l_2,\ldots, l_k\}|}{N},
\end{equation}

where $\{l_1, l_2,\ldots, l_k\}$ is the set of distinct lemmas and $N$ is the number of tokens contributing a lemma.

Values closer to 0 indicate strong repetition (e.g., boilerplate or duplicated templates), while higher values indicate broader vocabulary after normalization; unusually high values can also flag noisy or out-of-domain text.

Together, suffix entropy and lemma diversity provide complementary signals of morphological variation and lexical richness for Turkish corpus quality assessment.

\begin{figure}[H]

\centering

\includegraphics[width=0.98\columnwidth]{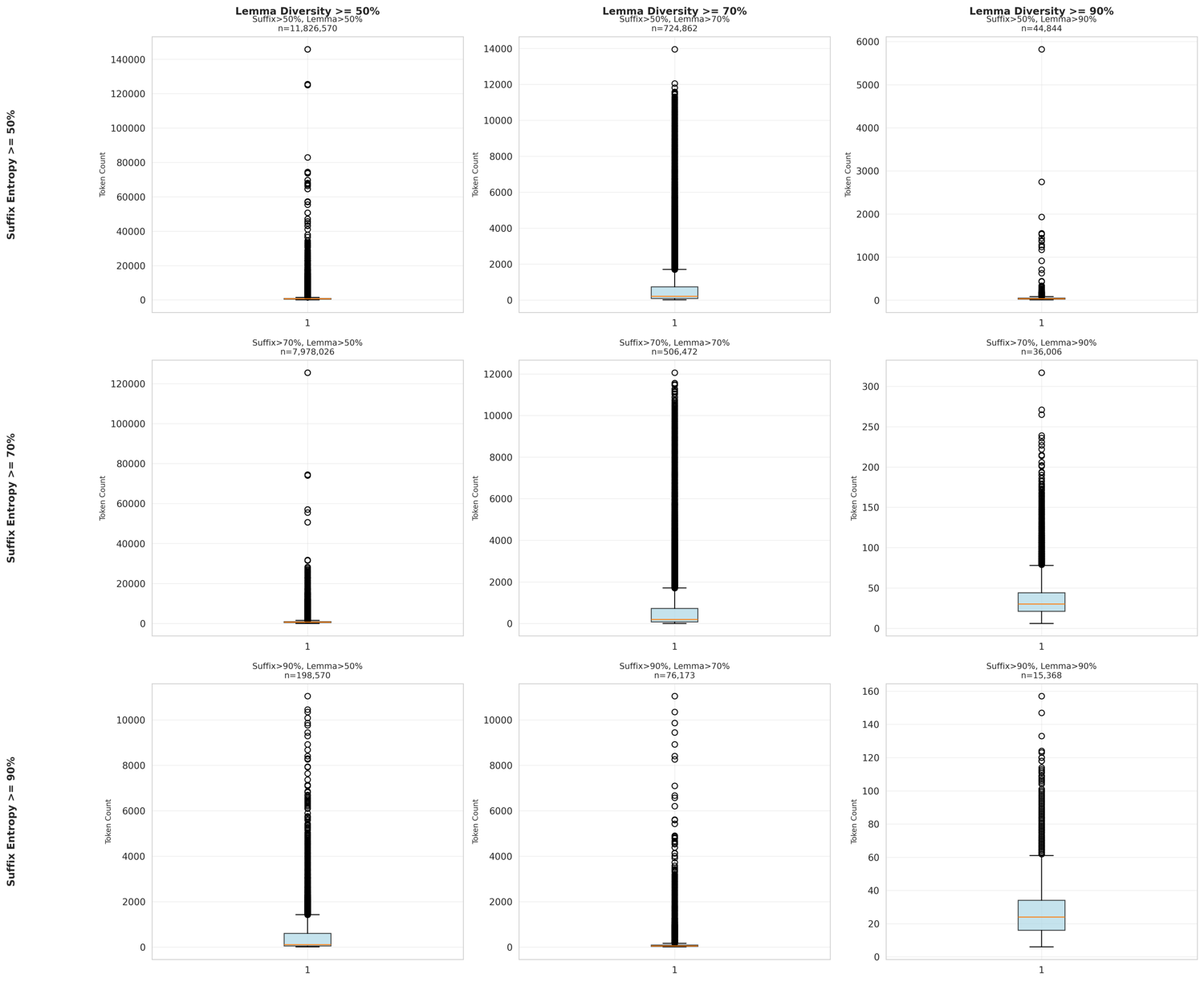}

\caption{Token Count Distribution Analysis Across All Threshold Combinations.}

\label{fig:box_plots_all_thresholds}

\end{figure}

Figure~\ref{fig:box_plots_all_thresholds} shows comprehensive token count distribution analysis across all threshold combinations, comparing distributions showing impact of Suffix Entropy (\%) and Lemma Diversity (\%) filters. Each subplot shows n(sample count), mean and median token counts for surviving data.

\subsubsection{Threshold selection for Turkish morphology filters}

To set filtering thresholds for \emph{suffix entropy} and \emph{lemma diversity}, we performed a coarse-to-fine sweep and tracked both data retention and token-count impact. First, we computed token counts for each sample using the Qwen1.7B tokenizer and evaluated a grid of candidate thresholds\(\{50,70,90\}\) for both metrics. For every\((\tau_{\text{suffix}},\tau_{\text{lemma}})\) pair, we filtered the corpus with \(\texttt{suffix\_entropy}>\tau_{\text{suffix}}\) and\(\texttt{lemma\_diversity}>\tau_{\text{lemma}}\), and recorded the number of surviving samples, total tokens, and token-length distribution statistics (min/max, mean/median, quartiles). As shown in Figure~\ref{fig:box_plots_all_thresholds}, the analysis reveals the impact of different threshold combinations on token count distributions. Table~\ref{tab:turkish_quality_filter} (see Appendix~\ref{app:turkish_quality_filter}) summarizes the main trade-off at fixed lemma diversity.

\begin{figure}[htbp]

\centering

\includegraphics[width=0.98\columnwidth]{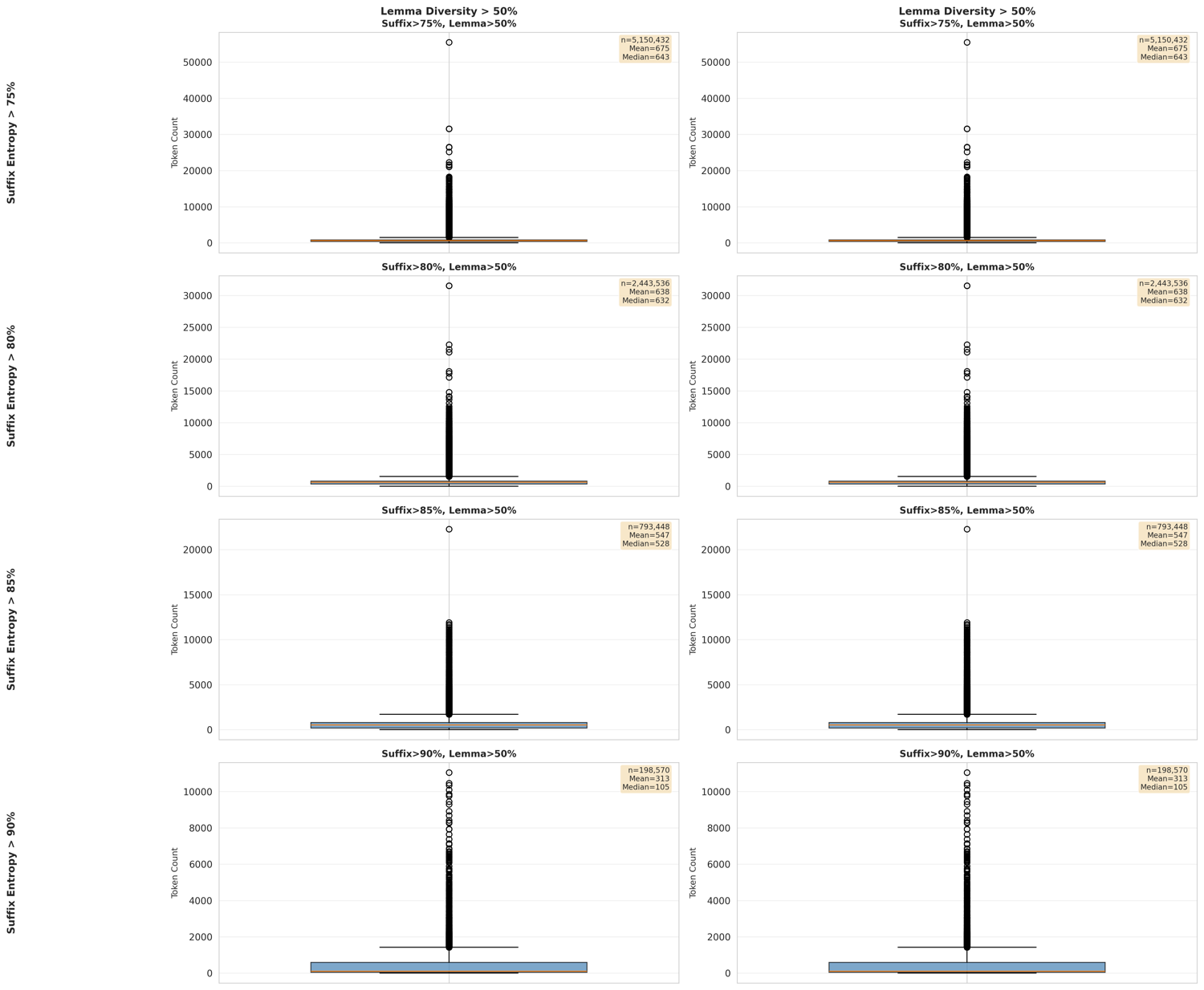}

\caption{Token Count Distribution Across Suffix Entropy Thresholds (Lemma Diversity $\ge$ 50\%)}

\label{fig:only_50s}

\end{figure}

Increasing the lemma-diversity threshold beyond  \textbf{50\%} caused a sharp drop in both remaining documents and total tokens (e.g., from 8.7M samples at 50--50 to 1.6M at 50--65), indicating an overly aggressive constraint that would remove substantial in-domain content. We therefore fixed the lemma-diversity threshold at \textbf{50\%} and refined the suffix-entropy threshold within the transition region between 70\% and 90\%. This second sweep\(\{75,80,85,90\}\) revealed a clear trade-off: while\(>70\%\) retained\(\sim 5.51\)B tokens,\(>90\%\) collapsed the corpus to\(\sim 0.062\)B tokens. Figure~\ref{fig:only_50s} visualizes the token count distribution across suffix entropy thresholds at fixed lemma diversity of 50\%. We selected\(\tau_{\text{suffix}}=\textbf{75\%}\) (i.e., 0.75 in normalized form) as a balanced operating point. This threshold enforces higher morphological variation while preserving a substantial token budget (\(\sim 3.48\)B tokens) and maintaining a stable token-length distribution in the surviving set.

\paragraph{Deduplication:}

To reduce redundancy and improve training efficiency, we apply a multi-stage deduplication pipeline that removes both exact copies and near-duplicate documents. The pipeline processes documents in stages, progressing from low-cost, high-precision exact matching to semantic similarity-based removal, retaining a single representative per duplicate cluster.

\textbf{Exact Match Deduplication:} We perform global deduplication using document text hashes to identify and remove all exact copies. This initial stage identifies approximately 5\% of the corpus as duplicates, reducing redundant training data.

\textbf{Semantic Deduplication:} For near-duplicate detection, we employ SemHash~\cite{semhash2024}, an embedding-based approach using dense embeddings with a similarity threshold of 0.75. This method effectively identifies documents that differ only in headers, footers, or minor formatting variations across multiple domains, significantly reducing redundant content while preserving semantic diversity essential for robust model training. Additionally, we apply URL-based content filtering to exclude samples whose URL fields match inappropriate keyword patterns (e.g., adult, explicit, or gambling-related terms) through keyword-driven string matching. To minimize false positives, we maintain an allowlist of common legitimate news domains and terms, preserving samples that match these exceptions. This semantic deduplication step improves dataset safety and topical relevance before downstream processing.\footnote{\href{https://github.com/MinishLab/semhash}{github.com/MinishLab/semhash}}

\paragraph{Decontamination:}

To mitigate false positives in our filtering pipeline, we developed task-specific heuristics that incorporate domain-specific knowledge and adaptive threshold mechanisms. First, we ignore matches of task-irrelevant text chunks, such as common generic phrases, where irrelevance is determined per task through manual inspection. Second, for specialized domains like mathematics, we apply stricter filtering criteria to ensure higher precision. These heuristics help maintain dataset quality while reducing over-filtering of legitimate content. Additional details on the decontamination module implementation, including profanity filtering, content quality assessment, and sensitive information anonymization, are provided in Appendix~\ref{app:decontamination}.

\textbf{Token and Document Statistics}

Training employed three different datasets containing approximately 3.7 billion, 57 billion, and 165 billion tokens respectively. This gradual structure supports the model's adaptation to increasing data complexity while steadily developing Turkish language proficiency and domain-specific contextual knowledge.

Table~\ref{tab:cpt_dataset_phases} summarizes the dataset statistics for Qwen3-1.7B four-phase curriculum learning and Qwen3-4B single-phase training.

\begin{table}[htbp]
\caption{CPT Training Dataset Statistics by Phase}
\label{tab:cpt_dataset_phases}
\centering
\small
\begin{tabular}{@{}lrr@{}}
\toprule
Phase & Model & Total Tokens\\
\midrule
Phase 1 & Qwen3-1.7B & 3,678,052,050\\
Phase 2 & Qwen3-1.7B & 57,045,791,446\\
Phase 3 & Qwen3-1.7B & 165,110,698,942\\
Phase 4 & Qwen3-1.7B & 24,905,934,016\\
Single Phase & Qwen3-4B & 270,791,712,595\\
\bottomrule
\end{tabular}

\end{table}

\begin{figure}[H]

\centering

\includegraphics[width=0.60\columnwidth]{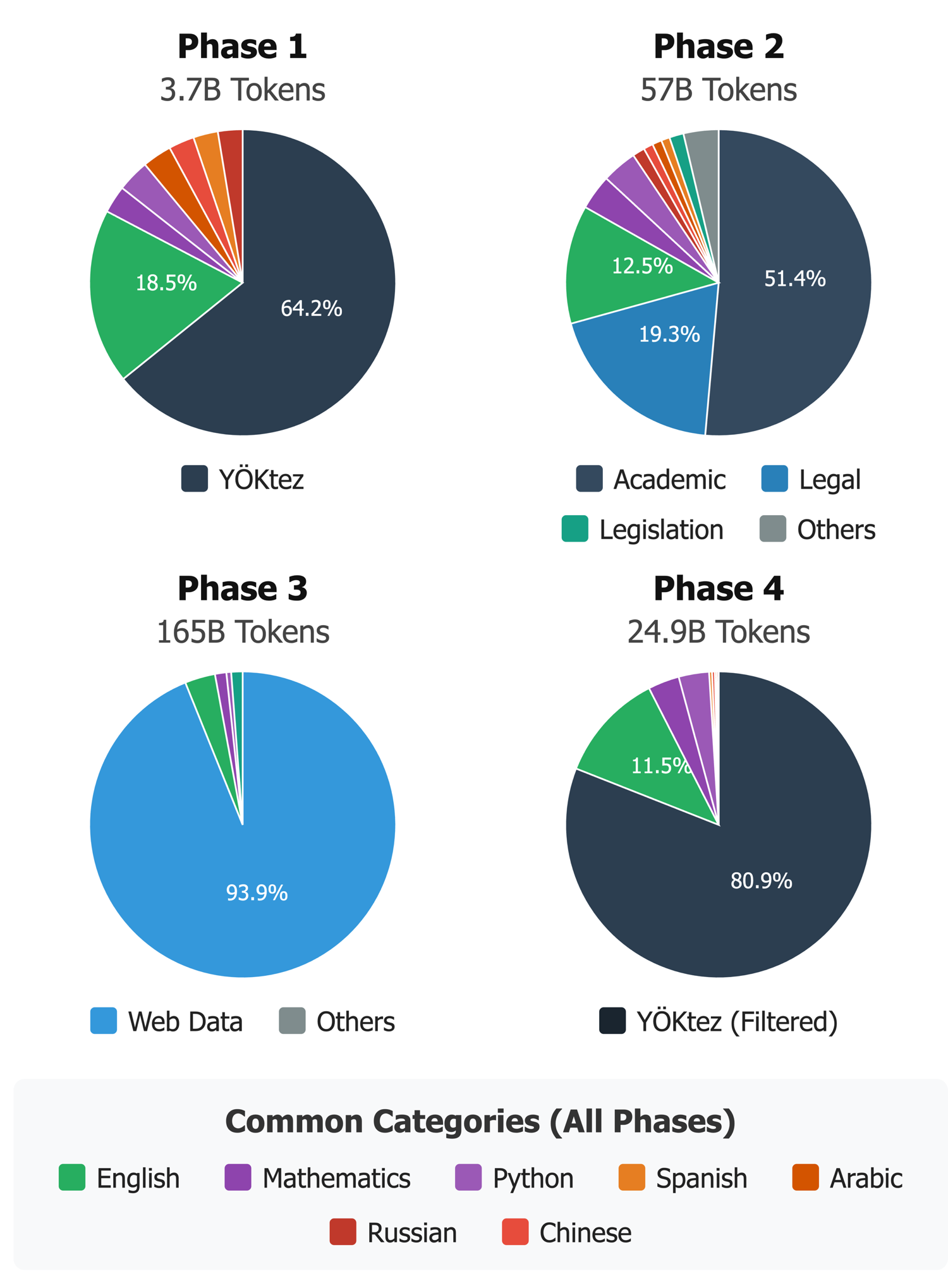}

\caption{Qwen3-1.7B CPT Dataset Distribution across Four Phases.}

\label{fig:qwen3-1.7_dataset}

\end{figure}

Figure~\ref{fig:qwen3-1.7_dataset} illustrates the dataset composition across the four training phases for Qwen3-1.7B, showing the progression from academic-focused data (Phase 1) through legal and web data integration (Phases 2-3) to refined domain-specific content (Phase 4). Common categories across all phases include English, Mathematics, Python, and multilingual data (Spanish, Arabic, Russian, Chinese), while phase-specific categories reflect the curriculum learning strategy.

In the first \textbf{phase} we applied semantic deduplication to remove near-duplicate records beyond exact string matches, followed by FineWeb quality filtering to discard structurally noisy or template-like content. Since theses are often split across multiple pages, we then merged consecutive pages belonging to the same document (preserving page order) to reconstruct coherent long-form samples.

In the second \textbf{phase} we used a lighter pipeline focused on quality control, applying FineWeb quality filtering to remove low-quality documents while preserving broad topical diversity and general-domain signal.

In \textbf{phase 3} we employed the full safety and normalization pipeline: (i) cleaning and normalization to discard malformed entries, extremely short/long samples, and HTML artifacts dominated by tables or image links; (ii) document-level language identification, retaining only target-language samples under a confidence-based criterion; (iii) semantic deduplication to remove near-duplicates at scale; (iv) FineWeb quality filtering for structural and stylistic quality and (v) for the web subset, URL-based filtering using a curated denylist of inappropriate keyword patterns with an allowlist to reduce false positives. Finally, in \textbf{phase 4} we expanded the academic portion used in earlier phases and re-applied the same thesis-specific procedure as in \textbf{phase 3} to maintain consistency while increasing academic coverage.

\begin{figure}[H]

\centering

\includegraphics[width=0.6\columnwidth]{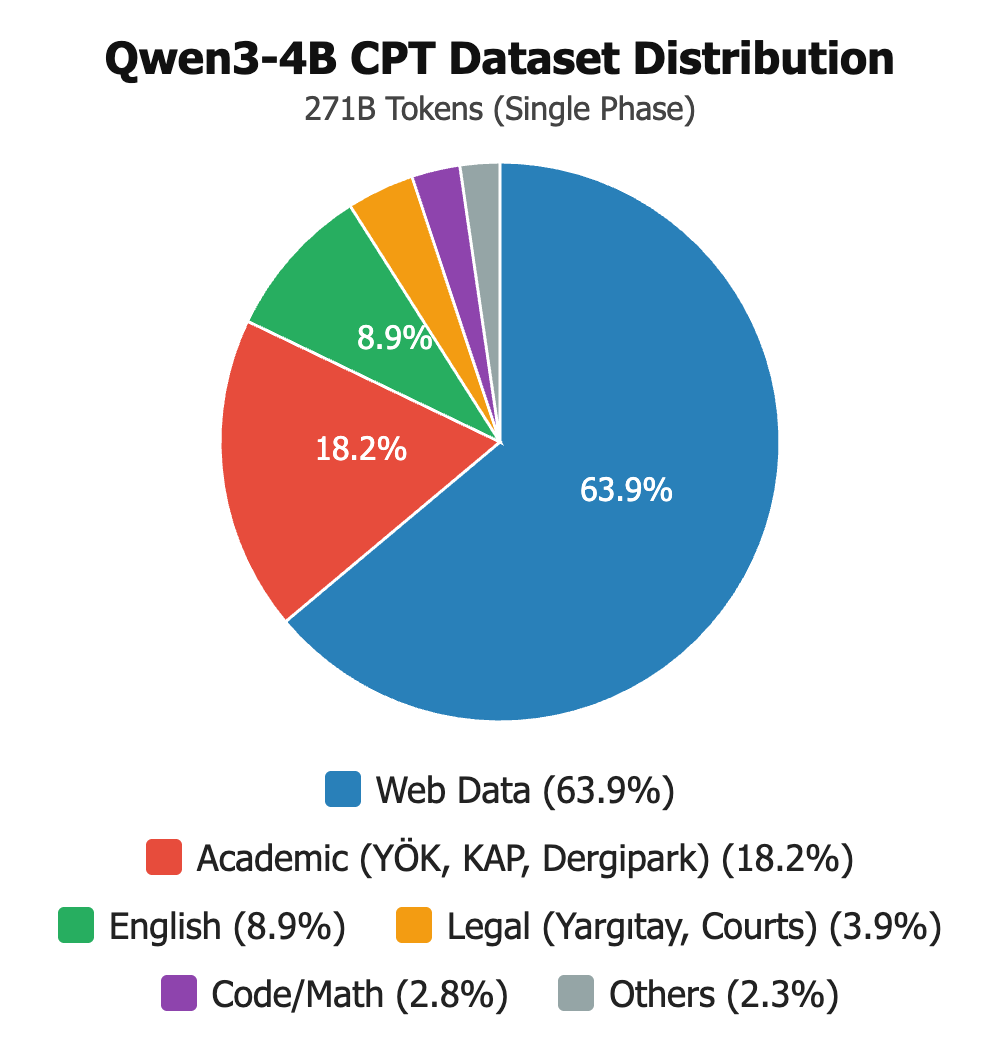}

\caption{Qwen3-4B CPT Dataset Distribution Single Phase.}

\label{fig:qwen4b_dataset}

\end{figure}

As shown in Figure~\ref{fig:qwen4b_dataset}, for Qwen3-4B, we combined academic theses (expanded from earlier phases) with legal, legislation, web, and other sources to obtain the final training mixture.

For both Qwen3-1.7B and Qwen3-4B CPT, supplementary training data were sourced from publicly available datasets: mathematics data from Nemotron-CC-Math~\cite{nemotronccmath2025} and OpenWebMath~\cite{fineweb22025,openwebmath2025}, multilingual content (Spanish, Arabic, Russian, Chinese) from FineWeb-2, and Python code from StarCoderData~\cite{starcoder2023} and The Stack Smol~\cite{starcoder2023}.




\subsection{Encoder Model: Pre-training from Scratch}

\subsubsection{Model Details}

The Mecellem Encoder project develops ModernBERT-based bidirectional encoder models pre-trained entirely from scratch on Turkish-dominant corpora. Unlike domain-adaptive approaches that continue training from existing checkpoints, our models are initialized randomly and trained on a carefully curated dataset totaling approximately 112.7 billion tokens combining Turkish legal text with general web data. The resulting models serve as foundation models for downstream embedding tasks, achieving state-of-the-art performance on Turkish legal retrieval benchmarks.

\subsubsection{Architecture}

ModernBERT incorporates modern architectural advances for bidirectional encoders, including alternating local and global attention to efficiently handle long contexts \cite{modernbert2025}. The architecture utilizes pre-layer normalization with RMSNorm and GeGLU activations (Gated Linear Units with GELU) in the MLP layers, while removing all bias terms to improve hardware throughput. To support sequences up to 8,192 tokens, it implements rotary positional embeddings (RoPE) across all layers, utilizing sliding window attention in its local layers (two out of every three layers) to reduce computational complexity. Table~\ref{tab:encoder_arch} summarizes the architecture hyperparameters.

\begin{table}[htbp]
\caption{ModernBERT Architecture Configuration}
\label{tab:encoder_arch}
\centering
\footnotesize
\begin{tabular}{@{}lrr@{}}
\toprule
Parameter & Base & Large\\
\midrule
Seq. Length & 1,024 & 2,048\\
Layers & 22 & 28\\
Hidden Size & 768 & 1,024\\
FFN Size & 1,152 & 2,624\\
Attn. Heads & 12 & 16\\
Head Dim. & 64 & 64\\
Activation & GLU+GELU & GLU+GELU\\
Pos. Enc. & RoPE & RoPE\\
RoPE $\theta$ & 10,000 & 20,000\\
Norm & RMSNorm & RMSNorm\\
Window Size & 128 & 128\\
Vocab Size & 59,008 & 59,008\\
MLM Prob. & 0.3 & 0.3\\
\midrule
\textbf{Total Params} &\textbf{155M} &\textbf{403M}\\
\bottomrule
\end{tabular}

\end{table}

\subsubsection{Tokenizer Design and Analysis}

Alongside the model architecture, a custom tokenizer was developed tailored to the distribution of Turkish web data and legal documents. To achieve optimal fertility and broad coverage of legal terminology, the tokenizer was trained on a large-scale dataset combining dense legal manuscripts with Common Crawl web data. The Byte Pair Encoding (BPE) algorithm with a regex-based pre-tokenization step was employed to ensure maximum efficiency. Specifically, the Llama pre-tokenization pattern was implemented, which demonstrated the best performance for the structural requirements of the Turkish language.

Tokenizer design directly impacts model performance on morphologically rich languages. Turkish is agglutinative—a single word can encode subject, object, tense, aspect, and mood through suffixation. Tokenizers trained on English often split Turkish words at arbitrary points, breaking morphological structure and hurting downstream performance~\cite{bayram2025tokenizationstandardslinguisticintegrity}.

\paragraph{Tokenizer Benchmark Methodology}

We evaluate tokenizer quality using metrics derived from the methodology proposed by Bayram et al.~\cite{bayram2025tokenizationstandardslinguisticintegrity} and the ITU Turkish NLP Web Service~\cite{eryigit2014itu}. Two primary metrics are computed:

\textbf{Turkish Token Count}: The number of vocabulary entries corresponding to valid Turkish words or morphemes, validated using morphological analysis via the ITU Turkish NLP pipeline~\cite{eryigit2014itu}.

\textbf{Token Purity}: The count of tokens representing complete morphological units rather than arbitrary subword fragments. A token is considered pure if it corresponds to a meaningful linguistic unit (root, suffix, or complete word) without semantic loss.

To illustrate, consider the word \textit{çalışıyorlar} (they are working). A morphologically-aware tokenizer produces \texttt{[çalış, ıyor, lar]}—three pure tokens representing the verb root, present continuous suffix, and plural marker respectively. A suboptimal tokenizer might produce \texttt{[ça, lı, şı, yor, lar]}, yielding only one pure token (\texttt{lar}), as the remaining fragments carry no morphological meaning.

\paragraph{Tokenizer Selection Results}

\begin{figure}[!th]

\centering

\includegraphics[width=0.75\linewidth]{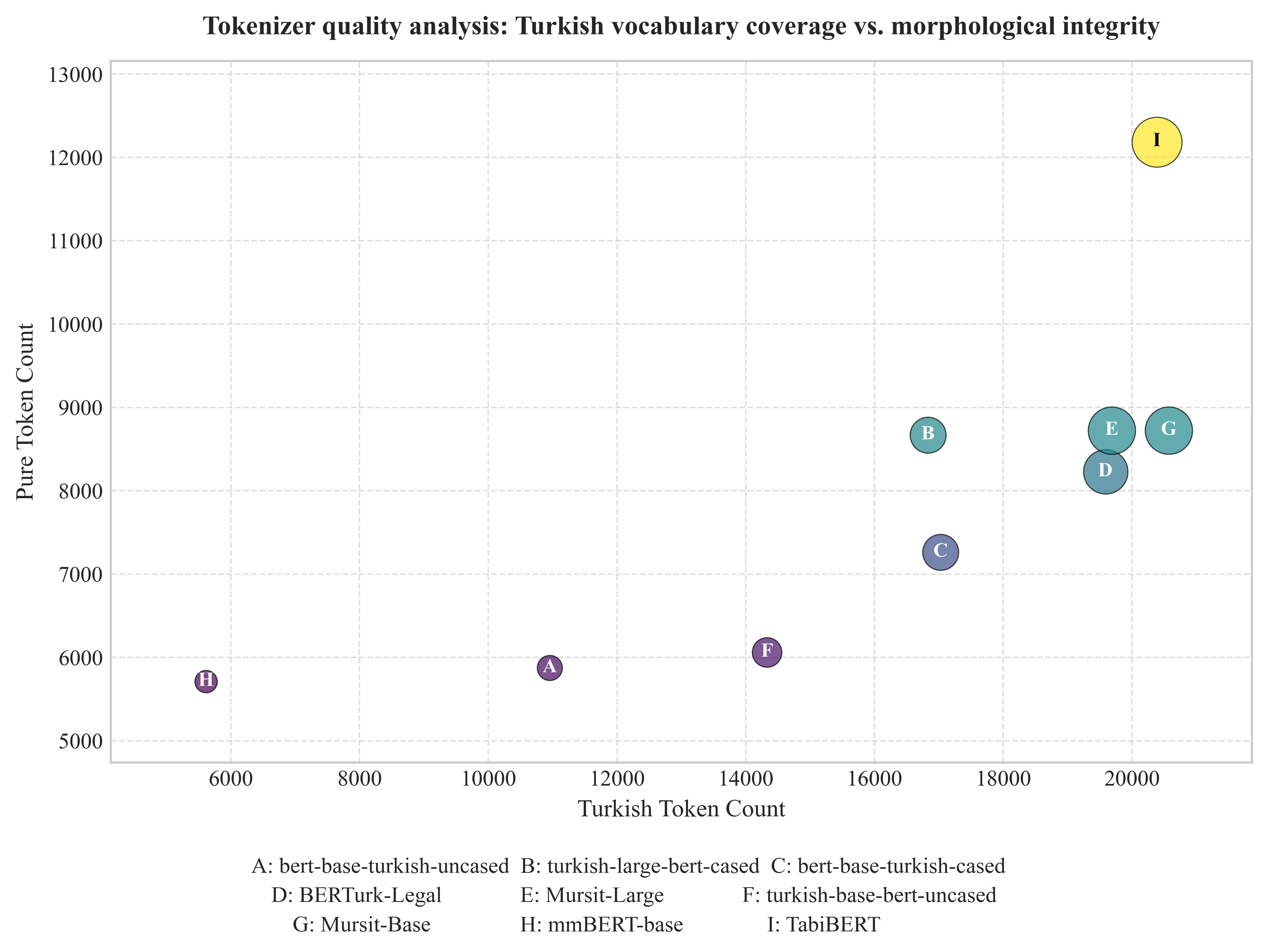}

\caption{Tokenizer quality analysis: Turkish vocabulary coverage vs. morphological integrity. Turkish Token Count (x-axis) measures vocabulary coverage of Turkish words and morphemes, while Pure Token Count (y-axis) represents morphological integrity—the count of tokens corresponding to complete morphological units. Bubble size indicates Unique Token Count, and bubble color (viridis colormap) represents Pure Token Count, where darker purple indicates lower values and brighter yellow indicates higher values. Only Turkish pre-trained MLM models (and one multilingual model, mmBERT-base) from Table~\ref{tab:comprehensive_embedding_results} are evaluated.}

\label{fig:tokenizer-quality}

\end{figure}

Figure~\ref{fig:tokenizer-quality} shows the relationship between tokenizer characteristics and model performance for Turkish pre-trained MLM models evaluated in Table~\ref{tab:comprehensive_embedding_results}. The bubble plot visualizes three key tokenizer quality metrics: Turkish Token Count (x-axis), Pure Token Count (y-axis), and Unique Token Count (bubble size), with bubble color representing Pure Token Count using a viridis colormap.

The analysis reveals a strong positive correlation between Turkish Token Count and Pure Token Count, indicating that models with broader Turkish vocabulary coverage also achieve better morphological integrity. Turkish pre-trained MLM models consistently demonstrate higher pure token counts (6,000–12,000) and Turkish token coverage (14,000–20,000) compared to the multilingual baseline (mmBERT-base), correlating with improved performance on both general and legal tasks. Turkish pre-trained models score 40–58\% vs. 22–23\% for mmBERT-base on general tasks (roughly 2.5$\times$ improvement), and 17–47\% vs. 2–3\% on legal retrieval (up to 20$\times$ difference). Legal text contains specialized vocabulary and complex nominal constructions that amplify tokenization errors, making morphological integrity critical for domain-specific tasks.

These results confirm that Turkish vocabulary coverage and morphological integrity in the tokenizer are necessary for strong performance, especially on domain-specific tasks. Based on this analysis, we selected the Llama pre-tokenization pattern with BPE, which achieved optimal Turkish token count and purity for our legal domain corpus.

\subsubsection{Pre-training}

\paragraph{Dataset Composition and Corpus Construction}

The pre-training corpus combines Turkish legal and general web data, totaling 112.7 billion tokens. Training data are aggregated from multiple sources: the legal subset includes decisions from the Court of Cassation (Yarg{\i}tay) with 10.3M sequences (3.43B tokens), the Council of State (Dan{\i}\c{s}tay) with 151K sequences (0.11B tokens), and academic theses from Y\"{O}KTEZ with 21.1M sequences (9.61B tokens) after OCR processing using DotsOCR \cite{dotsocr2025}. The general Turkish subset is constructed from FineWeb2~\cite{fineweb22025} and CulturaX, contributing 212M sequences (96.17B tokens).

The preprocessing pipeline prioritizes legal compliance, semantic diversity, and training stability over raw corpus size, employing conservative and fully reproducible transformations. All stages execute in batch-parallel form on the MareNostrum 5 supercomputer using GPU-accelerated cuDF workflows. Structurally invalid records are removed, followed by exact deduplication and strict legal filtering. Documents with unclear licensing were proactively excluded to ensure legal compliance. Web data undergoes additional filtering via GPU-parallel URL-based safety rules. OCR-induced noise, including scattered-text artifacts, is addressed through a recovery-oriented normalization pipeline. To eliminate near-duplicate legal templates, semantic deduplication is applied using dense document embeddings generated by the BGE-M3 encoder~\cite{bgem32024} and filtered with SemHash~\cite{semhash2024} at a cosine similarity threshold of 0.95. The final pipeline reduces the corpus from 132.6M to 72.2M documents while preserving linguistically meaningful content.

\paragraph{Training Dataset Evolution}

The training process was structured into multiple dataset versions, progressing from initial pipeline validation to comprehensive URL-based filtering, the integration of native Turkish legal and academic corpora, and optimized page-packing strategies:

\vspace{0.5em}

\noindent\textbf{Initial training datasets:} $\sim$8--52B tokens. These datasets were used to validate the training pipeline and tune hyperparameters, and consist primarily of Filtered FineWeb and CultureWeb data.

\vspace{0.5em}

\noindent\textbf{URL-filtered web corpus with academic and legal data:} $\sim$101.5B tokens. This corpus aggregates all previously used web data and applies a basic URL-level filtering to remove unwanted sources. It also includes native Turkish academic articles and legal documents collected from Y\"{O}KTEZ, Yarg{\i}tay (Court of Cassation), and Dan{\i}\c{s}tay (Council of State).

\vspace{0.5em}

\noindent\textbf{Extended legal and academic corpus:} $\sim$109.3B tokens. This dataset builds on the same web and legal foundation, augmented with a substantially larger collection of Y\"{O}KTEZ academic theses obtained via OCR.

\vspace{0.5em}

\noindent\textbf{Page-packed and quality-filtered corpus:} $\sim$112.7B tokens. This final corpus further enhances the previous data by applying additional quality and URL filtering to the web sources, and introduces page packing for Y\"{O}KTEZ documents to increase effective sequence length during training.

\paragraph{Training Configuration}

Pre-training was conducted on the MareNostrum 5 ACC supercomputer through an EuroHPC access grant, using the MosaicML Composer framework~\cite{mosaicmlcomposer2021}. Training employs Decoupled StableAdamW optimizer, warmup\_stable\_decay learning rate schedule, and BF16 mixed precision. The Masked Language Modeling (MLM) objective is used as the primary training signal, with a masking probability of 15\% following standard BERT practices. Rather than relying solely on MLM loss minimization, we implement a checkpoint selection strategy that evaluates downstream retrieval performance throughout training, revealing non-linear relationships between MLM loss and embedding quality. Table~\ref{tab:pretraining_hyperparams} summarizes the model-specific hyperparameters.
\vspace{-0.3\baselineskip}
\begin{table}[htbp]
\caption{Pre-training Hyperparameters for ModernBERT Models}
\label{tab:pretraining_hyperparams}
\centering
\small
\begin{tabular}{@{}lrr@{}}
\toprule
Parameter & Base & Large\\
\midrule
Nodes & 16 & 32\\
GPUs & 64 H100 & 128 H100\\
Learning Rate & 5$\times$10$^{-4}$ & 8$\times$10$^{-4}$\\
\bottomrule
\end{tabular}

\end{table}
\vspace{-0.3\baselineskip}
\subsubsection{Post-training for Embeddings}

For converting the pre-trained encoder into an embedding model, we explore multiple contrastive learning techniques on the MS MARCO-TR dataset.

\textbf{Training Dataset:}

\begin{itemize}

\item\textbf{MS MARCO-TR}: Turkish translation of MS MARCO passage ranking dataset, consisting of 920,106 triplets. Mean query length: 8.2 tokens, mean passage length: 69.9 tokens.

\end{itemize}

\textbf{Loss Functions Evaluated:}

\textbf{1. InfoNCE Loss}: Standard contrastive loss treating in-batch negatives as hard negatives:

\begin{equation}
\mathcal{L}_{\text{InfoNCE}} = -\frac{1}{N}\sum_i\log\frac{\exp(\text{sim}(q_i, d_i^+)/\tau)}{\sum_j\exp(\text{sim}(q_i, d_j)/\tau)}
\end{equation}

\textbf{2. Qwen3-Style InfoNCE}: Modified temperature scaling following Qwen3-Embedding approach with adjusted margin.

\textbf{3. GISTEmbed with Cached Guide Model}: Guided In-sample Selection of Training Negatives \cite{gistembed2024} with false-negative filtering. For each query $q_i$ and candidate negative $d_j$:
\vspace{-0.3\baselineskip}
\begin{equation}
\text{filter}(d_j) =\text{sim}_{\text{guide}}(q_i, d_j) <\text{sim}_{\text{guide}}(q_i, d_i^+) -\text{margin}
\end{equation}
\vspace{-0.5\baselineskip}

We evaluate two guide models: google/embeddinggemma-300m~\cite{bgem32024,embeddinggemma2025} (307M parameters, 768-dimensional) and BAAI/bge-m3 (568M parameters, 1024-dimensional). The cached implementation precomputes guide embeddings ($\sim$3.5GB for 920K samples) once before training, avoiding repeated forward passes through the guide model during each training step and thereby reducing computational overhead.

\textbf{Post-training Configuration:}

Through systematic ablation studies, we identify optimal hyperparameters balancing training efficiency with downstream task performance. Token distribution analysis of MS MARCO-TR reveals: queries mean=8.2 tokens (P99=16), documents mean=69.9 tokens (P99=116), with 99.9\% sample coverage at 256 tokens and 0.05-0.07\% token truncation loss. Post-training was conducted using AdamW optimizer (learning rate=2$\times$10$^{-5}$, weight decay=0.01) on 4$\times$H100 GPUs. Table~\ref{tab:posttraining_config} summarizes the hyperparameters used for post-training ModernBERT models.
\vspace{-0.8\baselineskip}
\begin{table}[htbp]
\caption{Post-training Configuration for ModernBERT Models}
\label{tab:posttraining_config}
\centering
\small
\begin{tabular}{@{}lrr@{}}
\toprule
Parameter & Base (155M) & Large (403M)\\
\midrule
Per-GPU Batch & 48 & 48\\
Gradient Accum. & 4 & 12\\
Effective Batch & 768 & 2,304\\
Sequence Length & 1,024 & 2,048\\
Temperature ($\tau$) & 0.05 & 0.05\\
\bottomrule
\end{tabular}
\vspace{-0.5\baselineskip}
\end{table}

\subsubsection{Pre-training Dataset Version Progression:}

Our iterative refinement of the pre-training dataset across multiple versions (v1 through v6) demonstrates the critical importance of data quality and domain coverage for Turkish legal embedding performance. Even minimal token truncation (0.05-0.07\%) results in performance degradation when sequence length is optimized solely for training data efficiency. Models trained with shorter sequences exhibit degradation in legal domain tasks, with regulation retrieval showing -23.6\% relative decrease and case law retrieval showing -11.6\% relative decrease, indicating that long-document understanding is essential for legal domain applications. Detailed results are presented in Section~\ref{sec:experiments}.




\subsection{Decoder Model: CPT}

\subsubsection{Theoretical Motivation for CPT}

CPT is an approach that aims to continue pre-training of large language models on domain-specific and language-specific data while preserving general language capabilities. Unlike classical fine-tuning, CPT updates all model parameters on large-scale data, providing more persistent and deeper domain adaptation. This characteristic is critical for maintaining conceptual consistency in highly formal domains such as law that require strong structural integrity and long-context understanding.

\subsubsection{Model Details}

The Mecellem Decoder project applies CPT to existing decoder-only language models, adapting them for Turkish legal domain understanding. Unlike pre-training from scratch, CPT preserves the general language capabilities of the base model while injecting domain-specific knowledge through extended training on specialized corpora. We select Qwen3-1.7B and Qwen3-4B \cite{qwen2024} as base models due to their strong multilingual capabilities and efficient architecture.

\subsubsection{Pretraining to CPT Transition Strategy}

In this work, CPT is applied directly as continuation training on Qwen3-Base models. The data distribution in the CPT phase is rebalanced to be Turkish-dominant and legal domain-focused. For the Qwen3-1.7B model, the process is conducted in multiple phases using a curriculum learning approach; the Qwen3-4B model is trained with a single-phase, large-scale CPT process.

\subsubsection{Architecture}

The decoder models follow the standard Qwen3 architecture with grouped query attention (GQA), RMSNorm, SwiGLU activations, and rotary positional embeddings. Table~\ref{tab:decoder_arch} summarizes the architecture.

\begin{table}[htbp]
\caption{Qwen3 Decoder Architecture Configuration}
\label{tab:decoder_arch}
\centering
\small
\begin{tabular}{@{}lrr@{}}
\toprule
Parameter & Qwen3-1.7B & Qwen3-4B\\
\midrule
Max Position Embeddings & 40,960 & 40,960\\
Number of Layers & 28 & 36\\
Hidden Size & 2,048 & 2,560\\
FFN Hidden Size & 6,144 & 9,728\\
Number of Heads & 16 & 32\\
Number of KV Heads (GQA) & 8 & 8\\
Activation Function & SwiGLU & SwiGLU\\
Position Encodings & RoPE & RoPE\\
Layer Norm & RMSNorm & RMSNorm\\
\midrule
\textbf{Total Parameters} &\textbf{1.7B} &\textbf{4B}\\
\bottomrule
\end{tabular}

\end{table}

\subsubsection{Addressing Catastrophic Forgetting}

Catastrophic forgetting poses severe practical challenges in legal domain applications, where forgetting general language capabilities while acquiring legal expertise would severely limit model utility. In Mecellem CPT, we employ two complementary strategies to mitigate catastrophic forgetting:

\begin{itemize}

\item\textbf{Curriculum Learning}: The four-phase training structure for Qwen3-1.7B organizes data from general texts to domain-specific legal content to complex normative documents and finally to extended domain-specific refinement. This progressive approach supports gradual domain adaptation, preventing the abrupt distribution shifts that typically cause forgetting.

\item\textbf{Replay Buffer}: To reinforce previously learned knowledge, we incorporate a portion of Y\"{O}KTEZ academic legal data from Phase 1 into Phase 2 training. This data replay strategy ensures that foundational legal terminology and document structures acquired in earlier phases are continuously reinforced during subsequent training stages.

\end{itemize}

These strategies are further supported by conservative learning rate scheduling (maximum 5e-5 with cosine decay) and extended warmup periods, which help stabilize parameter updates throughout the CPT process.

\subsubsection{Training Hyperparameters}

Hyperparameters for CPT training were selected to provide stable and scalable training in distributed environments. All experiments share the following common settings: sequence length of 4,096 tokens, Adam optimizer with cosine learning rate schedule (max: 5e-5, min: 5e-6), weight decay of 0.01, BF16 precision, 10 validation batches per evaluation, NVIDIA NeMo\footnote{\href{https://github.com/NVIDIA/NeMo}{github.com/NVIDIA/NeMo}}~\cite{nemo2024} framework with data-parallel distributed strategy on H100 GPUs at BSC. Phase-specific hyperparameters that vary across training configurations are summarized in Table~\ref{tab:cpt_config}.

\subsubsection{Pre-training (Continual)}

Qwen3 models were continually pre-trained on approximately 225 billion tokens using the MareNostrum 5 ACC supercomputer. The training employed a four-phase curriculum learning strategy designed for progressive domain adaptation, which also serves as a form of catastrophic forgetting mitigation by gradually transitioning the model from general to specialized knowledge. Phase 1 ($\sim$3.7B tokens) focuses on short, general-purpose Turkish texts to adapt the model to Turkish language patterns while maintaining stability, using higher learning rates with extended warmup. Phase 2 ($\sim$57B tokens) introduces legal content with domain-specific terminology including court decisions, legal articles, and regulatory documents, shifting the data mix toward specialized sources. Phase 3 ($\sim$165B tokens) incorporates long, structurally complex normative texts including full court decisions, legislative documents, and academic legal theses, refining the model's understanding of legal reasoning patterns. Phase 4 ($\sim$24.9B tokens) provides extended domain-specific refinement combining mixed complexity documents for final model optimization, consolidating knowledge from previous phases while improving generalization across legal subdomains.

\begin{table}[htbp]
\caption{CPT Training Configuration (Phase-Specific Parameters)}
\label{tab:cpt_config}
\centering
\footnotesize
\begin{tabular}{@{}lrrrr|r@{}}
\toprule
Parameter & P1 & P2 & P3 & P4 & 4B\\
\midrule
Max Steps & 4,170 & 32,260 & 46,803 & 14,122 & 153,508\\
Epochs & 2 & 1 & 1 & 1 & 1\\
Init Ckpt. & Base & P1 & P2 & P3 & Base\\
Warmup & 200 & 1,600 & 2,340 & 706 & 7,675\\
Micro Batch & 2 & 2 & 2 & 2 & 1\\
Global Batch & 400 & 400 & 800 & 400 & 400\\
Val Interval & 100 & 100 & 100 & 100 & 5,000\\
\bottomrule
\end{tabular}

\parbox{\columnwidth}{\scriptsize\textit{P1--P4: Qwen3-1.7B phases; 4B: Qwen3-4B single-phase training.}}

\end{table}

\subsubsection{Hardware and Distributed Training Architecture}

All CPT experiments were conducted on BSC MareNostrum5 supercomputing infrastructure using NVIDIA H100 SXM GPUs. Training utilized NVIDIA NeMo~\cite{megatronlm2020,nemo2024} and Megatron-Core based, data-parallel multi-node and multi-GPU distributed architecture with 4 GPUs per node. The InfiniBand network infrastructure enabled efficient processing of large-scale token flow and ensured high scalability and training stability in long-term CPT training.

\textbf{Hardware Utilization:} We measure training efficiency using Model FLOPs Utilization (MFU), where the theoretical peak for NVIDIA H100 SXM is 989 TFLOP/s for BF16 operations\footnote{\href{https://resources.nvidia.com/en-us-gpu-resources/h100-datasheet-24306}{resources.nvidia.com/en-us-gpu-resources/h100-datasheet-24306}}. Table~\ref{tab:mfu_stats} presents the MFU statistics for both Qwen3-1.7B curriculum training and Qwen3-4B single-phase training. For Qwen3-1.7B, median MFU values ranging from 20.3\% to 20.7\% represent effective hardware utilization after filtering checkpoint and validation overhead. Phase 3 achieved the highest throughput (7.35M tokens/sec) utilizing 400 GPUs. Qwen3-4B training achieved 18.7\% median MFU with 2.57M tokens/sec throughput, demonstrating consistent hardware efficiency across both model scales.

\begin{table}[htbp]
\caption{CPT Hardware Utilization (MFU) Statistics}
\label{tab:mfu_stats}
\centering
\footnotesize
\begin{tabular}{@{}llrrrrr@{}}
\toprule
Model & Phase & Nodes & GPUs & Mean & Median & Tok/s\\
\midrule
\multirow{4}{*}{1.7B} & P1 & 50 & 200 & 19.7\% & 20.7\% & 3.77M\\
& P2 & 50 & 200 & 18.8\% & 20.7\% & 3.59M\\
& P3 & 100 & 400 & 19.2\% & 20.3\% & 7.35M\\
& P4 & 50 & 200 & 17.0\% & 20.6\% & 3.25M\\
\midrule
4B & Single & 100 & 400 & 15.7\% & 18.7\% & 2.57M\\
\bottomrule
\end{tabular}
\vspace{-0.5\baselineskip}
\parbox{\columnwidth}{\scriptsize\textit{1.7B/4B: Qwen3-1.7B/Qwen3-4B; P1--P4: Phase 1--4; Mean/Median: MFU (\%); Tok/s: Tokens per second.}}

\end{table}

\subsubsection{Decoder-to-Encoder Conversion and Post-training for Embeddings}

To adapt the CPT decoder models for dense retrieval tasks, we implement a decoder-to-encoder conversion methodology following recent approaches for converting autoregressive models to embedding models \cite{qwen3embedding}. The Qwen3-1.7B and Qwen3-4B CPT models (2,048-dimensional embeddings, 151,936 vocabulary size) undergo architectural modifications to enable bidirectional embedding generation.

\textbf{Conversion Process:}

The conversion involves four architectural modifications:

\begin{enumerate}

\item\textbf{Removal of Language Modeling Head}: The autoregressive \texttt{lm\_head} layer is removed, eliminating next-token prediction capability while preserving transformer layer representations.

\item\textbf{Bidirectional Attention}: The causal attention mask is replaced with bidirectional attention, enabling the model to attend to all tokens in the input sequence.

\item\textbf{Mean Pooling}: Fixed-size representations are extracted from final hidden states using mean pooling over all token positions.

\item\textbf{Identity Projection}: A projection layer is initialized to maintain the original 2,048-dimensional embedding output.

\end{enumerate}

Our decoder-to-encoder conversion represents a simplified adaptation compared to the official Qwen3 Embedding approach \cite{qwen3embedding}. The official method employs large-scale synthetic data generation (150M samples), multi-stage training, and model merging, while our resource-constrained implementation uses single-stage supervised training on MS MARCO-TR (920K samples) with CachedGISTEmbedLoss. This limitation contributes to the performance gap observed in our results (Section~\ref{sec:experiments}).

\textbf{Post-training Configuration:}

Post-training was conducted on MS MARCO-TR dataset (920,106 triplets) using CachedGISTEmbedLoss with BAAI/bge-m3~\cite{bgem32024} guide model (568M parameters, 1024-dimensional embeddings) for false-negative filtering. Both models were trained on 4$\times$H100 GPUs using AdamW optimizer with learning rate 5$\times$10$^{-6}$. Qwen3-1.7B used per-GPU batch size 8 with gradient accumulation 8 (effective batch 256), while Qwen3-4B used per-GPU batch size 2 with gradient accumulation 16 (effective batch 128). Both models used max sequence length 1,024 tokens. Table~\ref{tab:qwen3_posttraining_config} presents the detailed hyperparameters for Qwen3 CPT models.

\begin{table}[htbp]
\caption{Post-training Configuration for Qwen3 CPT Models}
\label{tab:qwen3_posttraining_config}
\centering
\small
\begin{tabular}{@{}llr@{}}
\toprule
Parameter & Qwen3-1.7B CPT & Qwen3-4B CPT\\
\midrule
Per-GPU Batch & 8 & 2\\
Gradient Accumulation & 8 & 16\\
Effective Batch & 256 & 128\\
Max Sequence Length & 1,024 tokens & 1,024 tokens\\
Temperature ($\tau$) & 0.05 & 0.05\\
Training & 1 epoch & 1 epoch\\
\bottomrule
\end{tabular}

\end{table}

The smaller effective batch sizes (128-256 vs. official Qwen3's thousands) and limited training data diversity represent key constraints in our implementation.

\subsection{Evaluation Methodology}

To assess the quality of embedding models trained on our corpus, we developed MTEB-Turkish, an evaluation framework that extends the Massive Text Embedding Benchmark (MTEB) with Turkish-specific tasks and legal domain corpora. The framework evaluates embedding models across 17 tasks spanning five task types: classification, retrieval, clustering, pair classification, and semantic textual similarity.

\subsubsection{Task Taxonomy}

Table~\ref{tab:mteb-tasks} summarizes the task distribution across categories. The benchmark includes both general-domain tasks adapted for Turkish and three domain-specific legal retrieval tasks constructed from regulatory texts, case law, and contract documents.

\begin{table}[htbp]
\caption{Task Distribution in MTEB-Turkish Benchmark}
\label{tab:mteb-tasks}
\centering
\small
\begin{tabular}{@{}lrr@{}}
\toprule
Task Type & Tasks & Samples\\
\midrule
Classification & 6 & $\sim$9.5K\\
Retrieval & 4 & $\sim$157K\\
Pair Classification & 2 & $\sim$12.5K\\
Clustering & 1 & 99\\
STS & 1 & 208\\
Legal Retrieval & 3 & $\sim$122K\\
\midrule
\textbf{Total} &\textbf{17} & $\sim$\textbf{301K}\\
\bottomrule
\end{tabular}

\end{table}

Representative tasks include MassiveIntent and TurkishMovie for classification, WebFAQ and XQuAD for retrieval, XNLI for pair classification, and domain-specific legal retrieval tasks covering regulatory texts, case law, and contracts.

\subsubsection{Evaluation Metrics}

Each task type employs a primary metric appropriate to its evaluation objective. Classification tasks use accuracy as the primary metric, with F1-score and average precision as secondary measures. Retrieval tasks report nDCG@10 (Normalized Discounted Cumulative Gain) as the primary metric, supplemented by MAP@10, MRR@10, Recall@10, and Precision@10. Clustering performance is measured via V-measure, while semantic textual similarity tasks use Spearman correlation with Pearson correlation as a secondary metric. Pair classification tasks employ average precision as the primary metric.

Throughout this paper, we report \textbf{MTEB Score} (equivalent to Mean TaskType), calculated as the mean score across task types (classification, retrieval, clustering, pair classification, and STS), providing a unified metric for comparing model performance across diverse evaluation tasks.

\subsubsection{Benchmark Construction Methodology}

The MTEB-Turkish benchmark extends the standard MTEB framework with Turkish-specific adaptations and legal domain tasks. General-domain tasks were adapted from existing benchmarks by translating evaluation sets to Turkish or using Turkish-language datasets where available. Legal retrieval tasks were constructed from three domain-specific corpora: (1)\textbf{Contracts}: Turkish legal contract documents from publicly available sources, (2)\textbf{Regulation}: Regulatory texts from Turkish legal frameworks, and (3)\textbf{Caselaw}: Court decisions from Yarg{\i}tay (Court of Cassation) and Dan{\i}\c{s}tay (Council of State). Each legal task follows standard retrieval evaluation protocols with query-document pairs, where queries represent legal information needs and documents are candidate retrieval targets. Train/test splits were constructed to ensure no data leakage between training and evaluation sets, with approximately 80\% of data reserved for training and 20\% for evaluation across all tasks.

\subsubsection{Evaluation Protocol}

All models were evaluated using a consistent evaluation protocol to ensure fair comparison. For each task, models generate embeddings for queries and documents, and similarity scores are computed using cosine similarity. Evaluation metrics are computed on held-out test sets without any fine-tuning or task-specific adaptation. All reported scores represent single-run evaluations, as model checkpoints are deterministic and produce consistent results across evaluation runs. Baseline models (ModernBERT-base~\cite{modernbert2025}, ModernBERT-large~\cite{modernbert2025}) were selected based on their relevance to Turkish language processing and embedding generation tasks, representing encoder-only approaches.

\section{Experiments}

\label{sec:experiments}

Following the methodology described above, we now present detailed experimental analysis of encoder and decoder model training processes, followed by comprehensive evaluation results across all model types.




\subsection{ModernBERT Pre-training Progression}

We present results for encoder models pre-trained from scratch on Turkish-dominant corpora. We evaluate both base (155M parameters) and large (403M parameters) ModernBERT architectures across multiple pre-training dataset versions, examining the relationship between Masked Language Modeling (MLM) performance and downstream retrieval task effectiveness. Our analysis reveals critical insights into optimal checkpoint selection strategies, showing that MLM loss minimization does not guarantee superior embedding quality for Turkish legal domain tasks.

We evaluate the progression of ModernBERT-base models across six pre-training dataset versions (v1 through v6), where each version represents a distinct base model checkpoint trained on progressively refined datasets. Checkpoints v1-v4 underwent identical post-training procedures using GISTEmbed with BGE-M3 guide model (seq\_len=1024), enabling direct comparison of pre-training dataset quality effects on downstream retrieval performance. Checkpoint v5 was evaluated with BGE-M3 guide at extended sequence length (seq\_len=2048). Table~\ref{tab:modernbert_results} presents the performance on Turkish retrieval benchmarks across these versions, where MTEB Score is calculated as Mean TaskType.

\begin{table}[htbp]
\caption{ModernBERT-base Version Progression on Turkish Retrieval Benchmarks}
\label{tab:modernbert_results}
\centering
\small
\begin{tabular}{@{}lrrr@{}}
\toprule
Version & MTEB Score & Legal Score & Contracts\\
\midrule
v1 & 51.45 & 38.12 & 71.23\\
v2 & 53.23 & 39.95 & 73.51\\
v3 & 54.98 & 43.24 & 78.22\\
v4 & 55.79 & 46.31 & 80.81\\
v5 &\textbf{55.76} &\textbf{47.52} & 80.40\\
v6 & 52.76 & 41.89 & 73.86\\
\bottomrule
\end{tabular}

\end{table}

\begin{figure}[!th]

\centering

\includegraphics[width=0.75\linewidth]{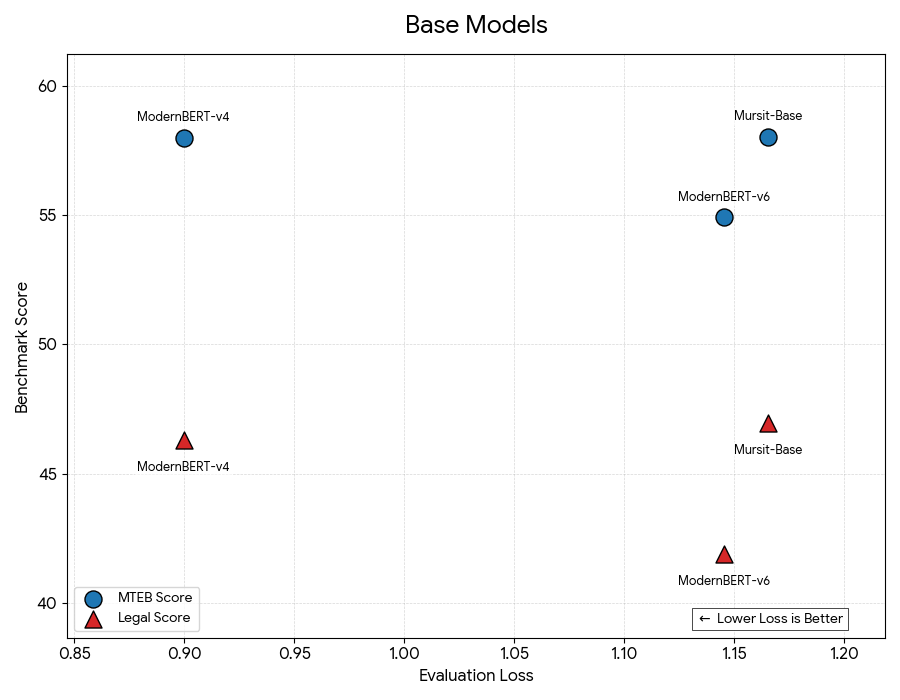}

\caption{Relationship between MLM validation loss and downstream retrieval performance across ModernBERT-base versions v4-v6. We restrict the comparison to these specific versions to ensure parity in pre-training token counts and validation set sizes.}

\label{fig:base_mlm_vs_downstream}

\end{figure}

\begin{figure}[!th]

\centering

\includegraphics[width=0.75\linewidth]{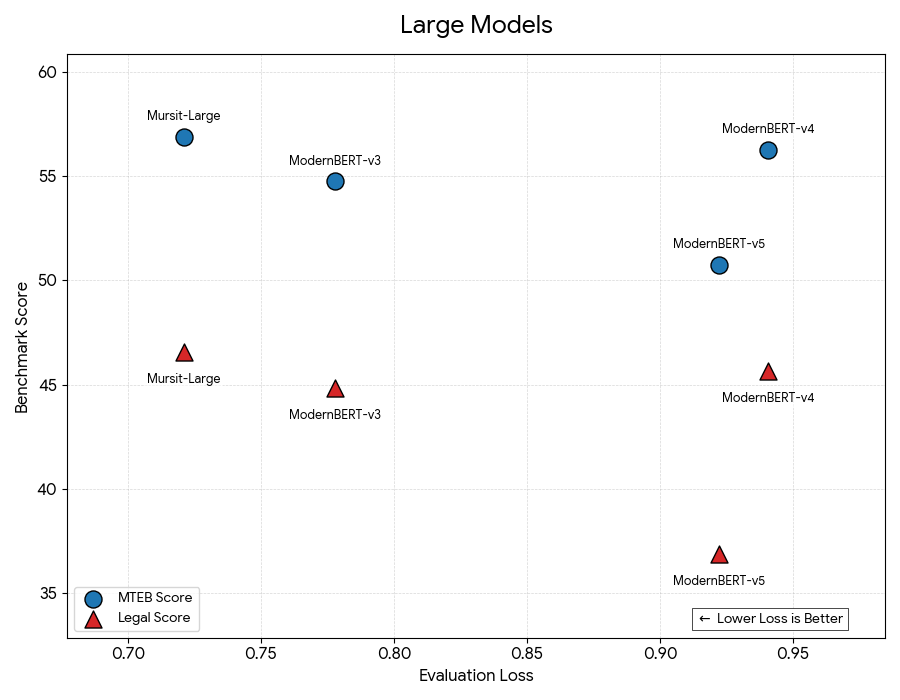}

\caption{Relationship between MLM validation loss and downstream retrieval performance across ModernBERT-large versions.}

\label{fig:large_mlm_vs_downstream}

\end{figure}

The progression demonstrates consistent improvement in downstream retrieval tasks across versions v1 through v5, with the largest performance jump occurring between v2$\rightarrow$v3 (7.2\% MTEB improvement, 8.2\% legal improvement), corresponding to improved URL filtering and extended legal corpus integration. Version v5 achieves the highest legal domain performance (47.52), representing cumulative gains of 12.8\% MTEB and 24.7\% legal improvement from v1. Legal subdomain analysis shows case law retrieval improved by 40.5\% from v2 to v5, indicating that enhanced legal corpus coverage directly benefits specialized legal retrieval tasks.

However, our analysis reveals a critical finding: improvements in Masked Language Modeling (MLM) pre-training objective do not translate linearly to downstream embedding task performance. This observation aligns with prior work demonstrating that MLM loss does not fully correlate with downstream task performance for low-resource and morphologically rich languages \cite{conneau2020xlmr,lester2021prompt,textembedding2024}. As shown in Figure~\ref{fig:base_mlm_vs_downstream}, while MLM loss decreases, downstream retrieval scores exhibit non-linear progression, with performance plateauing after v4-v5. Version v6 demonstrates a significant regression: despite achieving MLM loss lower than Mursit-Base (v5), it shows substantial degradation in both MTEB Score (52.76 vs. 55.76 for v5, -5.4\% relative decrease) and legal domain performance (41.89 vs. 47.52 for v5, -11.9\% relative decrease). This regression is most pronounced in contracts retrieval, where performance drops from 80.40 (v5) to 73.86 (v6), representing an 8.1\% relative decrease.

This observation shows that optimizing the pre-training objective alone is insufficient for embedding quality. Factors such as data quality, domain coverage, and training convergence dynamics play more critical roles in downstream utility. For Turkish, an agglutinative language with complex morphological systems, the optimal checkpoint selection depends on the interplay between morphological complexity, resource availability, and downstream task characteristics. Our results show that intermediate checkpoints (v5 for base models) achieve superior retrieval and classification performance compared to checkpoints with minimal MLM loss. This suggests that morphological complexity and domain-specific requirements create distinct optimization landscapes compared to morphologically simpler languages. The v6 regression indicates that continued pre-training beyond convergence can degrade embedding effectiveness, potentially due to overfitting to the pre-training objective at the expense of generalization to downstream tasks.

For ModernBERT-large models (403M parameters), we observed a distinct progression pattern compared to base models. Figure~\ref{fig:large_mlm_vs_downstream} illustrates the non-linear relationship between MLM loss and downstream retrieval performance for large models. Unlike base models, large models exhibited non-monotonic progression, with the v2 checkpoint achieving the best performance after post-training (56.43 MTEB Score, 46.42 Legal Score), outperforming both earlier (v1) and later versions (v3-v5). Subsequent versions showed progressive degradation: v3 achieved 54.75 overall (44.82 Legal Score), v4 achieved 56.27 overall (45.63 Legal Score), and v5 showed substantial regression to 50.73 overall (36.85 Legal Score). This pattern suggests that optimal pre-training duration and data quantity for large models differs from base models, with v2 representing the sweet spot before overfitting begins to degrade downstream embedding quality. While MLM loss generally decreases monotonically across versions, retrieval performance peaks at v5 for base models and at v2 for large models, then regresses significantly. The 403M parameter large model at v2 achieves competitive performance (56.43 overall) compared to the 155M base model at v5 (55.76 overall), demonstrating that larger architectures can achieve similar performance with less pre-training when optimally tuned.

These findings reinforce the critical importance of direct downstream task evaluation rather than relying solely on pre-training metrics for morphologically rich languages like Turkish where the divergence between MLM optimization and embedding utility is amplified by linguistic complexity and resource constraints \cite{conneau2020xlmr,lester2021prompt,textembedding2024}. The optimal performance range varies based on language structure, resource availability, and downstream task category. Retrieval and classification tasks show distinct optimization landscapes compared to morphologically simpler languages. This emphasizes that model evaluation must consider the specific requirements of target downstream applications, especially for low-resource languages where data scarcity amplifies the importance of checkpoint selection strategies that prioritize downstream utility over pre-training loss minimization.

\subsection{Masked Language Modeling Pre-training Strategy}

Our MLM pre-training follows the ModernBERT approach \cite{devlin2019bert,liu2019roberta,modernbert2025}, where 30\% of token positions are randomly selected for prediction, consistent with RoBERTa's masking strategy rather than BERT's 15\% approach. To mitigate the mismatch between pre-training and fine-tuning (since the [MASK] token does not appear during fine-tuning), the selected tokens are replaced using a three-way strategy: (1) 80\% of the time with the [MASK] token, enabling the model to learn word prediction from context alone; (2) 10\% of the time with a random token, forcing the model to rely on context rather than blindly trusting the input token; (3) 10\% of the time with the unchanged original token, encouraging alignment between word representation and context representation. This 80-10-10 distribution balances multiple learning objectives: context sensitivity, robustness to noisy inputs, and word-context alignment. It also reduces the pre-training/fine-tuning discrepancy inherent in masked language modeling.

We evaluate all models using a 15\% token masking rate, in line with standard BERT benchmarking protocols, even though ModernBERT was pretrained with a 30\% masking rate. This ensures fair comparison across models while preserving the original pretraining objectives.

Table~\ref{tab:mlm_accuracy} presents MLM accuracy scores (averaged across the 80-10-10 strategy) for our pre-trained models and baseline MLM models evaluated on Turkish datasets. Our Mursit models achieve competitive MLM accuracy, with TabiBERT~\cite{kesgin2023turkishbert,turker2025tabibert} achieving the highest score (69.57\%), followed by Mursit-Large (67.25\%), turkish-large-bert-cased~\cite{kesgin2023turkishbert} (65.03\%), and dbmdz/bert-base-turkish-cased (64.98\%). Mursit-Base achieves 64.05\%. The evaluation datasets include blackerx/turkish\_v2, fthbrmnby/turkish\_product\_reviews, hazal/Turkish-Biomedical-corpus-trM, and newmindai/EuroHPC-Legal. Notably, turkish-base-bert-uncased~\cite{kesgin2023turkishbert} was evaluated only on uncased datasets, resulting in a lower score (52.69\%) compared to models evaluated on mixed-case corpora.
\vspace{-0.8\baselineskip}
\begin{table}[htbp]
\caption{MLM Accuracy Scores (80-10-10 Strategy) on Turkish Datasets}
\label{tab:mlm_accuracy}
\centering
\small
\begin{tabular}{@{}lr@{}}
\toprule
Model & MLM Avg (\%)\\
\midrule
boun-tabilab/TabiBERT~\cite{turker2025tabibert} &\textbf{69.57}\\
newmindai/Mursit-Large & 67.25\\
ytu-ce-cosmos/turkish-large-bert-cased~\cite{kesgin2023turkishbert} & 65.03\\
dbmdz/bert-base-turkish-cased & 64.98\\
newmindai/Mursit-Base & 64.05\\
KocLab-Bilkent/BERTurk-Legal~\cite{ozturk2023berturklegal} & 54.10\\
ytu-ce-cosmos/turkish-base-bert-uncased~\cite{kesgin2023turkishbert} & 52.69\\
\bottomrule
\end{tabular}
\vspace{-0.3\baselineskip}
\parbox{\columnwidth}{\scriptsize\textit{MLM accuracy averaged across the 80-10-10 masking strategy. turkish-base-bert-uncased was evaluated only on uncased datasets. Evaluation datasets: blackerx/turkish\_v2, fthbrmnby/turkish\_product\_reviews, hazal/Turkish-Biomedical-corpus-trM, newmindai/EuroHPC-Legal. All experiments are reproducible (see Section~\ref{app:mlm_benchmark_results}).}}

\end{table}

\subsection{Encoder Post-training Analysis and Ablation Studies}\label{subsec:encoder_posttraining}

We evaluate three contrastive learning approaches for post-training ModernBERT-base-v4 checkpoint (155M parameters): standard InfoNCE, Qwen3-style InfoNCE, and GISTEmbed with cached guide models. All experiments used identical hyperparameters (batch size=768, learning rate=2$\times$10$^{-5}$, sequence length=1024) to ensure fair comparison. Table~\ref{tab:encoder_posttraining} presents the performance comparison on MTEB-Turkish benchmark, where MTEB Score is calculated as Mean TaskType (mean across task types: Classification, Clustering, Pair Classification, Retrieval, STS).

\begin{table}[htbp]
\caption{Comparison of Post-training Techniques on ModernBERT-base-v4}
\label{tab:encoder_posttraining}
\centering
\small
\begin{tabular}{@{}lrr@{}}
\toprule
Technique & MTEB Score & Legal Score\\
\midrule
InfoNCE (baseline) & 53.23 & 39.95\\
Qwen3-InfoNCE~\cite{qwen3embedding} & 56.70 & 42.60\\
GISTEmbed (BGE-M3~\cite{bgem32024}) &\textbf{55.67} &\textbf{46.36}\\
\bottomrule
\end{tabular}

\end{table}

GISTEmbed with guide models significantly outperforms standard contrastive learning, achieving 4.6\% overall improvement and 15.9\% legal domain improvement over the InfoNCE baseline, demonstrating the effectiveness of false-negative filtering for Turkish legal retrieval tasks.

To demonstrate the effectiveness of our post-training approach, we compare models across three categories: (1)\textbf{Pre-trained by owner}: Original ModernBERT models pretrained by the architecture authors; (2)\textbf{Pre-trained by us}: ModernBERT models pretrained from scratch on our Turkish-dominant corpus; (3)\textbf{Post-trained}: Our pre-trained models after embedding post-training with contrastive learning. Table~\ref{tab:pretrained_vs_posttrained} presents this comparison.

\begin{table}[htbp]
\caption{Comparison: Pre-trained vs Post-trained ModernBERT Models}
\label{tab:pretrained_vs_posttrained}
\centering
\small
\begin{tabular}{@{}lrr@{}}
\toprule
Model & MTEB Score & Legal Score\\
\midrule
\multicolumn{3}{l}{\textit{Post-trained:}}\\
Mursit-Large-TR-Retrieval &\textbf{56.43} & 46.42\\
Mursit-Base-TR-Retrieval & 55.86 &\textbf{47.52}\\
\midrule
\multicolumn{3}{l}{\textit{Pretrained:}}\\
Mursit-Large &\textbf{41.75} &\textbf{23.71}\\
Mursit-Base & 40.23 & 17.93\\
\midrule
\multicolumn{3}{l}{\textit{Original ModernBERT:}}\\
ModernBERT-base &\textbf{23.80} &\textbf{2.99}\\
ModernBERT-large & 23.74 & 2.44\\
\bottomrule
\end{tabular}

\end{table}

Key observations: (1) Our checkpoint 4 post-trained model with BGE-M3 guide (55.67 MTEB Score, 46.36 Legal Score) demonstrates that domain-specific pre-training from scratch combined with effective post-training achieves competitive performance; (2) Progressive improvement from checkpoint 1 to checkpoint 5 (12.8\% overall improvement, 23.2\% legal improvement) validates our iterative pre-training dataset refinement strategy.

We conduct ablation studies to investigate the relationship between training hyperparameters and domain-specific performance. The ablation was conducted on Mursit-Large-TR-Retrieval (403M parameters) using Qwen3-InfoNCE loss function, with all other hyperparameters held constant (batch size=2304, learning rate=2$\times$10$^{-5}$). Token distribution analysis of MS MARCO-TR indicates 99.9\% sample coverage at 256 tokens; however, reducing sequence length to match training data distribution resulted in significant performance degradation on downstream legal retrieval tasks. Table~\ref{tab:seq_ablation_exp} presents the impact of sequence length configuration on legal domain performance metrics.

\begin{table}[htbp]
\caption{Effect of Sequence Length on Legal Domain Performance (Mursit-Large-TR-Retrieval, Qwen3-InfoNCE)}
\label{tab:seq_ablation_exp}
\centering
\small
\begin{tabular}{@{}lrrrr@{}}
\toprule
Seq. Length & MTEB Score & Legal Score & $\Delta$ Legal Score\\
\midrule
256 & 57.85 & 42.60 & -8.5\%\\
1,024 & 58.21 & 44.89 & -3.6\%\\
2,048 & 56.43 & 46.42 & --\\
\bottomrule
\end{tabular}

\end{table}

Models trained with seq\_len=256 exhibited 8.5\% degradation in legal domain performance relative to seq\_len=2048, despite achieving equivalent MS MARCO-TR validation scores. The degradation is most pronounced in regulation retrieval (23.6\% relative decrease) and case law retrieval (17.8\% relative decrease), suggesting that long-context understanding is essential for legal document retrieval. These findings demonstrate that sequence length optimization must consider target evaluation corpus characteristics rather than training data distribution alone. While shorter sequences reduce computational requirements by approximately 67\%, the associated performance degradation in specialized domains may outweigh efficiency gains.




\subsection{Decoder Model Results}

This section presents results for the decoder models adapted through CPT on Turkish legal corpora.

\subsubsection{Training Process and Stability Analysis}

We evaluate the four-phase CPT process of the Qwen3-1.7B model and the single-phase CPT training of the Qwen3-4B model in terms of training stability and convergence behavior. Analyses focus primarily on loss curves, step-based convergence, and training stability.

\textbf{Qwen3-1.7B -- Phase 1:} In the first phase, the Qwen3-1.7B model was trained for 2 epochs over 4,170 steps. Training loss decreased from 2.358 to 1.927 (18.3\% reduction), achieving minimum loss of 1.877 at step 3,771. Validation loss decreased from 2.273 to 1.994, with minimum of 1.994. During the second epoch, validation loss stabilization was observed, indicating convergence completion for this phase.

\textbf{Qwen3-1.7B -- Phase 2:} In the second phase, training was extended to 32,260 steps for 1 epoch. Loss curves converged more aggressively, with training loss dropping from 1.486 to 1.053 (29.1\% reduction) and reaching minimum of 1.002 at step 28,942. Validation loss decreased from 1.409 to 1.081, with minimum of 1.081. Parameters inherited from Phase 1 were updated stably throughout training.

\textbf{Qwen3-1.7B -- Phase 3:} The third phase was conducted as a long-term CPT process of 46,803 steps for 1 epoch. Training loss decreased from 1.872 to 1.429 (23.7\% reduction), with minimum loss of 1.377 achieved at step 44,448. Validation loss decreased from 1.856 to 1.446, with minimum of 1.446. Knowledge learned in previous phases was preserved, and no signs of catastrophic forgetting were observed.

\textbf{Qwen3-1.7B -- Phase 4:} The fourth phase continued training for 14,122 steps for 1 epoch. Training loss decreased from 1.766 to 1.575 (10.8\% reduction), with minimum loss of 1.549 achieved at step 12,466. Validation loss decreased from 1.632 to 1.566, with minimum of 1.566. This final phase consolidated domain knowledge while maintaining stable convergence.

\begin{figure}[htbp]

\centering

\includegraphics[width=0.95\columnwidth]{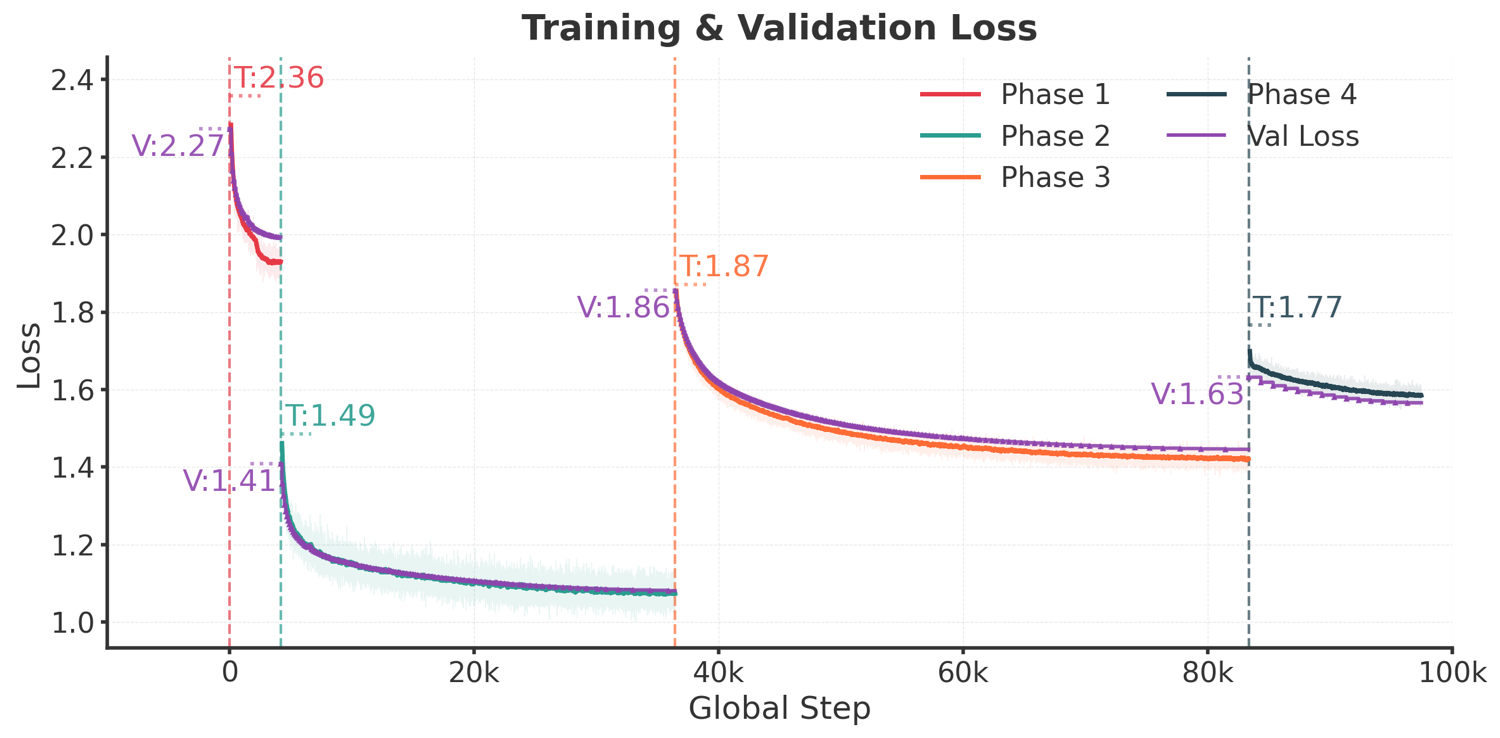}

\caption{Qwen3-1.7B CPT Training and Validation Loss Across Four Phases.}

\label{fig:qwen3-1.7b_loss}

\end{figure}

Figure~\ref{fig:qwen3-1.7b_loss} shows the training and validation loss curves across four phases. Vertical dashed lines indicate phase transitions. The validation loss (purple) consistently tracks training loss, indicating stable learning without overfitting.

\textbf{Qwen3-4B -- Single-Phase CPT:} The Qwen3-4B model was trained with a single-phase CPT process of 153,508 steps. Thanks to higher parameter capacity, the model exhibited smooth and stable convergence throughout long-term training.

\begin{figure}[htbp]

\centering

\includegraphics[width=0.95\columnwidth,keepaspectratio]{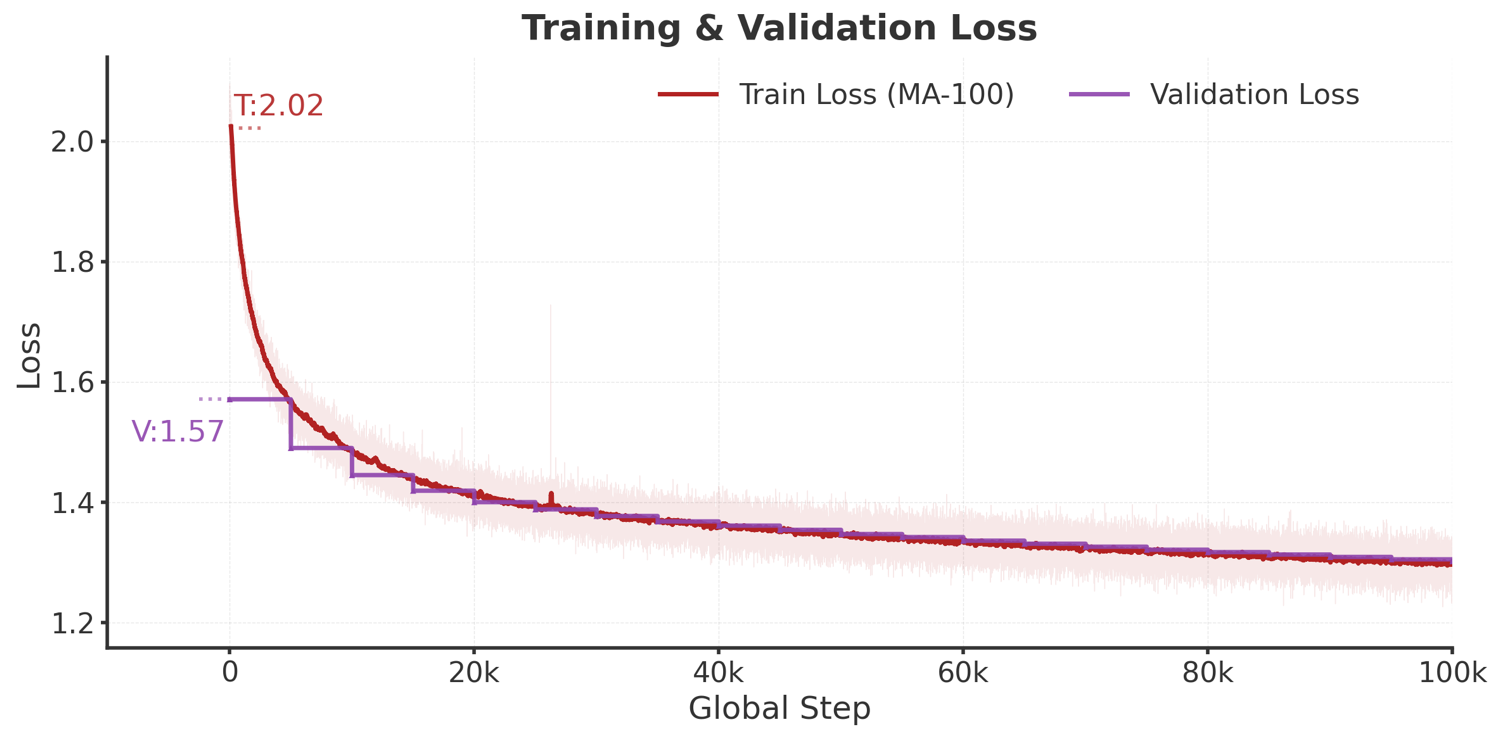}

\caption{Qwen3-4B CPT Training and Validation Loss Curves.}

\label{fig:qwen4b_loss}

\end{figure}

As shown in Figure~\ref{fig:qwen4b_loss}, training loss (blue) and validation loss (orange) decreased over 153,508 steps. Training loss decreased from 2.022 to 1.273 (37.1\% reduction), with minimum loss of 1.202 at step 116,651. Validation loss decreased from 1.571 to 1.284 (18.5\% reduction), converging to final validation loss of 1.284. No instabilities or sudden performance drops were observed during the training process.

\textbf{Comparative Evaluation:} The multi-phase CPT structure applied to the Qwen3-1.7B model provided an effective strategy for increasing training stability in small-scale models. The Qwen3-4B model, thanks to its high capacity, showed stable training behavior even in a single-phase CPT process. These observations indicate that the importance of gradual CPT strategies increases as model scale decreases.

\subsubsection{Evaluation Methodology}

Model performance was measured using the perplexity (PPL) metric on the EuroHPC-Legal evaluation dataset \cite{AITEAM2025124}\footnote{\href{https://huggingface.co/datasets/newmindai/EuroHPC-Legal}{huggingface.co/datasets/newmindai/EuroHPC-Legal}}. During evaluation, a context length of 512 tokens per example was used for Qwen3-1.7B and 1024 tokens per example for Qwen3-4B, based on the truth field content of the dataset.

Comparisons were conducted across two different model configurations. First, the pre-CPT state of the Qwen3-1.7B base model was compared with outputs from the four-phase CPT process (Phase-1, Phase-2, Phase-3, Phase-4). This approach enables direct observation of the effect of increasing CPT step count and curriculum structure on domain-specific language modeling performance.

Additionally, single-phase CPT training was applied to the Qwen3-4B model, and the results obtained were compared with the model's pre-CPT state. Thus, both the effects of different CPT strategies (multi-phase vs. single-phase) and model scale (1.7B vs. 4B) on the Turkish legal domain were evaluated together.
\vspace{-0.3\baselineskip}
\subsubsection{Qwen3 CPT Perplexity Results}

Table~\ref{tab:Qwen3-1.7B_results} shows the perplexity values for the Qwen3-1.7B model across different legal subdomains between the base model and CPT phases. Results show that PPL values decreased consistently and monotonically at each stage of the CPT process.

\begin{table}[htbp]
\caption{Qwen3-1.7B Perplexity Across CPT Phases}
\label{tab:Qwen3-1.7B_results}
\centering
\scriptsize
\begin{tabular}{@{}lrrrrrr@{}}
\toprule
Domain & Base & P1 & P2 & P3 & P4 & Impr.\\
\midrule
CL & 5.96 & 4.66 & 4.07 & 3.78 & 3.74 & 37.2\%\\
CCML & 9.90 & 7.11 & 5.97 & 5.58 & 5.53 & 44.2\%\\
ENRL & 9.77 & 7.39 & 6.15 & 5.81 & 5.80 & 40.7\%\\
EnvL & 10.40 & 7.34 & 5.89 & 5.52 & 5.56 & 46.6\%\\
FL & 9.98 & 7.12 & 5.70 & 5.30 & 5.35 & 46.4\%\\
HL & 4.93 & 4.09 & 3.67 & 3.40 & 3.35 & 32.1\%\\
IPL & 4.63 & 4.06 & 3.58 & 3.40 & 3.22 & 30.4\%\\
LabL & 4.78 & 3.53 & 3.20 & 2.88 & 2.76 & 42.4\%\\
LUZL & 9.11 & 7.01 & 5.62 & 5.17 & 5.16 & 43.4\%\\
PDPL & 8.36 & 6.62 & 5.23 & 4.73 & 4.71 & 43.7\%\\
TL & 7.29 & 5.20 & 4.57 & 4.09 & 3.98 & 45.4\%\\
\midrule
\textbf{OVERALL}$^\dagger$ &\textbf{9.22} &\textbf{6.93} &\textbf{5.66} &\textbf{5.26} &\textbf{5.24} &\textbf{43.1\%}\\
\bottomrule
\end{tabular}

\parbox{\columnwidth}{\scriptsize\textit{CL: Competition Law, CCML: Corporate Capital Market Law, ENRL: Energy and Natural Resources Law, EnvL: Environmental Law, FL: Fund Law, HL: Health Law, IPL: Intellectual Property Law, LabL: Labor Law, LUZL: Land Use and Zoning Law, PDPL: Personal Data Protection Law, TL: Tax Law. $^\dagger$Overall values are computed as token-weighted averages across all legal subdomains.}}

\end{table}

The phase-by-phase progression for Qwen3-1.7B reveals distinct improvement patterns across legal subdomains. Domains with initially higher perplexity values (CCML: 9.90, EnvL: 10.40, FL: 9.98) showed the most substantial reductions, achieving 44.2\%, 46.6\%, and 46.4\% improvements respectively, indicating that CPT effectively addresses domain-specific terminology and conceptual complexity. Conversely, domains with lower initial perplexity (HL: 4.93, IPL: 4.63) showed more modest but consistent improvements (32.1\% and 30.4\%), suggesting that CPT provides incremental gains even in domains where the base model already demonstrates reasonable performance. The largest single-phase improvement occurred between Phase 1 and Phase 2, where overall perplexity decreased from 6.93 to 5.66 (18.3\% reduction), corresponding to the introduction of legal content with domain-specific terminology. Subsequent phases (P2$\rightarrow$P3 and P3$\rightarrow$P4) showed diminishing returns, with Phase 4 achieving only marginal improvement over Phase 3 (5.26 vs. 5.24), indicating convergence of the curriculum learning strategy.

In parallel, Table~\ref{tab:qwen4b_results} presents perplexity results for the Qwen3-4B model between the base model and post-CPT state. Thanks to the larger model scale, Qwen3-4B achieved lower absolute PPL values across all legal subdomains and strongly benefited from the CPT process.
\begin{table}[htbp]
\caption{Qwen3-4B Perplexity Results}
\label{tab:qwen4b_results}
\centering
\scriptsize
\begin{tabular}{@{}lrrr@{}}
\toprule
Domain & Base & CPT & Impr.\\
\midrule
CL & 4.986 & 3.493 & 29.9\%\\
CCML & 8.189 & 5.140 & 37.2\%\\
ENRL & 8.159 & 5.388 & 34.0\%\\
EnvL & 8.601 & 5.107 & 40.6\%\\
FL & 8.277 & 4.897 & 40.8\%\\
HL & 4.155 & 3.205 & 22.9\%\\
IPL & 3.743 & 3.076 & 17.8\%\\
LabL & 3.926 & 2.702 & 31.2\%\\
LUZL & 7.559 & 4.818 & 36.3\%\\
PDPL & 6.873 & 4.451 & 35.2\%\\
TL & 6.111 & 3.796 & 37.9\%\\
\midrule
\textbf{OVERALL}$^\dagger$ &\textbf{7.649} &\textbf{4.884} &\textbf{36.2\%}\\
\bottomrule
\end{tabular}

\parbox{\columnwidth}{\scriptsize\textit{Domain abbreviations same as Table~\ref{tab:Qwen3-1.7B_results}. $^\dagger$Overall values are computed as token-weighted averages across all legal subdomains.}}

\end{table}

For Qwen3-4B, the single-phase CPT approach achieved comparable overall improvement (36.2\%) to the four-phase Qwen3-1.7B strategy (43.1\%), despite processing all training data in a single stage. This demonstrates that larger model capacity can compensate for the absence of curriculum structure, enabling effective domain adaptation through direct exposure to mixed-complexity legal content. This finding highlights a critical trade-off between training efficiency and model scale, suggesting that practitioners can choose simpler training protocols when computational resources permit larger architectures. The 4B model's superior absolute perplexity values (4.884 vs. 5.24 for 1.7B Phase-4) across all subdomains validate the parameter efficiency hypothesis, where increased model scale provides better generalization even with less structured training. These results provide compelling evidence that architectural capacity fundamentally influences learning dynamics, with larger models exhibiting enhanced robustness to training protocol variations and demonstrating superior adaptation capabilities across diverse legal subdomains.

Domain-specific analysis reveals that both models show strongest improvements in regulatory and financial law domains (CCML, FL, EnvL), where specialized terminology and normative structures require deeper domain understanding, while intellectual property and health law domains show more modest gains, potentially due to their more standardized terminology or smaller corpus representation in the training data. These domain-specific variations underscore the importance of corpus composition and domain complexity in determining CPT effectiveness. When results are evaluated together: For the Qwen3-1.7B model, PPL values improved stably across all legal subdomains at each CPT phase; at the end of Phase-4, the lowest perplexity values were reached, especially in conceptually and terminologically intensive legal domains; the Qwen3-4B model achieved high-rate PPL improvements even with single-phase CPT, and model scale significantly increased domain-specific language modeling performance; increasing CPT step count and model parameter scale directly and positively contributed to language modeling capacity in the Turkish legal domain, establishing clear guidelines for future model development and resource allocation strategies across different computational budgets.

Table~\ref{tab:qwen_checkpoints} shows checkpoint progression for both models.

\begin{table}[H]
\caption{CPT Checkpoint Progression}
\label{tab:qwen_checkpoints}
\centering
\footnotesize
\begin{tabular}{@{}lrr@{}}
\toprule
Model & Perplexity$^\dagger$ & Gain (\%)\\
\midrule
Mecellem-Qwen3-1.7B-TR (Ckpt. 1) & 6.93 & +24.9\%\\
Mecellem-Qwen3-1.7B-TR (Ckpt. 2) & 5.66 & +38.7\%\\
Mecellem-Qwen3-1.7B-TR (Ckpt. 3) & 5.26 & +43.0\%\\
Mecellem-Qwen3-1.7B-TR (Ckpt. 4) & 5.24 & +43.1\%\\
Qwen3-1.7B-Base & 9.22 & --\\
\midrule
Mecellem-Qwen3-4B-TR (Final) & 4.88 & +36.2\%\\
Qwen3-4B-Base & 7.65 & --\\
\bottomrule
\end{tabular}

\parbox{\columnwidth}{\scriptsize\textit{$^\dagger$Perplexity values are computed as token-weighted averages across all legal subdomains.}}

\end{table}

\subsubsection{Additional Experiment: Turkish Quality Filtering Refinement}

\label{subsec:turkish_filtering_refinement}

As a separate investigation independent of the main CPT training pipeline, we conducted an additional experiment to evaluate the effect of Turkish morphology-based quality filtering. This experiment used the final checkpoints from the main training runs (Mecellem-Qwen3-1.7B-TR Phase~4 and Mecellem-Qwen3-4B-TR ) as initialization and was not part of the primary CPT procedure described in Section~\ref{sec:methodology}.

Additional training was performed on a refinement dataset filtered using Turkish Quality Filtering criteria (suffix entropy $>75\%$, lemma diversity $>50\%$) and only used data where the token count is above 502 tokens. The dataset consisted mainly of previously seen Turkish data subjected to morphology-based filtering, with small amounts of unfiltered multilingual and technical content retained to preserve general capabilities. The goal was to assess whether post-hoc morphological filtering could further improve performance beyond the main CPT results.

Table~\ref{tab:filtered_dataset_composition} shows the composition of the refinement dataset, totaling 3.66B tokens, of which 94.6\% are filtered Turkish content.

\begin{table}[htbp]
\caption{Turkish Quality Filtering Refinement Dataset Composition}
\label{tab:filtered_dataset_composition}
\centering
\small
\begin{tabular}{@{}lrr@{}}
\toprule
Source & Tokens & Percentage\\
\midrule
Filtered Turkish & 3,464,365,893 & 94.6\%\\
English & 72,917,544 & 2.0\%\\
Math & 41,855,480 & 1.1\%\\
Python & 47,165,175 & 1.3\%\\
Arabic & 10,356,834 & 0.3\%\\
Chinese & 9,512,075 & 0.3\%\\
Spanish & 10,196,529 & 0.3\%\\
Russian & 6,500,781 & 0.2\%\\
\midrule
\textbf{Total} &\textbf{3,662,870,311} &\textbf{100\%}\\
\bottomrule
\end{tabular}

\end{table}

\subsubsection{Evaluating Decoder-Only Online Models with a Domain-Specific Armo Reward Model in Legal Tasks}

Although decoder-only online language models demonstrate strong surface-level fluency in legal text generation, their performance remains difficult to assess reliably with respect to domain-specific quality dimensions such as statutory compliance, case law grounding, and depth of legal reasoning \cite{legalai2020}. Widely adopted automatic metrics and likelihood-based scores primarily capture linguistic coherence, but fail to reflect the correctness, normative consistency, and evidentiary grounding required in legal reasoning. Traditional evaluation approaches often rely on perplexity or BLEU-like metrics that measure text similarity rather than legal accuracy, leaving critical gaps in assessing whether generated responses correctly cite relevant statutes, accurately interpret legal precedents, or provide comprehensive legal analysis. Moreover, legal text evaluation requires context-dependent assessment where the importance of different quality dimensions varies based on the type of legal question being addressed—for instance, regulatory queries may prioritize statute accuracy while case law questions may emphasize precedent citation quality. Consequently, there is a clear need for a fine-grained, domain-aware benchmarking framework capable of evaluating legal response quality beyond surface-level text statistics, one that can adaptively weight evaluation criteria based on the legal context and provide interpretable feedback across multiple legally grounded dimensions.

For evaluation, we employ Muhakim, a domain-specific multi-objective reward model trained for Turkish legal text assessment. Built upon a Llama-3.1-based \cite{liu2025skywork_reward_v2,llama3_2024} reward backbone and augmented with a mixture-of-experts gating mechanism, the model produces fine-grained quality scores across five legally grounded dimensions:
\begin{itemize}
\setlength{\itemsep}{0pt}
\setlength{\parsep}{0pt}
\item \textbf{Statute Reference}: {\small Accuracy of legal statute citations}
\item \textbf{Legal Accuracy}: {\small Correctness of legal information}
\item \textbf{Case Law Reference}: {\small Proper citation of legal precedents}
\item \textbf{Linguistic Coherence}: {\small Language quality and fluency}
\item \textbf{Depth Coverage}: {\small Comprehensiveness of the response}
\end{itemize}
The model is trained using a combination of regression-based supervision and preference learning, enabling stable and interpretable scoring suitable for benchmarking decoder-only online models.

The reward model architecture employs a multi-objective framework where: (i)\textbf{Gating}: Context-dependent importance weights assigned by the gating network (MoE layer) based on the input prompt. The gating layer outputs non-negative coefficients (summing to 1) that determine how much weight each reward objective should receive. (ii)\textbf{Rewards}: Multi-objective reward predictions from ArmoRM's regression layer, representing model performance on each objective. (iii)\textbf{Score}: Final preference score = $\Sigma(\text{gating}[i]\times\text{transformed\_rewards}[i])$. Critically, gating depends on the input prompt, while rewards depend on the model output, enabling context-aware evaluation that adapts importance weights based on the legal question while assessing response quality. Further technical details on the prompt-conditioned gating and response-conditioned reward mechanisms are provided in~\ref{app:reward_architecture}. The Muhakim reward model is publicly available.

\begin{figure}[H]

\centering

\includegraphics[width=0.85\columnwidth,keepaspectratio]{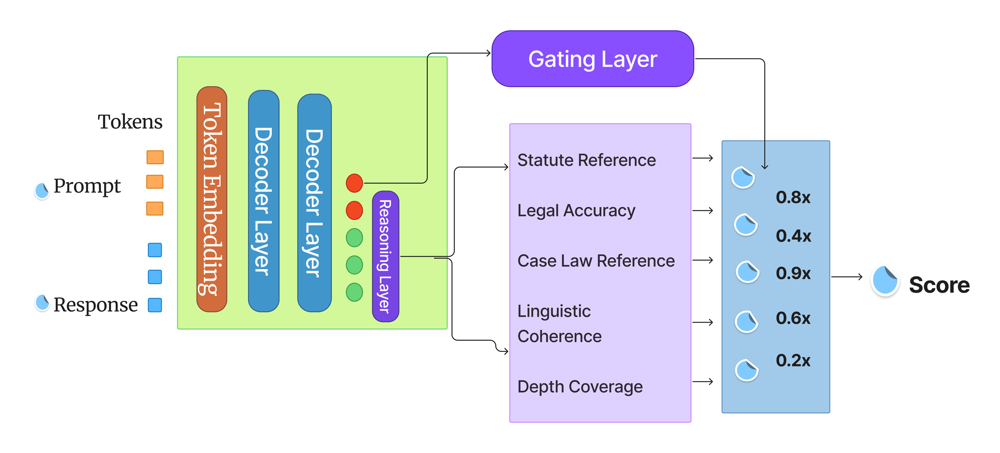}

\caption{Muhakim Model Training Pipeline.}

\label{fig:muhakim-pipeline}

\end{figure}

The ArmoRM-Turkish-Legal (Muhakim) reward model is built on the Skywork-Reward-V2-Llama-3.1-8B backbone and is trained through a multi-stage procedure tailored to the Turkish legal domain. This domain-specific adaptation addresses critical limitations of general-purpose reward models, which often fail to capture the nuanced requirements of legal text evaluation, including normative consistency, statutory accuracy, and precedent-based reasoning essential for reliable legal AI assessment. The training pipeline consists of three components: (i) multi-objective supervision that enables independent learning of five legal quality dimensions, (ii) preference-based training of a mixture-of-experts gating network to capture context-dependent importance of these dimensions, and (iii) a debiasing stage designed to mitigate length-related reward artifacts.\cite{ArmoRM} Figure~\ref{fig:muhakim-pipeline} illustrates the complete training pipeline. This training design allows the model to produce stable, interpretable, and context-aware reward signals, making it suitable for benchmarking decoder-only online language models in legal tasks.

The benchmark is designed to comparatively evaluate the performance of decoder-only online language models under varying contextual conditions in legal text generation. This evaluation framework addresses a fundamental challenge in legal AI: understanding how model performance scales with available context, which directly impacts practical deployment scenarios where reference documents may be partially available or truncated due to token limits. For this purpose, we use the newmindai/EuroHPC-Legal dataset, which consists of legal questions (question) paired with reference texts (truth). The dataset contains 116 high-quality question-answer pairs. From each reference text, the first 5, 10, 20, 50, and 100 tokens are extracted to construct five distinct context-length settings.

For each context length, the same question and partial reference context are provided using identical prompt structures and token constraints to ensure fair, controlled, and reproducible comparisons across model pairs. This approach isolates domain adaptation effects by controlling all variables except model training:
\begin{itemize}
\setlength{\itemsep}{0pt}
\setlength{\parsep}{0pt}
\item Qwen3-1.7B-Base vs Mecellem-Qwen3-1.7B-TR
\item Qwen3-4B-Base vs Mecellem-Qwen3-4B-TR
\end{itemize}
The generated responses are evaluated using the Muhakim reward model in a conversational format, where the user message contains both the legal question and partial reference text, while the assistant message consists of the model-generated response, enabling context-aware assessment of both linguistic quality and legal context alignment. For each evaluation instance, the reward model produces an overall quality score (Score), per-objective reward values (Rewards), and context-dependent gating outputs (Gating) across five quality dimensions: statutory grounding, legal accuracy and currency, case law referencing quality, linguistic coherence, and analytical depth and coverage.

\begin{figure}[H]

\centering

\includegraphics[width=0.85\columnwidth,keepaspectratio]{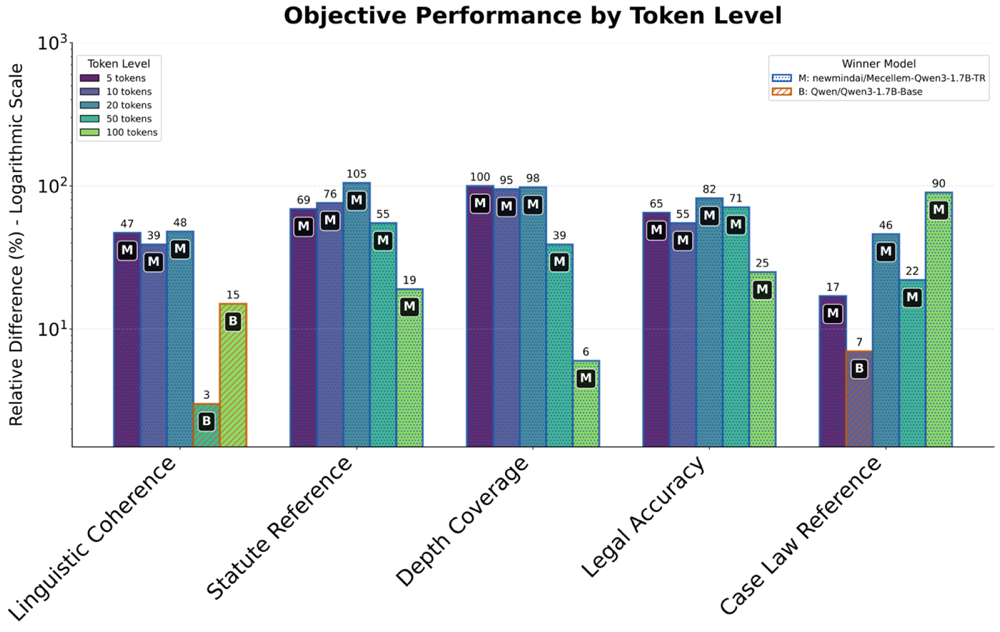}

\caption{Benchmark Performance of 1.7B Decoder-Only Models Across Context Lengths Using the Muhakim Reward Model.}

\label{fig:benchmark-1.7b}

\end{figure}

\begin{figure}[H]

\centering

\includegraphics[width=0.85\columnwidth,keepaspectratio]{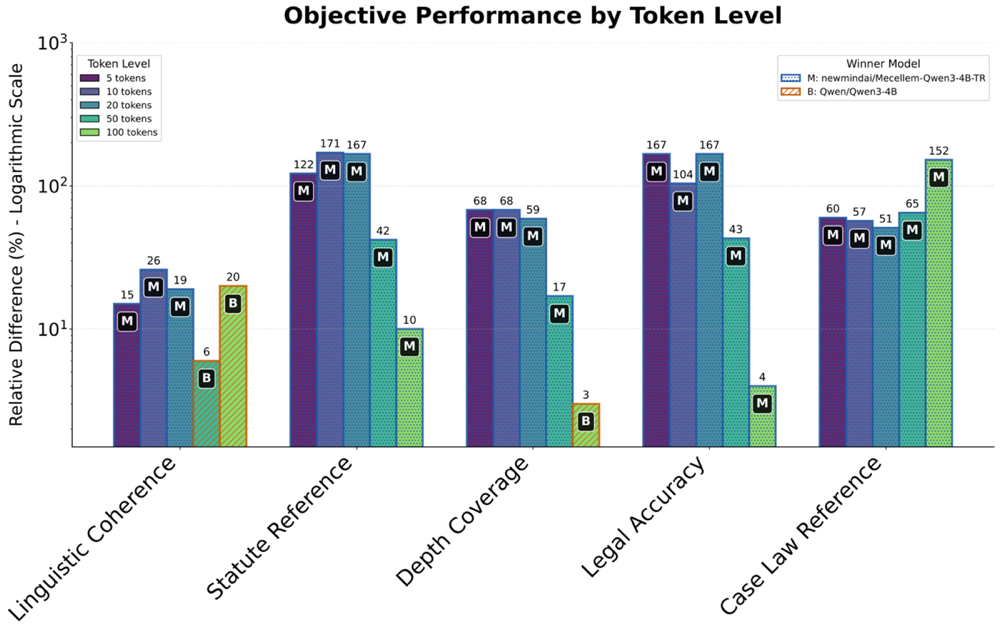}

\caption{Benchmark Performance of 4B Decoder-Only Models Across Context Lengths Using the Muhakim Reward Model.}

\label{fig:benchmark-4b}

\end{figure}

We first analyze the benchmark results shown in Figure~\ref{fig:benchmark-1.7b} for the 1.7B-scale decoder-only models under varying context-length constraints (5, 10, 20, 50, and 100 tokens). Under short and medium context settings (5–20 tokens), newmindai/Mecellem-Qwen3-1.7B-TR consistently outperforms the corresponding base model across all five legal quality objectives. Relative performance gains are most pronounced for depth of coverage, statute reference usage, and legal accuracy, with improvements frequently exceeding 100

As shown in Figure~\ref{fig:benchmark-4b}, we observe similar but more accentuated trends at the 4B parameter scale. Under severely reduced context conditions (5–20 tokens), newmindai/Mecellem-Qwen3-4B-TR consistently outperforms the corresponding base model across all evaluated objectives, demonstrating the effectiveness of legal domain alignment when contextual information is highly constrained. Relative performance differences are most pronounced for legal accuracy, depth coverage, and statute reference usage, with gains frequently exceeding 100

\paragraph{Rewards Comparison Analysis}

\begin{figure}[H]

\centering

\includegraphics[
  width=\columnwidth,
  height=0.42\textheight,
  keepaspectratio
]{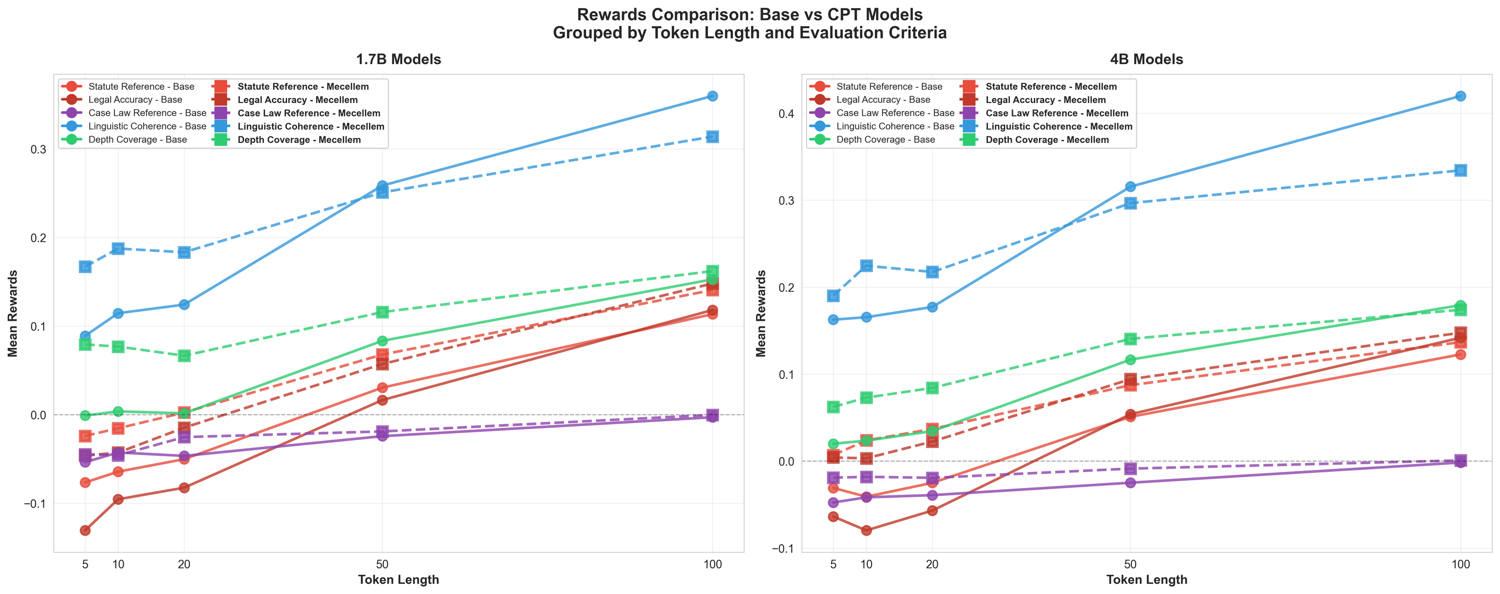}

\caption{Rewards Comparison: Base vs CPT Models Across Token Lengths.}

\label{fig:rewards-comparison}

\end{figure}

Figure~\ref{fig:rewards-comparison} summarizes mean reward scores for Base and CPT (Mecellem) models across token lengths for both the 1.7B and 4B scales. The figure shows mean reward values for five evaluation dimensions (statute reference, legal accuracy, case law reference, linguistic coherence, depth coverage) at token lengths 5, 10, 20, 50, 100. Overall, CPT models consistently outperform Base models and achieve higher mean rewards in short and medium context settings, with the largest gains observed under constrained contexts where Base model scores are near zero or negative. As context length increases, performance gaps gradually narrow. For the 1.7B models, CPT variants maintain a consistent advantage across all token lengths, whereas for the 4B models, convergence is observed at the longest context setting, where the Base model slightly outperforms CPT. Across both scales, mean reward scores generally increase with longer token lengths, indicating improved performance with richer contextual information.

\subsubsection{Mixed Precision Training Strategies}

The NVIDIA NeMo framework provides multiple mixed precision training strategies to optimize memory usage and accelerate training for large language models. These strategies enable storing and processing model parameters and computations in different numerical precision formats. The \texttt{MegatronMixedPrecision} class in NeMo configures these precision settings, offering predefined precision functions for common use cases.

\subsubsection*{Precision Strategy Options:}

\begin{itemize}

\item\textbf{BF16 Mixed Precision}: Basic 16-bit brain floating-point precision mode (configured via \texttt{bf16\_mixed}). Compatible with all GPUs, provides good numerical stability for gradient updates.

\item\textbf{FP16 Mixed Precision}: Basic 16-bit floating-point precision mode (configured via \texttt{fp16\_mixed}). Compatible with all GPUs, offers faster computation than BF16 but may have reduced numerical stability.

\item\textbf{BF16/FP16 with FP8 Hybrid}: FP8 hybrid format using E4M3 for forward pass and E5M2 for backward pass, with delayed scaling recipe (configured via \texttt{bf16\_with\_fp8\_mixed} or \texttt{fp16\_with\_fp8\_mixed}). Optimizes scaling factors using amax history tracking to maintain training stability while reducing memory usage.

\item\textbf{BF16/FP16 with FP8 Current Scaling}: FP8 hybrid format with tensorwise (per-tensor) scaling recipe (configured via \texttt{bf16\_with\_fp8\_current\_scaling\_mixed} or \texttt{fp16\_with\_fp8\_current\_scaling\_mixed}). Maintains first and last transformer layers in BF16/FP16 for gradient stability, providing a more conservative approach to FP8 adoption.

\end{itemize}

\subsubsection*{Ablation Study Methodology:}

We conducted an ablation study comparing these mixed precision strategies during Qwen3 CPT. All precision configurations were evaluated by training Qwen3-4B for 1,000 steps under identical settings to ensure a fair and controlled comparison. Table~\ref{tab:precision_ablation} presents training speed (step time), speedup relative to the BF16 baseline, and training stability metrics (first and final loss values).

\begin{table}[htbp]
\caption{Ablation Study: Mixed Precision Training Strategies}
\label{tab:precision_ablation}
\centering
\small
\begin{tabular}{@{}lrrrr@{}}
\toprule
Method & Step (s) & Speedup & First Loss & Final Loss\\
\midrule
FP16-FP8 & 0.467 & 20.5\% & 1.342 & 1.250\\
FP16-FP8-CS & 0.4683 & 20.2\% & 1.336 & 1.246\\
FP16 & 0.5001 & 12.5\% & 1.332 & 1.241\\
BF16-FP8 & 0.5208 & 8.0\% & 1.338 & 1.095\\
BF16-FP8-CS & 0.5378 & 4.6\% & 1.335 & 1.101\\
BF16 & 0.5627 &\textit{Ref.} & 1.328 & 1.090\\
\bottomrule
\end{tabular}

\parbox{\columnwidth}{\scriptsize\textit{FP16: fp16\_mixed, BF16: bf16\_mixed, FP16-FP8: fp16\_with\_fp8\_mixed (delayed scaling), FP16-FP8-CS: fp16\_with\_fp8\_current\_scaling\_mixed (tensorwise scaling), BF16-FP8: bf16\_with\_fp8\_mixed (delayed scaling), BF16-FP8-CS: bf16\_with\_fp8\_current\_scaling\_mixed (tensorwise scaling).}}

\end{table}

\subsubsection*{Key Findings:} Key findings from the ablation study:

\textbf{Training Speed}: FP16-based strategies with FP8 hybrid formats achieve the highest speedup (20.5\% and 20.2\% faster than BF16 baseline), reducing step time from 0.5627s to 0.467s. This represents a significant acceleration for large-scale training workloads.

\textbf{Training Stability}: BF16-based strategies demonstrate superior training stability, achieving lower final loss values (1.090--1.101) compared to FP16-based strategies (1.241--1.250). BF16 achieves the lowest final loss (1.090), indicating better numerical stability for gradient updates.

\textbf{Precision Trade-offs}: FP8 hybrid formats provide substantial speed improvements but may introduce slight numerical instability, as evidenced by higher final loss values. The delayed scaling (*-FP8) and tensorwise scaling (*-FP8-CS) recipes show similar performance, with delayed scaling achieving marginally better speedup.

\textbf{Practical Recommendation}: Based on the results, BF16-FP8 provides the best balance between training efficiency and convergence quality. Compared to the BF16 reference, it achieves an 8.0\% faster step time while maintaining a very similar final loss (1.095 vs. 1.090). This makes it a strong candidate as the recommended default configuration for large-scale and long-running CPT training setups.

\subsection{Evaluation Results}

To evaluate embedding model performance on Turkish legal retrieval tasks, we compare our models against a wide range of baselines including state-of-the-art embedding models, traditional BERT-based models, and decoder-to-encoder converted models. Table~\ref{tab:comprehensive_embedding_results} presents detailed performance metrics across all evaluated models on the MTEB-Turkish benchmark, including task-specific scores (Classification, Clustering, Pair Classification, Retrieval, STS), legal domain scores (Legal Score, Contracts, Regulation, Caselaw), model characteristics (Parameters, Model Type), and correlation metrics.

\begin{figure}[htbp]

\centering

\includegraphics[width=0.85\columnwidth]{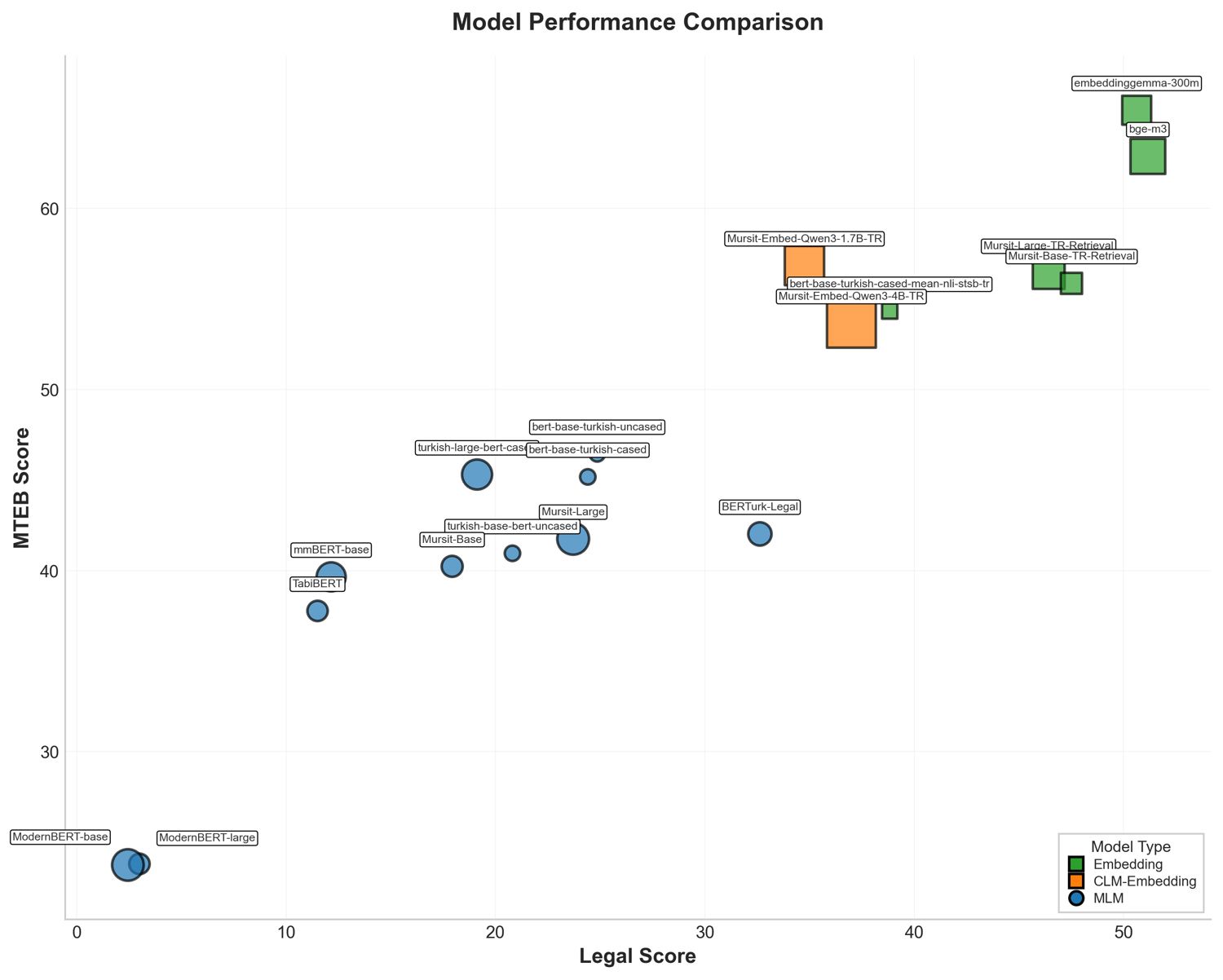}

\caption{Model Performance Comparison: Legal Score vs. MTEB Score.}

\label{fig:model_performance_2d}

\end{figure}

Table~\ref{tab:comprehensive_embedding_results} presents results across three distinct model types: Embedding models (specifically designed for retrieval tasks), CLM-Embedding models (decoder-to-encoder converted), and MLM models (Masked Language Models). While the standard MTEB framework typically focuses on embedding-optimized models and provides limited coverage for MLM architectures, we include MLM baselines to demonstrate baseline performance on downstream tasks, enabling fair comparison of our pre-trained models against traditional BERT-based approaches. Notably, ModernBERT-base~\cite{modernbert2025} and ModernBERT-large~\cite{modernbert2025} are monolingual English models pre-trained exclusively on English data, while all evaluations are conducted on Turkish datasets, providing insight into cross-lingual transfer capabilities and the importance of language-specific pre-training.

EmbeddingGemma-300M~\cite{bgem32024,embeddinggemma2025} (307M parameters), newmindai/bge-m3-stsb (63.53 MTEB), and BAAI/bge-m3 (62.87 MTEB) achieve the highest absolute MTEB scores, benefiting from multi-stage post-training strategies optimized for embedding tasks. These models leverage embedding-specific objectives and multi-phase training pipelines (RetroMAE, contrastive learning, and unified fine-tuning) that better capture Turkish morphological complexity. In contrast, our Mursit-Large-TR-Retrieval (403M parameters) and Mursit-Base-TR-Retrieval (155M parameters) achieve 56.43 and 55.86 MTEB scores respectively. Despite the absolute score gap, our models demonstrate competitive parameter efficiency: Mursit-Base-TR-Retrieval uses less than half the parameters of EmbeddingGemma-300M while achieving comparable performance. Task-specific analysis reveals that retrieval and STS categories show 3--7 point differences between EmbeddingGemma-300M and Mursit models, which becomes negligible when normalized by the 31--50\% parameter advantage of our models. Similarly, while BGE-M3 achieves highest absolute scores in clustering and retrieval tasks, Mursit models produce results within 2--5 points using significantly fewer parameters. These findings demonstrate that Turkish embedding performance depends more on language-specific continual pre-training, retrieval-focused objectives, and task-data alignment than raw model scale alone, with our models achieving competitive parameter-normalized performance compared to larger multi-stage post-trained baselines.

\begin{table*}[htbp]
\caption{Evaluation Results on MTEB-Turkish Benchmark}
\label{tab:comprehensive_embedding_results}
\centering
\tiny
\resizebox{\textwidth}{!}{%
\begin{tabular}{l*{12}{r}}
\toprule
Model & MTEB & Legal & Cls. & Clus. & Pair & Ret. & STS & Cont. & Reg. & Case & Params & Type\\
\midrule
embeddinggemma-300m &\textbf{65.42} & 50.63 &\textbf{77.74} &\textbf{45.05} &\textbf{80.02} &\textbf{55.06} & 69.22 & 83.97 &\textbf{39.56} & 28.38 & 307M & Emb.\\
bge-m3 & 62.87 &\textbf{51.16} & 75.35 & 35.86 & 78.88 & 54.42 &\textbf{69.83} &\textbf{86.08} & 38.09 &\textbf{29.30} & 567M & Emb.\\
\rowcolor{gray!15}Mursit-Embed-Qwen3-1.7B-TR$^*$ & 56.84 & 34.76 & 68.46 & 42.22 & 59.67 & 50.10 & 63.77 & 70.22 & 17.94 & 16.11 & 1.7B & CLM-E.\\
\rowcolor{gray!15}Mursit-Large-TR-Retrieval$^*$ & 56.43 & 46.42 & 67.47 & 38.76 & 59.88 & 51.59 & 64.44 & 81.63 & 32.39 & 25.24 & 403M & Emb.\\
\rowcolor{gray!15}Mursit-Base-TR-Retrieval$^*$ & 55.86 & 47.52 & 66.25 & 39.75 & 61.31 & 50.07 & 61.90 & 80.40 & 34.10 & 28.07 & 155M & Emb.\\
\rowcolor{gray!15}Mursit-Embed-Qwen3-4B-TR$^*$ & 53.65 & 37.00 & 67.29 & 36.68 & 58.36 & 51.12 & 54.77 & 69.25 & 24.21 & 17.56 & 4B & CLM-E.\\
\midrule
bert-base-turkish-uncased &\textbf{46.23} & 24.94 &\textbf{68.05} & 33.81 &\textbf{60.44} &\textbf{32.01} & 36.85 & 52.47 & 12.05 & 10.29 & 110M & MLM\\
turkish-large-bert-cased~\cite{kesgin2023turkishbert} & 45.30 & 19.12 & 67.43 & 34.24 & 60.11 & 28.68 & 36.04 & 47.57 & 5.93 & 3.85 & 337M & MLM\\
bert-base-turkish-cased & 45.17 & 24.41 & 66.39 &\textbf{35.28} & 60.05 & 30.52 & 33.62 & 54.03 & 10.13 & 9.07 & 110M & MLM\\
BERTurk-Legal~\cite{ozturk2023berturklegal} & 42.02 &\textbf{32.63} & 60.61 & 26.24 & 59.51 & 25.80 &\textbf{37.94} &\textbf{61.40} &\textbf{15.51} &\textbf{20.99} & 184M & MLM\\
\rowcolor{gray!15}Mursit-Large$^*$ & 41.75 & 23.71 & 62.95 & 25.34 & 58.04 & 27.40 & 35.01 & 42.74 & 11.29 & 17.10 & 403M & MLM\\
turkish-base-bert-uncased~\cite{kesgin2023turkishbert} & 44.68 & 27.58 & 66.22 & 30.23 & 58.84 & 31.40 & 36.74 & 56.60 & 13.39 & 12.74 & 110M & MLM\\
\rowcolor{gray!15}Mursit-Base$^*$ & 40.23 & 17.93 & 59.78 & 25.48 & 58.65 & 20.82 & 36.45 & 36.00 & 7.40 & 10.40 & 155M & MLM\\
mmBERT-base~\cite{marone2025mmbert} & 39.65 & 12.15 & 61.84 & 26.77 & 59.25 & 15.83 & 34.56 & 34.45 & 1.33 & 0.68 & 306M & MLM\\
TabiBERT~\cite{turker2025tabibert} & 37.77 & 11.50 & 59.63 & 25.75 & 58.19 & 14.96 & 30.32 & 32.02 & 1.86 & 0.63 & 148M & MLM\\
ModernBERT-base~\cite{modernbert2025} & 23.80 & 2.99 & 39.06 & 2.01 & 53.95 & 2.10 & 21.91 & 7.92 & 0.62 & 0.43 & 149M & MLM\\
ModernBERT-large~\cite{modernbert2025} & 23.74 & 2.44 & 39.44 & 3.90 & 53.73 & 1.80 & 19.85 & 6.12 & 0.62 & 0.59 & 394M & MLM\\
\bottomrule
\end{tabular}%

}

{\tiny\textit{Column abbreviations: MTEB = mean performance across task types; Legal = weighted average of Contracts, Regulation, Caselaw; Classification = accuracy on Turkish classification tasks; Clustering = V-measure on clustering tasks; Pair Classification = average precision on pair classification tasks like NLI; Retrieval = nDCG@10 on information retrieval tasks; Semantic Textual Similarity = Spearman correlation; Contracts = nDCG@10 on legal contract retrieval; Regulation = nDCG@10 on regulatory text retrieval; Caselaw = nDCG@10 on case law retrieval; Number of Parameters = number of model parameters; Model Type = model type (Embedding, CLM-Embedding, Masked Language Model).}}

\end{table*}

Figure~\ref{fig:model_performance_2d} visualizes the relationship between Legal Score and MTEB Score across all evaluated models, showing model performance trade-offs. The scatter plot compares 18 models across three architecture types: Embedding (green squares), CLM-Embedding (orange squares), and MLM (blue circles), with bubble size representing model parameter count (logarithmically scaled). The plot reveals distinct performance clusters corresponding to model architectures: Embedding models (green squares) dominate the top-right quadrant, achieving the highest scores on both metrics, with embeddinggemma-300m (Legal: 50.63, MTEB: 65.42), newmindai/bge-m3-stsb (Legal: 49.97, MTEB: 63.53), and BAAI/bge-m3 (Legal: 51.16, MTEB: 62.87) representing the state-of-the-art. Our encoder-only models pre-trained from scratch (Mursit-Base-TR-Retrieval and Mursit-Large-TR-Retrieval) cluster closely with these top performers, indicating that domain-specific pre-training from scratch combined with contrastive learning post-training achieves competitive performance despite significantly fewer parameters. Notably, Mursit-Base-TR-Retrieval (155M parameters) achieves superior Legal Score (47.52) compared to Mursit-Large-TR-Retrieval (46.56), while maintaining comparable MTEB Score (55.86 vs. 56.87), highlighting the parameter efficiency of our base architecture.

CLM-Embedding models (orange squares) exhibit an interesting performance pattern: while Mursit-Embed-Qwen3-1.7B-TR achieves competitive MTEB Score (56.84), it shows substantial underperformance in Legal Score (34.76), indicating that decoder-to-encoder conversion without extensive multi-stage training struggles with domain-specific legal retrieval tasks. This architectural limitation becomes particularly evident in specialized domains where precise semantic understanding and normative reasoning are critical for accurate retrieval performance. The larger Mursit-Embed-Qwen3-4B-TR model shows even greater degradation (Legal: 37.00, MTEB: 53.65), suggesting that increased model capacity alone does not compensate for the architectural mismatch between autoregressive generation and bidirectional embedding tasks.

MLM models (blue circles) form a distinct cluster in the lower-left quadrant, with most models achieving Legal Scores below 25 and MTEB Scores below 45. For MLM models, embeddings are generated using standard mean pooling over all token representations followed by L2 normalization. The pre-trained-only ModernBERT models (base~\cite{marone2025mmbert,modernbert2025} and large~\cite{modernbert2025}), which are monolingual English models pre-trained exclusively on English data, occupy the bottom-left corner (Legal: 2.44--2.99, MTEB: 23.74--23.80) when evaluated on Turkish datasets. Baseline MLM models include Turkish BERT variants (turkish-base-bert-uncased~\cite{kesgin2023turkishbert}, turkish-large-bert-cased~\cite{kesgin2023turkishbert}), BERTurk-Legal~\cite{ozturk2023berturklegal}, mmBERT-base~\cite{marone2025mmbert}, and TabiBERT~\cite{turker2025tabibert}. This shows that pre-training alone without embedding-specific fine-tuning yields minimal utility for retrieval tasks, and further highlights the importance of language-specific pre-training, as monolingual English models show limited cross-lingual transfer to Turkish. Our pre-trained Mursit models (Mursit-Base and Mursit-Large) show substantial improvement over ModernBERT baselines, validating the effectiveness of Turkish-dominant pre-training, though they remain below post-trained embedding models.

The visualization reveals a strong positive correlation between Legal Score and MTEB Score, indicating that models performing well on general Turkish tasks also excel in legal domain tasks. However, the performance gap between model types is more pronounced in Legal Score than MTEB Score, suggesting that legal domain adaptation requires specialized training beyond general-purpose embedding optimization. The bubble sizes, proportional to model parameters, show that parameter count alone does not guarantee superior performance: our 155M Mursit-Base-TR-Retrieval outperforms many larger models, including the 4B Mursit-Embed-Qwen3-4B-TR, emphasizing the importance of architecture selection and training methodology over raw model scale. The evaluation shows that our encoder-only models pre-trained from scratch achieve competitive performance with strong parameter efficiency, ranking among the top models on the Turkish retrieval leaderboard despitely fewer parameters compared to larger baselines. Our Mursit-Base-TR-Retrieval model (155M parameters) achieves 55.86 MTEB Score and 47.52 Legal Score, outperforming many larger models including converted decoder architectures. The decoder-to-encoder converted models (Mursit-Embed-Qwen3-1.7B-TR and Mursit-Embed-Qwen3-4B-TR) demonstrate the significant challenges of adapting autoregressive models for embedding tasks without extensive multi-stage training infrastructure, achieving 56.84 and 53.65 MTEB Scores respectively, with notable underperformance in legal domain tasks compared to encoder-only architectures, highlighting the fundamental architectural advantages of purpose-built embedding models.

\section{Discussion}

\label{sec:discussion}
\vspace{-0.3\baselineskip}
\textbf{Domain Adaptation Effectiveness}: CPT on Turkish legal corpus significantly improves performance across all tasks. Mursit-Base-TR-Retrieval achieved 17.5\% legal domain improvement over checkpoint 2, while Mecellem-Qwen3-4B-TR demonstrated 36.2\% perplexity reduction.

\textbf{Encoder Model Parameter Efficiency:} Our low-parameter encoder models demonstrate exceptional parameter efficiency compared to larger baselines. Mursit-Base-TR-Retrieval (155M parameters) achieves 55.86 MTEB Score and 47.52 Legal Score, while Mursit-Large-TR-Retrieval (403M parameters) achieves 56.43 MTEB Score and 46.42 Legal Score, ranking 2nd and 1st respectively on the Turkish retrieval leaderboard. This demonstrates the effectiveness of domain-specific pre-training from scratch combined with contrastive learning post-training for Turkish legal tasks. Our findings, consistent with prior observations for low-resource and morphologically rich languages \cite{conneau2020xlmr,lester2021prompt,textembedding2024}, indicate that improvements in Masked Language Modeling (MLM) accuracy do not always result in better performance on downstream tasks. While increasing model parameters generally reduces MLM cross-entropy loss, these improvements do not consistently lead to higher-quality embeddings. This suggests that downstream utility depends more on data quality, training convergence dynamics, and the alignment between pre-training objectives and downstream task requirements than on pre-training objective optimization alone, as evidenced by checkpoints achieving better MLM scores but worse retrieval performance. For Turkish, an agglutinative language with complex morphological systems, the optimal checkpoint selection depends on the interplay between morphological complexity, resource availability, and downstream task characteristics, with intermediate checkpoints often outperforming those with minimal MLM loss. Consequently, lower MLM loss is not a guaranteed indicator of embedding effectiveness for morphologically rich languages where linguistic complexity amplifies the divergence between pre-training optimization and downstream utility.

\textbf{Curriculum Learning Benefits}: The four-phase curriculum strategy for Qwen3-1.7B achieved comparable results to single-phase Qwen3-4B despite 2.4$\times$ fewer parameters, validating the importance of staged training from general to specialized content.

\textbf{False-Negative Filtering Critical}: GISTEmbed with guide model outperformed standard InfoNCE by 8.9\% overall (15.9\% legal improvement), demonstrating the importance of preventing false-negative learning in contrastive training.

\textbf{Sequence Length Configuration}: Ablation studies reveal that sequence length optimization based solely on training data distribution leads to significant performance degradation in specialized domains. Models configured with seq\_len=256 (matching MS MARCO-TR token distribution) exhibited 8.5\% degradation in legal domain performance compared to seq\_len=2048, with regulation retrieval showing 23.6\% relative decrease. These findings indicate that sequence length must accommodate target evaluation corpus characteristics for domains requiring long-context understanding.

\textbf{Turkish Quality Filtering Impact:} As a separate experimental investigation (Section~\ref{subsec:turkish_filtering_refinement}), we evaluated the impact of Turkish morphology-based quality filtering (suffix entropy $>75\%$, lemma diversity $>50\%$) applied as post-training refinement. This experiment, conducted independently from the main CPT training pipeline, revealed scale-dependent effects. Qwen3-1.7B showed improvement (5.243 $\rightarrow$ 5.212 perplexity, 0.6\% reduction), while Qwen3-4B exhibited slight degradation (4.884 $\rightarrow$ 4.935, 1.0\% increase). This suggests that smaller models benefit more from aggressive morphological filtering due to limited capacity, while larger models may require greater data diversity once convergence is reached. These experimental results indicate that Turkish Quality Filtering should be applied comprehensively during initial dataset preparation rather than as a post-hoc refinement step.

\textbf{Decoder-to-Encoder Conversion:} The converted Qwen3 CPT models underperform, with Mursit-Embed-Qwen3-4B-TR ranking 9th despite 4B parameters. This demonstrates that decoder-to-encoder conversion requires sophisticated training infrastructure (large-scale synthetic data, multi-stage training, model merging) as described in \cite{qwen3embedding}, which was unavailable in our implementation. The 155M Mursit-Base-TR-Retrieval (Rank 2) outperforms the 4B Mursit-Embed-Qwen3-4B-TR (Rank 9) by 4.1\% in MTEB Score and 28.4\% in Legal Score improvement, indicating that encoder-only architectures pre-trained for embedding tasks are more parameter-efficient than converted decoder models under resource constraints.

\section{Conclusion}

\label{sec:conclusion}

This work demonstrates that effective Turkish legal language models can be developed through two complementary strategies: encoder models pre-trained from scratch with downstream-aware checkpoint selection, and decoder models adapted through controlled continual pre-training. Our key finding is that trainable MLM models serve as effective foundations for embedding tasks when training quality is assessed through downstream performance rather than pre-training loss alone. Our experimental results, consistent with prior observations \cite{conneau2020xlmr,lester2021prompt,textembedding2024}, demonstrate that MLM loss does not fully correlate with downstream task performance for morphologically rich and low-resource languages. The optimal performance range varies based on language structure, resource availability, and downstream task category. For Turkish, an agglutinative language with complex morphological systems, intermediate checkpoints achieve superior retrieval and classification performance compared to checkpoints with minimal MLM loss, suggesting that morphological complexity and domain-specific requirements create distinct optimization landscapes. This finding emphasizes the importance of direct downstream task evaluation rather than relying solely on pre-training metrics, especially for low-resource languages where data scarcity amplifies the divergence between pre-training objectives and downstream utility.

For decoder models, systematic ablation studies of initialization strategies and dataset compositions enable controlled domain adaptation while preserving general capabilities. The resulting models achieve competitive performance on Turkish legal retrieval benchmarks, with our encoder models ranking among the top positions on the Turkish retrieval leaderboard. These contributions advance the state of Turkish legal NLP and provide practical tools for legal information retrieval applications. Future work will explore advanced decoder-to-encoder conversion techniques and develop practical RAG applications tailored for Turkish legal professionals.

All models are released as open-source on HuggingFace under NewmindAI\footnotemark[1].

\vspace{-1.0\baselineskip}
\thispagestyle{conclusionfooter}

\vspace{-0.5\baselineskip}
\section*{Funding}
\vspace{-0.3\baselineskip}

This work was supported by the EuroHPC Joint Undertaking with access to the MareNostrum 5 supercomputer, hosted by Barcelona Supercomputing Center (BSC), Spain. The funding sources had no involvement in the study design, collection, analysis and interpretation of data, writing of the report, or decision to submit the article for publication.

\vspace{-0.5\baselineskip}
\section*{Author Contributions}
\vspace{-0.3\baselineskip}

Author contributions using CRediT (Contributor Roles Taxonomy):
\textbf{Özgür Uğur}: Conceptualization, Methodology, Project administration, Writing - Review \& Editing; \textbf{Mahmut Göksu}: Methodology, Software, Writing - Original Draft; \textbf{Mahmut Çimen}: Methodology, Data curation, Software, Writing - Original Draft; \textbf{Musa Yılmaz}: Methodology, Software, Writing - Original Draft; \textbf{Alp Talha Demir}: Validation, Software, Writing - Original Draft; \textbf{Esra Savirdi}: Data curation, Methodology, Writing - Original Draft; \textbf{İclal Çetin}: Data curation, Writing - Original Draft; \textbf{Ömer Can Sağbaş}: Data curation, Writing - Original Draft; \textbf{Rumeysa Güllüce}: Investigation, Writing - Original Draft.

\vspace{-0.5\baselineskip}
\section*{Data Availability}
\vspace{-0.3\baselineskip}

The datasets, models, code, and benchmarks used in this study are publicly available as follows: Pre-trained and post-trained models are available through the HuggingFace model collection\footnotemark[1]. Legal benchmark datasets used for evaluation (Contracts, Regulation, Caselaw) are available through the HuggingFace collection\footnotemark[2]. The complete model collection, training code, evaluation scripts, and benchmark datasets including all datasets generated for benchmark evaluation are available in the repository\footnotemark[3]. Legal corpus data sources include publicly available court decisions and academic theses; specific datasets are described in Section~\ref{sec:methodology}.

\footnotetext[1]{\url{https://huggingface.co/collections/newmindai/mecellem-models}}
\footnotetext[2]{\url{https://huggingface.co/collections/newmindai/mleb-dataset}}
\footnotetext[3]{\url{https://github.com/newmindai/mecellem-models}}

\vspace{-0.5\baselineskip}
\section*{Acknowledgments}
\vspace{-0.3\baselineskip}

The numerical calculations reported in this paper were fully/partially performed using the EuroHPC Joint Undertaking (EuroHPC JU) supercomputer MareNostrum 5, hosted by the Barcelona Supercomputing Center (BSC). Access to MareNostrum 5 was provided through a national access call coordinated by the Scientific and Technological Research Council of Turkey (T\"{U}B\.{I}TAK). We gratefully acknowledge T\"{U}B\.{I}TAK, BSC and the EuroHPC JU for providing access to these resources and supporting this research. The authors gratefully acknowledge the know-how provided by the MINERVA Support for expert guidance and collaboration opportunities in HPC-AI integration.


\bibliographystyle{elsarticle-num}

\bibliography{references}

@article{turkish2024,
  title={Bridging the Bosphorus: Advancing Turkish Large Language Models through Strategies for Low-Resource Language Adaptation and Benchmarking},
  author={Acikgoz, Emre Can and Erdogan, Mete and Yuret, Deniz},
  journal={arXiv preprint arXiv:2405.04685},
  year={2024},
  eprint={2405.04685},
  archivePrefix={arXiv},
  primaryClass={cs.CL}
}

@article{emergent2022,
  title={Emergent abilities of large language models},
  author={Wei, Jason and Tay, Yi and Bommasani, Rishi and others},
  journal={arXiv preprint arXiv:2206.07682},
  year={2022},
  eprint={2206.07682},
  archivePrefix={arXiv},
  primaryClass={cs.CL}
}

@article{gpt4,
  title={GPT-4 technical report},
  author={Achiam, Josh and Adler, Steven and Agarwal, Sandhini and others},
  journal={arXiv preprint arXiv:2303.08774},
  year={2023},
  eprint={2303.08774},
  archivePrefix={arXiv},
  primaryClass={cs.CL}
}

@article{legalai2020,
  title={How does NLP benefit legal system: A summary of legal artificial intelligence},
  author={Zhong, Haoxi and Xiao, Chaojun and Tu, Cunchao and Zhang, Tianyang and Liu, Zhiyuan and Sun, Maosong},
  journal={arXiv preprint arXiv:2004.12158},
  year={2020},
  eprint={2004.12158},
  archivePrefix={arXiv},
  primaryClass={cs.CL}
}

@article{legalbert2020,
  title={LEGAL-BERT: The muppets straight out of law school},
  author={Chalkidis, Ilias and Fergadiotis, Manos and Malakasiotis, Prodromos and Aletras, Nikolaos and Androutsopoulos, Ion},
  journal={arXiv preprint arXiv:2010.02559},
  year={2020},
  eprint={2010.02559},
  archivePrefix={arXiv},
  primaryClass={cs.CL}
}

@misc{fineweb2024,
  title={FineWeb: A Large-Scale Web Dataset for Language Model Training},
  author={{HuggingFace}},
  howpublished={GitHub Repository},
  year={2024},
  note={\url{github.com/huggingface/datatrove}}
}

@misc{semhash2024,
  title={SemHash: Semantic Hash-based Deduplication for Large Language Models},
  author={{MinishLab}},
  howpublished={GitHub Repository},
  year={2024},
  note={\url{github.com/MinishLab/semhash}}
}

@article{butler2025massivelegalembeddingbenchmark,
  title={The Massive Legal Embedding Benchmark (MLEB)},
  author={Butler, Umar and Butler, Abdur-Rahman and Malec, Adrian Lucas},
  journal={arXiv preprint arXiv:2510.19365},
  year={2025},
  eprint={2510.19365},
  archivePrefix={arXiv},
  primaryClass={cs.CL},
  note={Developed by Isaacus}
}

@misc{turkishbert2024,
  title={Turkish BERT/DistilBERT, ELECTRA, ConvBERT and T5 models},
  author={{Stefan Schweter}},
  year={2024},
  howpublished={GitHub repository}
}

@article{vbart2024,
  title={VBART: The Turkish LLM},
  author={Turker, M. and Ari, M. E. and Han, A.},
  journal={arXiv preprint arXiv:2403.01308},
  year={2024},
  eprint={2403.01308},
  archivePrefix={arXiv},
  primaryClass={cs.CL}
}

@article{AITEAM2025124,
  title={Tailoring AI for Turkish Law: Domain-Specific Fine-Tuning of Small Language Models for Legal Expertise},
  author={New Mind AI Team},
  journal={Procedia Computer Science},
  volume={267},
  pages={124--135},
  year={2025},
  doi={10.1016/j.procs.2025.08.239}
}

@article{ulmfit2018,
  title={Universal language model fine-tuning for text classification},
  author={Howard, Jeremy and Ruder, Sebastian},
  journal={arXiv preprint arXiv:1801.06146},
  year={2018},
  eprint={1801.06146},
  archivePrefix={arXiv},
  primaryClass={cs.CL}
}

@inproceedings{modernbert2025,
  title={Smarter, better, faster, longer: A modern bidirectional encoder for fast, memory efficient, and long context finetuning and inference},
  author={Warner, Benjamin and Chaffin, Antoine and Clavi{\'e}, Benjamin and Weller, Orion and Hallstr{\"o}m, Oskar and Taghadouini, Said and Gallagher, Alexis and Biswas, Rohan and Ladhak, Faisal and Aarsen, Tyler and Adams, Garrett T.},
  booktitle={Proceedings of the 63rd Annual Meeting of the Association for Computational Linguistics (Volume 1: Long Papers)},
  pages={2526--2547},
  year={2025}
}

@article{qwen3embedding,
  title={Qwen3 Embedding: Ultra-Large Multilingual Embedding Model},
  author={Qwen Team},
  journal={arXiv preprint arXiv:2506.05176},
  year={2025},
  eprint={2506.05176},
  archivePrefix={arXiv},
  primaryClass={cs.CL}
}

@article{ewc2017,
  title={Overcoming catastrophic forgetting in neural networks},
  author={Kirkpatrick, James and Pascanu, Razvan and Rabinowitz, Neil and others},
  journal={Proceedings of the National Academy of Sciences},
  volume={114},
  number={13},
  pages={3521--3526},
  year={2017}
}

@article{nayak2025sculpting,
  title={Sculpting Subspaces: Constrained Full Fine-Tuning in LLMs for Continual Learning},
  author={Nayak, Nikhil Shivakumar and Killamsetty, Krishnateja and Han, Ligong and Bhandwaldar, Abhishek and Chanda, Prateek and Xu, Kai and Wang, Hao and Pareja, Aldo and Silkin, Oleg and Eyceoz, Mustafa and Srivastava, Akash},
  journal={arXiv preprint arXiv:2504.07097},
  year={2025},
  eprint={2504.07097},
  archivePrefix={arXiv},
  primaryClass={cs.CL}
}

@misc{catastrophicforgetting2023,
  title={The Elephant in the Room: Catastrophic Forgetting},
  author={{AI Innovation Team}},
  year={2023},
  howpublished={AI Innovation Team Blog}
}

@article{climb2025,
  title={CLIMB: CLustering-based Iterative Data Mixture Bootstrapping for Language Model Pre-training},
  author={Diao, Shizhe and Yang, Yiming and Fu, Yuli and others},
  journal={arXiv preprint arXiv:2504.13161},
  year={2025},
  eprint={2504.13161},
  archivePrefix={arXiv},
  primaryClass={cs.CL}
}

@article{weborganizer2025,
  title={Organize the Web: Constructing Domains Enhances Pre-Training Data Curation},
  author={Wettig, Alexander and Lo, Kyle and Min, Sewon and others},
  journal={arXiv preprint arXiv:2502.10341},
  year={2025},
  eprint={2502.10341},
  archivePrefix={arXiv},
  primaryClass={cs.CL}
}

@article{fineweb22025,
  title={FineWeb2: One Pipeline to Scale Them All--Adapting Pre-Training Data Processing to Every Language},
  author={Penedo, Guilherme and Kydl{\'i}{\v{c}}ek, Hynek and Sabol{\v{c}}ec, Vinko and Messmer, Bettina and Foroutan, Negar and Kargaran, Amir Hossein and Raffel, Colin and Jaggi, Martin and Von Werra, Leandro and Wolf, Thomas},
  journal={arXiv preprint arXiv:2506.20920},
  year={2025},
  eprint={2506.20920},
  archivePrefix={arXiv},
  primaryClass={cs.CL}
}

@article{glotlid2023,
  title={GlotLID: Language identification for low-resource languages},
  author={Kargaran, Amir Hossein and Imani, Ayyoob and Yvon, Fran{\c{c}}ois and Sch{\"u}tze, Hinrich},
  journal={arXiv preprint arXiv:2310.16248},
  year={2023},
  eprint={2310.16248},
  archivePrefix={arXiv},
  primaryClass={cs.CL}
}

@article{Akin2007Zemberek,
  title   = {Zemberek, an Open Source NLP Framework for Turkic Languages},
  author  = {Ak{\i}n, Ahmet Af{\c{s}}{\i}n and Ak{\i}n, Mehmet D{\"u}ndar},
  journal = {Structure},
  volume  = {10},
  number  = {2007},
  pages   = {1--5},
  year    = {2007}
}

@article{Shannon1948,
  title={A mathematical theory of communication},
  author={Shannon, Claude E.},
  journal={Bell System Technical Journal},
  volume={27},
  number={3},
  pages={379--423},
  year={1948}
}

@article{nemotronccmath2025,
  title={Nemotron-cc-math: A 133 billion-token-scale high quality math pretraining dataset},
  author={Mahabadi, Rabeeh Karimi and Satheesh, Surya and Prabhumoye, Shrimai and Patwary, Mostofa and Shoeybi, Mohammad and Catanzaro, Bryan},
  journal={arXiv preprint arXiv:2508.15096},
  year={2025},
  eprint={2508.15096},
  archivePrefix={arXiv},
  primaryClass={cs.CL}
}

@article{openwebmath2025,
  title={Openwebmath: An open dataset of high-quality mathematical web text (2023)},
  author={Paster, Keiran and Dos Santos, Marco and Azerbayev, Zhangir and Ba, Jimmy},
  journal={arXiv preprint arXiv:2310.06786},
  year={2025},
  eprint={2310.06786},
  archivePrefix={arXiv},
  primaryClass={cs.CL}
}

@article{starcoder2023,
  title={Starcoder: may the source be with you!},
  author={Li, Raymond and Allal, Leandro B. and Zi, Yangtian and Muennighoff, Nikolai and Kocetkov, Denis and Mou, Chenghao and de Vries, Harm},
  journal={arXiv preprint arXiv:2305.06161},
  year={2023},
  eprint={2305.06161},
  archivePrefix={arXiv},
  primaryClass={cs.SE}
}

@article{bayram2025tokenizationstandardslinguisticintegrity,
    title={Tokenization Standards for Linguistic Integrity: Turkish as a Benchmark}, 
    author={Bayram, M. Ali and Fincan, Ali Arda and Gümüş, Ahmet Semih and Karakaş, Sercan and Diri, Banu and Yıldırım, Savaş},
    journal={arXiv preprint arXiv:2502.07057},
    year={2025},
    eprint={2502.07057},
    archivePrefix={arXiv},
    primaryClass={cs.CL}
}

@inproceedings{eryigit2014itu,
    title={ITU Turkish NLP Web Service},
    author={Eryiğit, Gülşen},
    booktitle={Proceedings of the Demonstrations at the 14th Conference of the European Chapter of the Association for Computational Linguistics (EACL 2014)},
    year={2014},
    pages={1--4},
    address={Gothenburg, Sweden},
    month={April},
    publisher={Association for Computational Linguistics}
}

@article{bgem32024,
  title={BGE M3-Embedding: Multi-Lingual, Multi-Functionality, Multi-Granularity Text Embeddings Through Self-Knowledge Distillation},
  author={Chen, Jianlv and Xiao, Shitao and Zhang, Peitian and Luo, Kun and Lian, Defu and Liu, Zheng},
  journal={arXiv preprint arXiv:2402.03216},
  year={2024},
  eprint={2402.03216},
  archivePrefix={arXiv},
  primaryClass={cs.CL}
}

@misc{mosaicmlcomposer2021,
  author={{The Mosaic ML Team}},
  title={composer},
  year={2021},
  howpublished={\url{github.com/mosaicml/composer}}
}

@article{gistembed2024,
  title={GISTEmbed: Guided In-sample Selection of Training Negatives for Text Embedding Fine-tuning},
  author={Solatorio, Aivin V. and Dupont, Olivier},
  journal={arXiv preprint arXiv:2402.16829},
  year={2024},
  eprint={2402.16829},
  archivePrefix={arXiv},
  primaryClass={cs.CL}
}

@article{embeddinggemma2025,
  title={Embeddinggemma: Powerful and lightweight text representations},
  author={Vera, Henrique Schechter and Dua, Shreya and Zhang, Bo and Salz, David and Mullins, Ryan and Panyam, Surya R. and others},
  journal={arXiv preprint arXiv:2509.20354},
  year={2025},
  eprint={2509.20354},
  archivePrefix={arXiv},
  primaryClass={cs.CL}
}

@article{qwen2024,
  title={Qwen Technical Report},
  author={Yang, An and Yang, Baosong and Hui, Binyuan and others},
  journal={arXiv preprint arXiv:2309.16609},
  year={2024},
  eprint={2309.16609},
  archivePrefix={arXiv},
  primaryClass={cs.CL}
}

@misc{nemo2024,
  title={NVIDIA NeMo: A Toolkit for Building AI Applications Using Neural Modules},
  author={{NVIDIA Corporation}},
  year={2024},
  howpublished={Open-source toolkit for conversational AI and LLM training}
}

@article{megatronlm2020,
  title={Megatron-LM: Training Multi-Billion Parameter Language Models Using Model Parallelism},
  author={Shoeybi, Mohammad and Patwary, Mostofa and Puri, Raul and LeGresley, Patrick and Casper, Jared and Catanzaro, Bryan},
  journal={arXiv preprint arXiv:1909.08053},
  year={2020},
  eprint={1909.08053},
  archivePrefix={arXiv},
  primaryClass={cs.LG}
}

@article{conneau2020xlmr,
  title={Unsupervised Cross-lingual Representation Learning at Scale},
  author={Conneau, Alexis and Khandelwal, Kartikay and Goyal, Naman and Chaudhary, Vishrav and Wenzek, Guillaume and Guzm{\'a}n, Francisco and Grave, Edouard and Ott, Myle and Zettlemoyer, Luke and Stoyanov, Veselin},
  journal={arXiv preprint arXiv:1911.02116},
  year={2020},
  eprint={1911.02116},
  archivePrefix={arXiv},
  primaryClass={cs.CL}
}

@article{lester2021prompt,
  title={The Power of Scale for Parameter-Efficient Prompt Tuning},
  author={Lester, Brian and Al-Rfou, Rami and Constant, Noah},
  journal={arXiv preprint arXiv:2104.08691},
  year={2021},
  eprint={2104.08691},
  archivePrefix={arXiv},
  primaryClass={cs.CL}
}

@article{textembedding2024,
  title={Improving text embeddings with large language models},
  author={Wang, Liang and Yang, Nan and Huang, Xiaolong and others},
  journal={arXiv preprint arXiv:2401.00368},
  year={2024},
  eprint={2401.00368},
  archivePrefix={arXiv},
  primaryClass={cs.CL}
}

@article{dotsocr2025,
  title={dots. ocr: Multilingual document layout parsing in a single vision-language model},
  author={Li, Yumeng and Yang, Guoxin and Liu, Haoyu and Wang, Bo and Zhang, Changshui},
  journal={arXiv preprint arXiv:2512.02498},
  year={2025},
  eprint={2512.02498},
  archivePrefix={arXiv},
  primaryClass={cs.CV}
}

@inproceedings{devlin2019bert,
  title={BERT: Pre-training of Deep Bidirectional Transformers for Language Understanding},
  author={Devlin, Jacob and Chang, Ming-Wei and Lee, Kenton and Toutanova, Kristina},
  booktitle={Proceedings of the 2019 Conference of the North American Chapter of the Association for Computational Linguistics: Human Language Technologies, Volume 1 (Long and Short Papers)},
  pages={4171--4186},
  year={2019},
  publisher={Association for Computational Linguistics}
}

@article{liu2019roberta,
  title={RoBERTa: A Robustly Optimized BERT Pretraining Approach},
  author={Liu, Yinhan and Ott, Myle and Goyal, Naman and Du, Jingfei and Joshi, Mandar and Chen, Danqi and Levy, Omer and Lewis, Mike and Zettlemoyer, Luke and Stoyanov, Veselin},
  journal={arXiv preprint arXiv:1907.11692},
  year={2019},
  eprint={1907.11692},
  archivePrefix={arXiv},
  primaryClass={cs.CL}
}

@article{kesgin2023turkishbert,
  title={Developing and evaluating tiny to medium-sized Turkish BERT models},
  author={Kesgin, Himmet Toprak and Yuce, Muzaffer Kaan and Amasyali, Mehmet Fatih},
  journal={arXiv preprint arXiv:2307.14134},
  year={2023},
  eprint={2307.14134},
  archivePrefix={arXiv},
  primaryClass={cs.CL}
}

@article{turker2025tabibert,
  title={TabiBERT: A Large-Scale ModernBERT Foundation Model and Unified Benchmarking Framework for Turkish},
  author={T{\"u}rker, Melik{\c{s}}ah and K{\i}z{\i}lo{\u{g}}lu, A. Ebrar and G{\"u}ng{\"o}r, Onur and {\"U}sk{\"u}darl{\i}, Susan},
  journal={arXiv preprint arXiv:2512.23065},
  year={2025},
  eprint={2512.23065},
  archivePrefix={arXiv},
  primaryClass={cs.CL}
}

@inproceedings{ozturk2023berturklegal,
  title={A Transformer-Based Prior Legal Case Retrieval Method},
  author={{\"O}zt{\"u}rk, Ceyhun E. and {\"O}z{\c{c}}el{\i}K, {\c{S}}. Bar{\i}{\c{s}} and Ko{\c{c}}, Aykut},
  booktitle={2023 31st Signal Processing and Communications Applications Conference (SIU)},
  pages={1--4},
  year={2023},
  organization={IEEE}
}

@article{liu2025skywork_reward_v2,
  title   = {Skywork-Reward-V2: Scaling Preference Data Curation via Human-AI Synergy},
  author  = {Liu, Chenyu and Zeng, Liang and Xiao, Yifan and He, Jing and Liu, Junjie and Wang, Chen and Yan, Rui and Shen, Wei and Zhang, Fan and Xu, Jun and Liu, Yang and Zhou, Yu},
  journal = {arXiv preprint arXiv:2507.01352},
  year    = {2025},
  eprint={2507.01352},
  archivePrefix={arXiv},
  primaryClass={cs.CL}
}

@article{llama3_2024,
  title   = {The Llama 3 Herd of Models},
  author  = {{Llama Team} and Meta AI},
  journal = {arXiv preprint arXiv:2407.21783},
  year    = {2024},
  eprint={2407.21783},
  archivePrefix={arXiv},
  primaryClass={cs.CL}
}

@article{ArmoRM,
  title={Interpretable Preferences via Multi-Objective Reward Modeling and Mixture-of-Experts},
  author={Wang, Haoxiang and Xiong, Wei and Xie, Tengyang and Zhao, Han and Zhang, Tong},
  journal={arXiv preprint arXiv:2406.12845},
  year={2024},
  eprint={2406.12845},
  archivePrefix={arXiv},
  primaryClass={cs.CL}
}

@article{marone2025mmbert,
  title={mmbert: A modern multilingual encoder with annealed language learning},
  author={Marone, Marc and Weller, Orion and Fleshman, William and Yang, Eugene and Lawrie, Dawn and Van Durme, Benjamin},
  journal={arXiv preprint arXiv:2509.06888},
  year={2025},
  eprint={2509.06888},
  archivePrefix={arXiv},
  primaryClass={cs.CL}
}

@article{c4dataset2019,
  title={Exploring the Limits of Transfer Learning with a Unified Text-to-Text Transformer},
  author={Raffel, Colin and Shazeer, Noam and Roberts, Adam and Lee, Katherine and Narang, Sharan and Matena, Michael and Zhou, Yanqi and Li, Wei and Liu, Peter J.},
  journal={Journal of Machine Learning Research},
  volume={21},
  pages={1--67},
  year={2020}
}

@article{gopher2021,
  title={Scaling Language Models: Methods, Analysis \& Insights from Training Gopher},
  author={Rae, Jack W. and Borgeaud, Sebastian and Cai, Trevor and Millican, Katie and Hoffmann, Jordan and Song, Francis and Aslanides, John and Henderson, Sarah and Ring, Roman and Young, Susannah and others},
  journal={arXiv preprint arXiv:2112.11446},
  year={2021},
  eprint={2112.11446},
  archivePrefix={arXiv},
  primaryClass={cs.CL}
}

\appendix

\pagestyle{resultsreproduction}

\section{Results Reproduction}\label{app:results_reproduction}

\subsection{MLM Benchmark Results}\label{app:mlm_benchmark_results}

The MLM accuracy scores presented in Table~\ref{tab:mlm_accuracy} can be reproduced using the repository\footnote{\href{https://github.com/newmindai/mecellem-models/benchmark/mlm}{github.com/newmindai/mecellem-models/benchmark/mlm}}, which contains all necessary code, evaluation configurations, and instructions for reproducing the MLM benchmark results across all evaluated models on Turkish datasets using the 80-10-10 masking strategy.

\subsection{Evaluation Results}\label{app:comprehensive_evaluation_results}

The evaluation results presented in Table~\ref{tab:comprehensive_embedding_results} can be reproduced using the repository\footnote{\href{https://github.com/newmindai/mecellem-models/benchmark/embedding_model}{github.com/newmindai/mecellem-models/benchmark/embedding\_model}}, which contains all necessary code, evaluation configurations, and instructions for reproducing the benchmark results across all 17 models evaluated on the MTEB-Turkish benchmark.

\subsection{Post-Training Benchmark Results}\label{app:post_training_benchmark_results}

This appendix presents detailed performance comparisons of post-trained retrieval models across all evaluation tasks. All models were fine-tuned using contrastive learning on the MS MARCO-TR dataset following the methodology described in Section~\ref{subsec:encoder_posttraining}. The results demonstrate the effectiveness of post-training for improving retrieval performance on Turkish legal tasks.

\begin{figure}[htbp]
\centering
\includegraphics[width=0.85\columnwidth]{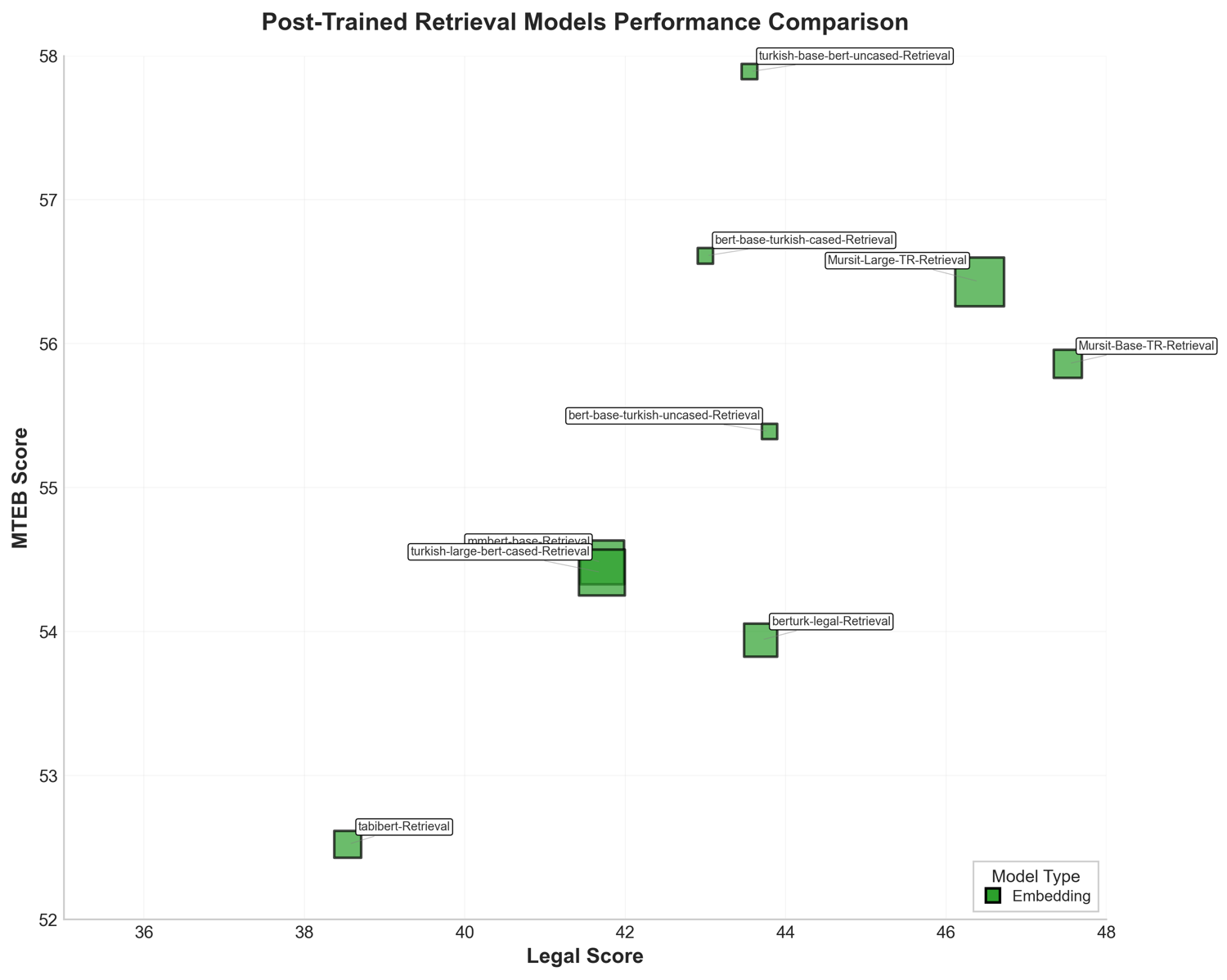}
\setlength{\belowcaptionskip}{-5pt}
\caption{Post-Trained Retrieval Models Performance Comparison: Legal Score vs. MTEB Score.}
\label{fig:post_train_retrieval_2d}
\end{figure}

Table~\ref{tab:post_training_results} presents comprehensive performance metrics for post-trained models, comparing our Mursit models against established monolingual Turkish baselines and multilingual reference models. Our Mursit-Base-TR-Retrieval achieves the highest Legal Score (47.52) among all evaluated models, while Mursit-Large-TR-Retrieval achieves the highest MTEB Score (56.43). The results show consistent improvements across all task types, with strong performance in legal domain retrieval tasks (Contracts, Regulation, Caselaw). Legal Score represents the weighted average of Contracts, Regulation, and Caselaw scores. MTEB Score is the mean performance across task types (Classification, Clustering, Pair Classification, Retrieval, STS). Classification reports accuracy on Turkish classification tasks, Clustering reports V-measure on clustering tasks, Pair Classification reports average precision on pair classification tasks like NLI, Retrieval reports nDCG@10 on information retrieval tasks, and STS reports Spearman correlation on semantic similarity tasks. Contracts, Regulation, and Caselaw report nDCG@10 on their respective legal domain retrieval tasks. Figure~\ref{fig:post_train_retrieval_2d} visualizes the relationship between Legal Score and MTEB Score for the models presented in this table. The models presented in Table~\ref{tab:post_training_results} and Table~\ref{tab:comprehensive_embedding_results} can be evaluated and trained using the repository\footnote{\href{https://github.com/newmindai/mecellem-models/benchmark/embedding_model}{github.com/newmindai/mecellem-models/benchmark/embedding\_model}} and\footnote{\href{https://github.com/newmindai/mecellem-models/training/post-training-retrieval}{github.com/newmindai/mecellem-models/training/post-training-retrieval}}.

Beyond retrieval tasks, we also evaluated our models on non-retrieval downstream tasks to assess their general language understanding capabilities. Table~\ref{tab:non_retrieval_tasks} presents average performance across task types (classification, named entity recognition, question answering, semantic textual similarity, natural language inference, and part-of-speech tagging) for TabiBERT, Mursit-Large, and Mursit-Base. Performance is averaged across datasets within each task type. Overall represents the mean across all task types. CLF (Classification) reports F1-macro averaged across 6 datasets; NER (Named Entity Recognition) reports F1-micro averaged across 2 datasets; QA (Question Answering) reports F1-score averaged across 2 datasets; STS (Semantic Textual Similarity) reports Pearson correlation on STS dataset; NLI (Natural Language Inference) reports F1-macro on MedNLI dataset; POS (Part-of-Speech Tagging) reports F1-micro averaged across 2 datasets. Hyperparameters were tuned following the recommendations in the reference implementation. Bold values indicate the highest score in each column. Models are sorted by Overall score in descending order.

\begin{table}[htbp]
\caption{Average Performance Across Task Types (Non-Retrieval Tasks)}
\label{tab:non_retrieval_tasks}
\centering
\footnotesize
\resizebox{0.9\columnwidth}{!}{%
\begin{tabular}{l*{7}{c}}
\toprule
Model & Overall & CLF & NER & QA & STS & NLI & POS\\
\midrule
Mursit-Large &\textbf{0.818} &\textbf{0.748} &\textbf{0.848} &\textbf{0.685} &\textbf{0.862} &\textbf{0.843} &\textbf{0.921}\\
Mursit-Base & 0.798 & 0.732 & 0.835 & 0.632 & 0.849 & 0.826 & 0.913\\
TabiBERT & 0.772 & 0.722 & 0.844 & 0.492 & 0.850 & 0.816 & 0.910\\
\bottomrule
\end{tabular}%

}

\end{table}

The results demonstrate that our Mursit models achieve competitive or superior performance compared to TabiBERT across all task types, with Mursit-Large showing the strongest overall performance. Notably, Mursit models maintain strong performance across diverse task types, indicating robust language understanding capabilities beyond retrieval-specific optimizations. All experiments are reproducible at the repository\footnote{\href{https://github.com/newmindai/mecellem-models/benchmark/mlm}{github.com/newmindai/mecellem-models/benchmark/mlm}}.

\begin{table*}[htbp]
\caption{Post-Trained Retrieval Models Performance Comparison}
\label{tab:post_training_results}
\centering
\small
\resizebox{\textwidth}{!}{%
\begin{tabular}{l*{10}{r}}
\toprule
Model & Legal & MTEB & Cls. & Clus. & Pair & Ret. & STS & Cont. & Reg. & Case\\
\midrule
\rowcolor{gray!15}Mursit-Base-TR-Retrieval$^*$ (NewmindAI) &\textbf{47.52} & 55.86 & 66.25 & 39.75 & 61.31 & 50.07 & 61.90 & 80.40 &\textbf{34.10} &\textbf{28.07}\\
\rowcolor{gray!15}Mursit-Large-TR-Retrieval$^*$ (NewmindAI) & 46.42 & 56.43 & 67.47 & 38.76 & 59.88 &\textbf{51.59} &\textbf{64.44} &\textbf{81.63} & 32.39 & 25.24\\
bert-base-turkish-uncased-Retrieval$^*$ (dbmdz) & 43.80 & 55.39 & 69.05 & 41.52 & 61.39 & 50.20 & 54.77 & 80.40 & 29.14 & 21.85\\
berturk-legal-Retrieval (KocLab-Bilkent) & 43.69 & 53.94 & 64.95 & 39.77 & 58.81 & 48.64 & 57.54 & 79.80 & 26.41 & 24.86\\
turkish-base-bert-uncased-Retrieval$^*$ (YTU-CE-Cosmos) & 43.55 &\textbf{57.89} &\textbf{72.29} &\textbf{43.23} &\textbf{61.77} & 47.95 & 64.18 & 80.94 & 28.84 & 20.86\\
bert-base-turkish-cased-Retrieval (dbmdz) & 43.00 & 56.61 & 69.45 & 42.25 & 60.95 & 49.70 & 60.71 & 78.93 & 28.03 & 22.04\\
mmbert-base-Retrieval (JHU-CLSP) & 41.71 & 54.48 & 65.16 & 33.24 & 61.51 & 48.93 & 63.56 & 76.19 & 27.76 & 21.19\\
turkish-large-bert-cased-Retrieval (YTU-CE-Cosmos) & 41.71 & 54.41 & 68.13 & 41.50 & 60.77 & 48.23 & 53.40 & 72.70 & 29.64 & 22.79\\
tabibert-Retrieval (BOUN-TabiLab) & 38.54 & 52.52 & 65.36 & 36.30 & 60.10 & 47.68 & 53.17 & 73.65 & 24.30 & 17.68\\
\bottomrule
\end{tabular}%

}

\parbox{\textwidth}{\scriptsize\textit{Column abbreviations: Legal = Legal Score; MTEB = MTEB Score; Cls. = Classification; Clus. = Clustering; Pair = Pair Classification; Ret. = Retrieval; STS = Semantic Textual Similarity; Cont. = Contracts; Reg. = Regulation; Case = Caselaw. Models are sorted by Legal Score in descending order.}}

\end{table*}

\vspace{1.0\baselineskip}
\subsubsection{Production Deployment Efficiency}\label{subsec:production_deployment}
\vspace{0.3\baselineskip}

\begin{table}[!t]
\caption{Production-Ready Evaluation: Embedding Model Efficiency for Vector Database Deployment}
\label{tab:production_efficiency}
\centering
\setlength{\belowcaptionskip}{-8pt}
\normalsize
\resizebox{\columnwidth}{!}{%
\begin{tabular}{l*{7}{r}}
\toprule
Model & Legal & Ret. & Max Seq Length & Params & Emb. Dim & Avg Time & Efficiency \\
\midrule
embeddinggemma-300m & 50.63 & 55.06 & 2,048 & 307M & 768 & 1.41 & 100.00 \\
bge-m3 & 51.16 & 54.42 & 8,194 & 567M & 1,024 & 1.49 & 99.54 \\
\textbf{newmindai/bge-m3-stsb} & 49.97 & 50.14 & 8,194 & 567M & 1,024 & 1.37 & 94.38 \\
\textbf{newmindai/Mursit-Base-TR-Retrieval} & 47.52 & 50.07 & 1,024 & 155M & 768 & 1.17 & 92.36 \\
\textbf{newmindai/Mursit-Large-TR-Retrieval} & 46.42 & 51.59 & 2,048 & 403M & 1,024 & 1.67 & 91.26 \\
bert-base-turkish-uncased-Retrieval & 43.80 & 50.20 & 512 & 110M & 768 & 0.84 & 88.84 \\
bert-base-turkish-cased-Retrieval & 43.00 & 49.70 & 512 & 110M & 768 & 0.84 & 87.55 \\
berturk-legal-Retrieval & 43.69 & 48.64 & 512 & 184M & 768 & 0.72 & 87.21 \\
turkish-base-bert-uncased-Retrieval & 43.55 & 47.95 & 512 & 110M & 768 & 1.16 & 86.31 \\
mmbert-base-Retrieval & 41.71 & 48.93 & 8,192 & 306M & 768 & 1.53 & 86.01 \\
\bottomrule
\end{tabular}%
}
\parbox{\columnwidth}{\scriptsize\textit{Column abbreviations: Legal = Legal Score; Ret. = Retrieval (nDCG@10); Max Seq Length = Maximum Sequence Length; Params = Number of Parameters; Emb. Dim = Embedding Dimension; Avg Time = Average Ingestion Time per 100 Points (s); Efficiency = Production Efficiency (normalized across 25 SOTA models).}}
\vspace{-0.3\baselineskip}
\end{table}
\vspace{0.5\baselineskip}
\noindent To evaluate production deployment efficiency, we introduce a composite Production Efficiency metric that balances retrieval performance, legal domain specialization, and computational resource requirements. This comprehensive framework addresses critical real-world deployment challenges by systematically quantifying the complex interplay between model accuracy, operational scalability, and infrastructure costs. Higher scores indicate better trade-offs between performance and deployment costs. The metric combines normalized scores with the following weights: 40\% for Retrieval performance, 40\% for Legal domain score, -5\% penalty for Parameter count, -5\% penalty for Average Ingestion Time per 100 Points, +5\% bonus for Maximum Sequence Length (if > 512) or -5\% penalty (if $\leq$ 512), and +5\% bonus for Embedding Dimension (if < 2048) or -5\% penalty (if $\geq$ 2048). Normalization is performed across 25 state-of-the-art embedding models to ensure fair comparison and meaningful relative performance assessment.

\noindent We measure vector ingestion latency by recording the time required to insert 100 vectors into a Qdrant database under production settings, including INT8 scalar quantization (quantile=0.98), on-disk HNSW indexing (m=16, ef\_construct=100), and WAL-enabled persistence. All ingestion experiments use a fixed batch size of 100 vectors, with embedding inference time excluded to isolate database insertion performance. This framework enables practitioners to make decisions when selecting models for legal RAG systems, where accuracy and efficiency impact scalability and cost-effectiveness. By quantifying trade-offs between model performance and infrastructure requirements, the Production Efficiency metric addresses a critical gap in embedding model evaluation, providing insights for deployment scenarios.

Table~\ref{tab:production_efficiency} presents Production Efficiency scores for top-performing models from Table~\ref{tab:comprehensive_embedding_results}. Scores are normalized across 25 state-of-the-art embedding models to enable fair comparison. Our \textbf{newmindai/bge-m3-stsb} (567M parameters) achieves a Production Efficiency score of 94.38, ranking third overall. Our \textbf{newmindai/Mursit-Base-TR-Retrieval} (155M parameters) achieves a score of 92.36, ranking fourth overall despite using fewer parameters than the top-performing models (embeddinggemma-300m: 307M, bge-m3: 567M). This shows that domain-specific pre-training combined with efficient architecture design enables practical deployment scenarios where computational resources are constrained. Our \textbf{newmindai/Mursit-Large-TR-Retrieval} (403M parameters) achieves a score of 91.26, ranking fifth. Maximum sequence length, embedding dimension, and parameter count are critical factors that directly and significantly influence overall system production efficiency, with smaller models benefiting from reduced memory requirements and faster indexing speeds.
\pagestyle{plain}
\section{Reward Model Architecture Details}\label{app:reward_architecture}
The Muhakim reward model employs a multi-objective framework distinguishing input and output-dependent components. The \emph{gating mechanism} operates in a prompt-conditioned manner, adjusting priorities based on legal domain or question type. This separation addresses a limitation of fixed-weight evaluation systems, where the same importance is assigned to all quality dimensions regardless of question characteristics. It is implemented as a mixture-of-experts layer that processes the prompt through a shared encoder and routes the representation to expert networks specialized for legal types. The networks learn to recognize patterns in legal prompts, such as regulatory queries requiring accuracy versus case law emphasizing citation quality. The MoE gating network outputs non-negative coefficients that sum to unity, ensuring weights form a probability distribution. The routing mechanism employs a top-k selection strategy where only relevant experts contribute to the final gating weights, reducing computational overhead compared to full expert activation while maintaining evaluation quality. Conversely, \emph{reward predictions} operate response-conditioned, evaluating text quality across each dimension through five parallel regression heads. Each regression head processes the response through the shared Llama-3.1 encoder to produce scalar reward scores normalized via sigmoid transformation, ensuring reward values remain bounded in the [0,1] interval and facilitating gradient flow during training. The mathematical formulation combines gating coefficients $\alpha_i$ (prompt-dependent) with reward scores $r_i$ (response-dependent) as: $\text{score} = \sum_{i=1}^{5} \alpha_i(\text{prompt}) \cdot r_i(\text{response})$. This design enables the model to weight quality dimensions based on input context while maintaining interpretability through per-dimension scores, allowing practitioners to understand which aspects of legal text quality drive the final assessment.

The training pipeline consists of three stages ensuring stable, interpretable, context-aware reward signals. First, \emph{multi-objective supervision} enables learning of five legal quality dimensions through labeled scores, allowing each head to specialize without interference. This stage establishes baseline performance for each dimension independently, providing a foundation for subsequent preference learning. Second, \emph{preference-based training} of the MoE gating network uses response pairs to learn importance weights, enabling the model to adapt priorities based on context. This stage learns when different quality dimensions should receive higher weights, such as prioritizing statutory accuracy for regulatory questions versus case law citation quality for precedent-based queries. Third, a \emph{debiasing stage} mitigates length-related artifacts, ensuring quality evaluation based on content rather than response length, which is important for legal text where concise and comprehensive responses may both be valid depending on context.

\section{Turkish Quality Filtering Threshold Selection}\label{app:turkish_quality_filter}

To determine optimal filtering thresholds for Turkish morphology-based quality filters, we systematically evaluated threshold combinations for suffix entropy and lemma diversity metrics across the corpus. This evaluation is critical for Turkish, an agglutinative language where morphological richness directly impacts model performance on downstream tasks requiring precise linguistic understanding. The selected configuration (SE $\geq$ 75\%, LD $\geq$ 50\%) balances morphological variation requirements with sufficient token budget, retaining approximately 3.48 billion tokens while enforcing higher morphological diversity compared to less restrictive thresholds. This approach ensures that training data captures the full spectrum of Turkish morphological patterns, including complex nominal case constructions and verb conjugations essential for legal domain applications, thereby improving model generalization across diverse linguistic contexts. Table~\ref{tab:turkish_quality_filter} presents the comprehensive results of this threshold selection process.

The threshold selection process reveals that more restrictive configurations (SE $\geq$ 80\%, LD $\geq$ 70\%) result in excessive data reduction (85-99\%), significantly limiting corpus diversity and potentially degrading model performance on rare morphological patterns. Conversely, less restrictive thresholds (SE $\geq$ 50\%, LD $\geq$ 50\%) retain more tokens but fail to enforce sufficient morphological diversity, compromising the quality of linguistic representations. The optimal configuration (SE $\geq$ 75\%, LD $\geq$ 50\%) achieves a balanced trade-off, maintaining 31\% of the original corpus while ensuring comprehensive morphological coverage essential for robust Turkish legal language modeling.

\begin{table}[H]
\setlength{\abovecaptionskip}{-5pt}
\caption{Data reduction and token-length statistics across suffix entropy and lemma diversity cutoffs.}
\label{tab:turkish_quality_filter}
\centering
\footnotesize
\begin{tabular}{@{}llrrrrr@{}}
\hline
SE (\%) & LD (\%) & DR (\%) & RT & Mean$\pm$STD\\
\hline
50 & 50 & 28.8\% & 8,301,450,232 & 586 $\pm$ 702\\
50 & 70 & 95.6\% & 337,745,214 & 635 $\pm$ 466\\
50 & 75 & 98.2\% & 78,401,681 & 525 $\pm$ 261\\
50 & 80 & 99.1\% & 17,703,932 & 326 $\pm$ 119\\
50 & 85 & 99.5\% & 5,224,718 & 177 $\pm$ 65\\
50 & 90 & 99.7\% & 1,791,945 & 45 $\pm$ 40\\
\hline
70 & 50 & 52.0\% & 5,511,412,625 & 542 $\pm$ 691\\
70 & 70 & 97.0\% & 221,509,571 & 594 $\pm$ 437\\
70 & 75 & 98.7\% & 45,029,331 & 462 $\pm$ 213\\
70 & 80 & 99.3\% & 8,571,350 & 230 $\pm$ 78\\
70 & 85 & 99.6\% & 2,921,785 & 96 $\pm$ 47\\
70 & 90 & 99.8\% & 1,290,702 & 22 $\pm$ 36\\
\hline
\textbf{75} &\textbf{50} &\textbf{69.0\%} &\textbf{3,476,853,258} &\textbf{506 $\pm$ 675}\\
75 & 70 & 97.7\% & 148,219,275 & 554 $\pm$ 396\\
75 & 75 & 99.0\% & 30,395,603 & 414 $\pm$ 184\\
75 & 80 & 99.5\% & 6,218,802 & 192 $\pm$ 69\\
75 & 85 & 99.7\% & 2,284,263 & 75 $\pm$ 44\\
75 & 90 & 99.8\% & 1,068,025 & 21 $\pm$ 35\\
\hline
80 & 50 & 85.3\% & 1,559,029,748 & 475 $\pm$ 638\\
80 & 70 & 98.5\% & 73,866,169 & 486 $\pm$ 304\\
80 & 75 & 99.2\% & 16,917,511 & 332 $\pm$ 134\\
80 & 80 & 99.6\% & 4,266,928 & 144 $\pm$ 58\\
80 & 85 & 99.7\% & 1,762,443 & 54 $\pm$ 40\\
80 & 90 & 99.8\% & 879,816 & 19 $\pm$ 33\\
\hline
85 & 50 & 95.2\% & 434,347,440 & 468 $\pm$ 547\\
85 & 70 & 99.2\% & 24,891,788 & 357 $\pm$ 180\\
85 & 75 & 99.5\% & 7,378,437 & 206 $\pm$ 86\\
85 & 80 & 99.7\% & 2,605,621 & 91 $\pm$ 48\\
85 & 85 & 99.8\% & 1,210,163 & 49 $\pm$ 36\\
85 & 90 & 99.9\% & 645,765 & 18 $\pm$ 31\\
\hline
90 & 50 & 98.8\% & 62,193,294 & 417 $\pm$ 313\\
90 & 70 & 99.5\% & 6,308,954 & 200 $\pm$ 83\\
90 & 75 & 99.7\% & 2,975,307 & 105 $\pm$ 55\\
90 & 80 & 99.8\% & 1,407,302 & 32 $\pm$ 38\\
90 & 85 & 99.9\% & 731,717 & 20 $\pm$ 31\\
90 & 90 & 99.9\% & 421,114 & 16 $\pm$ 27\\
\hline
\end{tabular}

\parbox{\columnwidth}{\scriptsize\textit{SE: Suffix Entropy Ratio; LD: Lemma Diversity Ratio; DR: Data Reduction; RT: Remaining Tokens}}

\end{table}

\section{Data Preprocessing and Corpus Construction}\label{app:data_preprocessing}

The preprocessing pipeline prioritizes legal compliance, semantic diversity, and training stability over raw corpus size, employing conservative and fully reproducible transformations. All stages execute in batch-parallel form on the MareNostrum 5 supercomputer using GPU-accelerated cuDF workflows, achieving order-of-magnitude speedups over CPU-based processing. Structurally invalid records are removed, followed by exact deduplication and strict legal filtering. Documents with unclear licensing were proactively excluded to ensure legal compliance. Web data undergoes additional filtering via GPU-parallel URL-based safety rules. OCR-induced noise, particularly scattered-text artifacts, is addressed through a recovery-oriented normalization pipeline using LLM-based reconstruction. To eliminate near-duplicate legal templates, semantic deduplication is applied using dense document embeddings generated by the BGE-M3 encoder~\cite{bgem32024} and filtered with SemHash~\cite{semhash2024} at a cosine similarity threshold of 0.95. The final pipeline—spanning cleaning, filtering, OCR normalization, exact and semantic deduplication, and curriculum assignment—reduces the corpus from 132.6M to 72.2M documents while preserving linguistically meaningful content and ensuring a legally compliant, high-quality dataset suitable for large-scale pretraining.

\section{Decontamination Module}\label{app:decontamination}

The decontamination module applies additional filtering stages to the preprocessing pipeline described in Section~\ref{app:data_preprocessing}. It removes profanity, filters low-quality content using C4~\cite{c4dataset2019}, FineWeb~\cite{fineweb22025}, and Gopher~\cite{glotlid2023,gopher2021} filters, anonymizes sensitive personal information (emails, phone numbers, URLs, national identifiers, financial numbers), filters high-risk domains, normalizes OCR artifacts, and applies language confidence filtering using GlotLID~\cite{glotlid2023}. The module operates using GPU-accelerated cuDF workflows with deterministic execution.

\end{document}